%% file: paper.tex
\documentclass[twoside]{article}

\usepackage[accepted]{aistats2019}
\usepackage[
    natbib,
    citestyle=numeric-comp,
    backref=false,
    maxbibnames=99,
    firstinits=false,
    maxcitenames=2,
    doi=false,
    isbn=false,
    url=false,
    backend=bibtex8
]{biblatex}
\AtEveryBibitem{
 \clearlist{address}
  \clearfield{pages}
  \clearfield{volume}
 \clearfield{date}
 \clearfield{eprint}
 \clearfield{isbn}
 \clearfield{issn}
 \clearlist{location}
 \clearfield{month}
 \clearfield{series}
 
 \ifentrytype{book}{}{
  \clearlist{publisher}
  \clearname{editor}
 }
}

\usepackage[utf8]{inputenc} 
\usepackage[T1]{fontenc}    
\usepackage{hyperref}       
  
\usepackage{url}            
\usepackage{booktabs}       
\usepackage{multicol}
\usepackage{multirow}
\usepackage{array}			
\usepackage{amsfonts}       
\usepackage{nicefrac}       
\usepackage{graphicx}       
\usepackage[table]{xcolor}
	\definecolor{lightgray}{gray}{0.95}
    \definecolor{C1}{HTML}{1F77B4}
    \definecolor{C2}{HTML}{FF7F0E}
    \definecolor{C3}{HTML}{2CA02C}
    \definecolor{C4}{HTML}{D62728}
    \definecolor{C5}{HTML}{9467BD}
    \colorlet{C1light}{C1!20!white}
    \colorlet{C2light}{C2!20!white}
    \colorlet{C3light}{C3!20!white}
    \colorlet{C4light}{C4!20!white}
    \colorlet{C5light}{C5!20!white}

\usepackage[activate={true,nocompatibility},final,tracking=true,kerning=true,spacing=true,factor=1100,stretch=10,shrink=20]{microtype}
\SetTracking{encoding={*}, shape=sc}{20}

\usepackage{fontawesome}
    \DeclareFontFamily{U}{fontawesome1}{}
    \DeclareFontShape{U}{fontawesome1}{m}{n}{<->FontAwesome--fontawesomeone}{}
    \DeclareFontFamily{U}{fontawesome2}{}
    \DeclareFontShape{U}{fontawesome2}{m}{n}{<->FontAwesome--fontawesometwo}{}
    \DeclareFontFamily{U}{fontawesome3}{}
    \DeclareFontShape{U}{fontawesome3}{m}{n}{<->FontAwesome--fontawesomethree}{}

\usepackage[linesnumbered, ruled, vlined]{algorithm2e}
    \newcommand{\nosemic}{\SetEndCharOfAlgoLine{\relax}}

\usepackage[normalem]{ulem}	
\usepackage[inline]{enumitem}
  \newlist{inlinelist}{enumerate*}{1}
  \setlist*[inlinelist,1]{%
          label=(\roman*),
      }
\usepackage{bm}
\usepackage{bbm} 
\usepackage{mathtools}
\usepackage{pgfmath}
\usepackage{pgf}
\usepackage{tikz}
\usepackage{pgfplots}
	\usepgfplotslibrary{groupplots}
  \usetikzlibrary{calc}
  \usetikzlibrary{positioning}
  \usetikzlibrary{angles,quotes}
  \usetikzlibrary{backgrounds}
  \usetikzlibrary{external}
  \usetikzlibrary{fit}
  \usetikzlibrary{bayesnet}
  \usetikzlibrary{calc}
  \usetikzlibrary{fit}
  \usetikzlibrary{spy, shadows}
  \usetikzlibrary{arrows.meta}
  \usetikzlibrary{shadings}
  \usetikzlibrary{fadings}
\usetikzlibrary{matrix}
\usepackage{caption}
\usepackage[capitalize]{cleveref}
\usepackage{subcaption}
\usepackage{amsfonts}
\usepackage{amsmath}
\usepackage{standalone}
\usepackage{xargs}
\usepackage{amssymb}
\usepackage{mathtools}

\usepackage{placeins}
\usepackage{dblfloatfix}

\usepackage{scrextend}
\usepackage{footnote}
\makesavenoteenv{tabular}
\makesavenoteenv{table}

\newcommand{\vz}{\mathbf{z}}
\newcommand{\vx}{\mathbf{x}}
\newcommand{\vX}{\mathbf{X}}
\DeclareMathOperator{\Prob}{\mathrm{Pr}}
\DeclareMathOperator{\Exp}{\mathbb{E}}
\DeclareMathOperator{\KL}{\mathrm{KL}}
\DeclarePairedDelimiterX{\infdivx}[2]{(}{)}{%
  #1\;\delimsize\|\;#2%
}
\newcommand{\kld}{\KL\infdivx}

\renewcommand{\d}{\mathrm{d}}
\newcommand{\Normal}[1]{\mathcal{N}\!\left(#1\right)}
\newcommand{\diag}[1]{\operatorname{diag}\!\left(#1\right)}

\newcommand{\VAE}{VAE}
\newcommand{\HVAE}{HVAE}

\newcommand{\ConvHVAE}{ConvHVAE}

\newcommand{\MLP}[1]{\ensuremath{\mathrm{MLP}[#1]}}
\newcommand{\CNN}[1]{\ensuremath{\mathrm{CNN}[#1]}}

\usepackage[colorinlistoftodos,prependcaption,textsize=tiny]{todonotes}

\newcommand{\mbadd}[1]{{\color{red} #1}}
\newcommand{\amadd}[1]{{\color{blue} #1}}

\newcommand{\Lars}{\textsc{Lars}}

\usepackage[compact]{titlesec}
\titlespacing*{\paragraph} {0pt}{1ex plus 0.5ex minus .2ex}{0.5em}

\begin{document}
\begin{refsection}

\twocolumn[

\aistatstitle{Resampled Priors for Variational Autoencoders}

\aistatsauthor{ {Matthias Bauer\footnotemark} 
\And {Andriy Mnih} }
\aistatsaddress{ {Max Planck Institute for Intelligent Systems} \\  {University of Cambridge} \And  {DeepMind} }
]

\footnotetext{Work done during an internship at DeepMind.}

\begin{abstract}
We propose Learned Accept/Reject Sampling (\Lars{}), a method for constructing richer priors using rejection sampling with a learned acceptance function. This work is motivated by recent analyses of the VAE objective, which pointed out that commonly used simple priors can lead to underfitting.
As the distribution induced by 
\Lars{} involves an intractable normalizing constant, we show how to estimate it and its gradients efficiently.
We demonstrate that \Lars{} priors improve VAE performance on several standard datasets both when they are learned jointly with the rest of the model and when they are fitted to a pretrained model. 
Finally, we show that \Lars{} can be combined with existing methods for defining flexible priors for an additional boost in performance. 

\end{abstract}

\section{Introduction}

Variational Autoencoders (VAEs) \citep{Kingma_vae, Rezende2014_vae} are powerful and widely used probabilistic generative models. 
In their original formulation, both the prior and the variational posterior are parameterized by factorized Gaussian distributions. Many approaches have proposed using more expressive distributions to overcome these limiting modelling choices; most of these focus on flexible variational posteriors \citep[e.g.][]{Salimans2015_bridging_gap,Rezende2015_flows,KingmaIAF2016,Tran_Variational_GP_2015}.
More recently, both the role of the prior and its mismatch to the \emph{aggregate} variational posterior have been investigated in detail \citep{ELBO_Surgery, Rosca_Distribution_Matching_2018}. 
It has been shown that the prior minimizing the KL term in the ELBO is given by the corresponding aggregate posterior \citep{Jakub_Vamp_2018}; however, this choice usually leads to overfitting as this kind of non-parametric prior essentially memorizes the training set. The VampPrior \citep{Jakub_Vamp_2018} models the aggregate posterior explicitly using a mixture of posteriors of learned virtual observations with a fixed number of components. Other ways of specifying expressive priors include flow-based \citep{Chen_variational_lossy_vae} and autoregressive models \citep{gulrajani2016pixelvae}, which provide different tradeoffs between expressivity and efficiency. 

In this work, we present Learned Accept/Reject Sampling, an alternative way to address the mismatch between the prior and the aggregate posterior in VAEs. The proposed \Lars{} prior is the result of applying a rejection sampler with a learned acceptance function to the original prior, which now serves as the proposal distribution.
The acceptance function, typically parameterized using a neural network, can be learned either jointly with the rest of the model by maximizing the ELBO, or separately by maximizing the likelihood of samples from the aggregate posterior of a trained model. Our approach
is orthogonal to most existing approaches for making priors more expressive
and can be easily combined with them by using them as proposals for the sampler. The price we pay for richer priors is iterative sampling, which can require rejecting a large number of proposals before accepting one. 
To improve the acceptance rate and, thus, the sampling efficiency of the sampler, we introduce a truncated version that always accepts the next proposal if some predetermined number of samples have been rejected in a row.

\Lars{} is neither limited to variational inference nor to continuous variables. In fact, it can be thought of as a generic method for transforming a tractable base distribution into a more expressive one by modulating its density while keeping learning or exact sampling tractable.

Our main contributions are as follows:
\begin{itemize}[topsep=-2pt,itemsep=2pt,partopsep=2pt, parsep=2pt, 
]
    \item We introduce Learned Accept/Reject Sampling,  
    a method for transforming a tractable base distribution into a better approximation to a target density, based only on samples from the latter.
    \item We develop a truncated version of \Lars{} which enables a principled trade-off between approximation accuracy and sampling efficiency.
    \item We apply \Lars{} to VAEs, replacing the standard Normal priors with \Lars{} priors. This consistently yields improved ELBOs, with further gains achieved using expressive flow-based proposals.
    \item We demonstrate the versatility of our method by applying it to the high-dimensional discrete output distribution of a VAE.
\end{itemize}

\section{Variational Autoencoders}
A VAE \citep{Rezende2014_vae, Kingma_vae} models the distribution of observations $\vx$ using a stochastic latent vector $\vz$, by specifying its prior $p(\vz)$ along with a likelihood $p_\theta(\vx|\vz)$ that connects it with the observation.
As computing the marginal likelihood of an observation $p_\theta(\vx) = \int p_\theta(\vx,\vz) d\vz $ and, thus, maximum likelihood learning are intractable in VAEs, they are trained by maximizing the variational \emph{evidence lower bound} (ELBO) on the marginal log-likelihood:
\begin{equation}
    \label{eq:elbo}
    \begin{aligned}
     \hspace{-0.25em} \Exp_{q(\vx)}& \left[ \Exp_{q_\phi(\vz|\vx)}\log p_\theta(\vx|\vz)  - \KL(q_\phi(\vz|\vx) || p_\lambda(\vz))\right] \hspace{-0.25em}
    \end{aligned}
\end{equation}
where $q(\vx)$ is the empirical distribution of the training set and $q_\phi(\vz|\vx)$ is the \emph{variational posterior}. The neural networks used to parameterize $q_\phi(\vz|\vx)$ and $p_\theta(\vx|\vz)$ are referred to as the \emph{encoder} and \emph{decoder}, respectively. In the rest of the paper we omit the parameter subscripts to reduce notational clutter. The gradients of the ELBO w.r.t.~all parameters can be estimated efficiently using the reparameterization trick \citep{Kingma_vae, Rezende2014_vae}.

The first term in \cref{eq:elbo}, the (negative) reconstruction error, measures how well a sample $\vx$ can be reconstructed by the decoder from its latent space representation produced by the encoder. The second term, the Kullback-Leibler (KL) divergence between the variational posterior and the prior, acts as a regularizer that limits the amount of information passing through the latents and ensures that the model can also generate the data rather than just reconstruct it.
We can rewrite the KL in terms of the \emph{marginal} or \emph{aggregate (variational) posterior} $q(\vz) = \Exp_{q(\vx)}[q(\vz|\vx)]$ \citep{ELBO_Surgery}:
    \begin{align}
        \hspace{-.75em} \Exp_{q(\vx)} \mathrm{KL}(q(\vz|\vx) || p(\vz))  = \mathrm{KL}(q(\vz) || p(\vz)) +  I(\vx; \vz),
    \end{align}
where the second term is the mutual information between the latent vector and the training example.
This formulation makes it clear that maximizing the ELBO corresponds to minimizing the KL between the aggregate posterior and the prior. Despite this, there usually is a mismatch between the two distributions in trained models \citep{Rosca_Distribution_Matching_2018, ELBO_Surgery}. This phenomenon is sometimes described as ``holes in the aggregate posterior'', referring to the regions of the latent space that have high density under the prior but very low density under the aggregate posterior, meaning that they are almost never encountered during training. Samples from these regions are typically decoded to observations that do not lie on the data manifold \citep{Rosca_Distribution_Matching_2018}. 

Our goal is to make the prior $p(\vz)$ sufficiently expressive to make it substantially closer to $q(\vz)$ compared to the original prior, without making the two identical, as that can result in overfitting. 
In the following we introduce a simple and effective way of achieving this by learning to reject the samples from the proposal that have low density under the aggregate posterior. 

\section{Learned Accept/Reject Sampling}

\begin{figure}[htb]
    \centering
    \includestandalone[mode=buildnew]{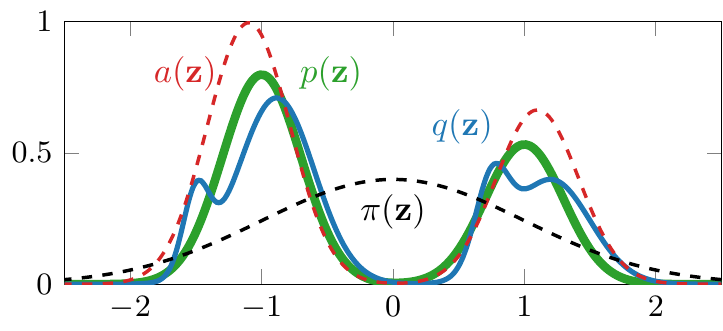}
     \vskip-1.25ex
    \caption[]{\Lars{} approximates a target density $q(\vz)$ (\protect\tikz[baseline=(current bounding box.base),inner sep=0pt]{\draw[line width=1.5pt, solid, C1] (0pt,3pt) --(15pt,3pt);}) by a resampled density $p(\vz)$ (\protect\tikz[baseline=(current bounding box.base),inner sep=0pt]{\draw[line width=2.5pt, solid, C3] (0pt,3pt) --(15pt,3pt);}) obtained by multiplying a proposal $\pi(\vz)$ (\protect\tikz[baseline=(current bounding box.base),inner sep=0pt]{\draw[line width=1.5pt, dashed] (0pt,3pt) --(15pt,3pt);}) with a learned acceptance function $a(\vz) \in [0, 1]$ (\protect\tikz[baseline=(current bounding box.base),inner sep=0pt]{\draw[line width=1.5pt, solid, C4, dashed] (0pt,3pt) --(15pt,3pt);}) and normalizing.}
    \label{fig:learned_rejection_sampler}
\end{figure}

In this section we present our main contribution: \emph{Learned Accept/Reject Sampling (\Lars{}), a practical method to approximate a target density $q$ from which we can sample but whose density is either too expensive to evaluate or unknown}. It works by reweighting a simpler proposal distribution $\pi$ using a learned acceptance function $a$ as visualized in \cref{fig:learned_rejection_sampler}. The aggregate posterior of a VAE, which is a very large mixture, is one example of such a target distribution.
For simplicity, we only deal with continuous variables in this section; however, our approach is equally applicable to discrete random variables as we explain in \cref{sec:output_vae}. 


\vskip-0.25em
\vskip-0.5em

%
%
\begin{figure*}[ht!]
    \includestandalone[width=\textwidth, mode=tex]{figures/2d_toy_figure}
     \vskip-1ex
    \caption[short]{Learned acceptance functions $a(\vz)$ (red) that approximate a fixed target $q(\vz)$ (blue) by reweighting a $\mathcal{N}(0,1)$ (\protect\tikz[baseline=(current bounding box.base),inner sep=0pt]{\draw[line width=1pt, dashed] (0pt,3pt) --(15pt,3pt);}) or a RealNVP (green) proposal to obtain an approximate density $p(\vz)$ (green). $\kld*{q\!}{\!\pi_{\mathcal{N}(0,1)}}= 1.8$.}
    \label{fig:toy}
\end{figure*}

\paragraph{Approximating a target by modulating a proposal.} 
Our goal is to approximate a complex target distribution $q(\vz)$ that we can sample from but whose density we are unable to evaluate.  We start with an easy-to-sample-from base distribution $\pi(\vz)$ with a tractable density (with the same support as $q$), which might have learnable parameters but is insufficiently expressive to approximate $q(\vz)$ well. We then transform $\pi(\vz)$ into a more expressive distribution $p_\infty(\vz)$ by multiplying it by an \emph{acceptance function} $a_\lambda(\vz) \in [0, 1]$ and renormalizing to obtain the density
\begin{align}
    \label{eq:resampled_density_indefinite}
    p_\infty(\vz) & = \frac{\pi(\vz) a_\lambda(\vz)}{Z}; && Z = \int \pi(\vz)a_\lambda(\vz)\d \vz,
\end{align}
where $Z$ is the normalization constant and $\lambda$ are the learnable parameters of $a_\lambda(\vz)$; 
we drop the subscript $\lambda$ in the following.
Similar to other approaches for specifying expressive densities, the parameters of $a(\vz)$ can be learned end-to-end by optimizing the natural objective for the task at hand.
$p_\infty$ is also the distribution we obtain by using rejection sampling \citep{vonNeumann_rejection} with $\pi(\vz)$ as the proposal and $a(\vz)$ as the acceptance probability function. Thus, we can sample from $p_\infty$ as follows: draw a candidate sample $\vz$ from the proposal $\pi$ and accept it with probability $a(\vz)$; if accepted, return $\vz$ as the sample; otherwise repeat the process. This means we can think of $a(\vz)$ as a filtering function which --- by rejecting different fractions of samples at different locations --- can transform $\pi(\vz)$ into a distribution closer to $q(\vz)$. 
In fact, as we show in Section \ref{sec:rejection_sampling_101}, if the capacity of $a(\vz)$ is unlimited, we can obtain $p_\infty(\vz)$ that matches the target distribution perfectly.

The efficiency of the sampler is closely related to the normalizing constant $Z$, 
which can be interpreted as the average probability of accepting a candidate sample. Thus, on average, we will consider $1/Z$ candidates before accepting one.
Interestingly, the same distribution can be induced by rejection samplers of very different efficiency, since scaling $a(\vz)$ by a constant (while keeping $a$ in the $[0, 1]$ range) has no effect on the resulting $p_\infty(\vz)$ because $Z$ gets scaled accordingly. This means that learning an efficient sampler might require explicitly encouraging $a$ to be as large as possible.

\paragraph{Truncated rejection sampling.}
We now describe a simple and effective way to guarantee a minimum sampling efficiency that uses a modified density $p_T$ instead of $p_\infty$. We derive it from a \emph{truncated sampling scheme} (see \cref{app:resampling_truncated} for the derivation):
instead of allowing an arbitrary number of candidate samples to be rejected before accepting one,
we cap this number and always accept the $T^\text{th}$ sample if the preceding $T-1$ samples have been rejected. The corresponding density is a mixture of $p_\infty(\vz)$ and the proposal $\pi(\vz)$, with the mixing weights determined by the average rejection probability $(1-Z)$ and the truncation parameter $T$:
\begin{align}
    \label{eq:resampled_density_truncated}
    p_T(\vz) & = (1-\alpha_T) \frac{a(\vz) \pi(\vz)}{Z}  + \alpha_T \pi(\vz) ,
\end{align}
where $\alpha_T=(1-Z)^{T-1}$. Note that for $T=1$ we recover the proposal density $\pi$, while letting $T\rightarrow\infty$ yields the density for untruncated sampling (\cref{eq:resampled_density_indefinite}).

While the truncated sampling density will in general be less expressive due to smoothing by interpolation with $\pi$, its acceptance rate is guaranteed to be at least $1/T$, with the expected number of resampling steps being $(1-(1-Z)^T)/Z \leq T$; see \cref{app:resampling_truncated} for details. Thus we can trade off approximation accuracy against sampling efficiency by varying $T$. 
Using the truncated sampling density also has the benefit of providing a natural scale for $a$ in an interpretable and principled manner as, unlike $p_\infty(\vz)$, $p_T(\vz)$ is not invariant under scaling $a$  by $c \in (0, 1]$.



\paragraph{Estimating $Z$.} Estimating the normalizing constant $Z$ is the main technical challenge of our method. An estimate of $Z$ is needed both for updating the parameters of $p_T(\vz)$ and for evaluating trained models. For reliable model evaluation, we use the Monte Carlo estimate
\begin{align}
    \label{eq:Z_MC}
    Z \approx \frac{1}{S} \sum_{s=1}^S a(\vz_s) && \vz_s \sim \pi(\vz),
\end{align}
which we compute only once, using a very large $S$ when the model is fully trained.
For training, we found it sufficient to estimate $Z$ once per batch using a modest number of samples. In the forward pass, we use a version of $Z$ that is smoothed with exponential averaging, whereas the gradients in the backward pass are computed on samples from the current batch only. We also include a reweighted sample from the target in the estimator to decrease the variance of the gradients (similar to~\citep{botev17acomplementary}). See \cref{app:sec:Z_MC} for a detailed explanation.

\subsection{Illustrative examples in 2D}
Before applying \Lars{} to VAEs, we investigate its ability to fit a mixture of Gaussians (MoG) target distribution in 2D, see \cref{fig:toy} \textit{(left)}. We use a standard Normal distribution as the proposal and train an acceptance function parameterized by a small MLP, to reweight $\pi$ to $q$ by maximizing the supervised objective
\begin{equation}
    \label{eq:supervised_learning_a}
    \Exp_{q(\vz)} \left[ \log p_\infty(\vz) \right] = \Exp_{q(\vz)} \left[ \log \frac{\pi(\vz)a(\vz)}{Z} \right],
\end{equation}
which amounts to maximum likelihood learning. 

We find that a small network for $a(\vz)$ (an MLP with two layers of 10 units) is sufficient to separate out the individual modes almost perfectly (\cref{fig:toy} (\textit{left})). However, an even simpler network (one layer with 10 units) already can learn to cut away the unnecessary tails of the proposal and leads to a better fit than the proposal (\cref{fig:toy} (\textit{middle})). While samples from the higher-capacity model are closer to the target (lower KL to the target), the sampling efficiency is lower (smaller $Z$). 

As a second example, we consider the same target distribution but use a RealNVP \citep{RealNVP_2017} proposal which is trained jointly with the acceptance function, see \cref{fig:toy} (\text{right}). Trained alone, the RealNVP warps the standard Normal distribution into a multi-modal distribution, but fails to isolate all the modes (\cref{fig:app:toy_realnvp} in the Appendix, also shown by \citet{NAF_2018}). On the other hand, the model obtained by using \Lars{} with a RealNVP proposal manages to isolate all the modes. It also achieves higher average accept probability than \Lars{} with a standard Normal proposal, as the RealNVP proposal approximates the target better, see \cref{fig:toy} (right).


\subsection{Relationship to rejection sampling} 
\label{sec:rejection_sampling_101}
Here we briefly review classical rejection sampling (RS) \citep{vonNeumann_rejection} and relate it to \Lars{}. RS provides a way to sample from a target distribution with a known density $q(\vz)$ which is hard to sample from directly but is easy to evaluate point-wise. Note that this is \emph{exactly the opposite} of our setting, in which $q(\vz)$ is easy to sample from but  infeasible to evaluate. By drawing samples from a tractable proposal distribution $\pi(\vz)$ and accepting them with probability $a(\vz) = \frac{q(\vz)}{M \pi(\vz)}$, where $M$ is chosen so that $\pi(\vz) M \geq q(\vz), \forall \vz$, RS generates exact samples from $q(\vz)$. The average sampling efficiency, or acceptance rate,  of this sampler is $1/M$, which means the less the proposal needs to be scaled to dominate the target, the fewer samples are rejected. In practice, it can be difficult to find the smallest valid $M$ as $q(\vz)$ is usually a complicated multi-modal distribution.

We can match $q(\vz)$ exactly with \Lars{} by emulating classical RS and setting $a^\ast(\vz) = c \frac{q(\vz)}{\pi(\vz)}$ with the constant $c$ chosen so that $a^\ast(\vz)\leq 1$ for all $\vz$.\footnote{This assumes that $a^\ast$ has unlimited capacity.} Noting that this results in $Z = c$ and comparing this to standard RS, we find that $Z$ plays the role of $1/M$. However, with \Lars{} our primary goal is learning a compact and fast-to-evaluate density that approximates $q(\vz)$, rather than constructing a way to sample from $q(\vz)$ exactly, which is already easy in our setting. 

\section{VAEs with resampled priors}

While \Lars{} can be used to enrich several distributions in a VAE, we will concentrate on applying it to the prior; that is, we use the original VAE prior as a proposal distribution $\pi(\vz)$ and define a \emph{resampled prior} $p(\vz)$ through a rejection sampler with learned acceptance probability $a(\vz)$. We parameterize $a(\vz)$ with a flexible neural network and use the terms ``\Lars{} prior'' and ``resampled prior'' interchangeably.

In general, the acceptance function is trained jointly with the other parts of the model by optimizing the ELBO objective (\cref{eq:elbo}) but with the resampled prior instead of the standard Normal prior. This only modifies the prior term $\log p(\vz)$ in the objective, with $p(\vz)$  replaced by \cref{eq:resampled_density_indefinite} for indefinite and by \cref{eq:resampled_density_truncated} for truncated resampling. See \cref{app:sec:vae_training} for pseudo-code and further details. 
Thus, training a VAE with a resampled prior only requires \emph{evaluating the resampled prior density} on samples from the variational posterior as well as generating a modest number of samples from the proposal to update the moving average estimate of $Z$ as described above. \emph{Crucially, we never need to perform accept/reject sampling during training}. Thus, we can choose the maximum number of resampling steps $T$ based on computational budget \emph{at sampling time}. 



\paragraph{Hierarchical models.} In addition to VAEs with a single stochastic layer, 
we also consider hierarchical VAEs with two stochastic layers. 
We closely follow the proposed inference and generative structure of \citep{Jakub_Vamp_2018} and apply the resampled prior to the top-most latent layer. Thus, the generative distribution and the variational posterior factorize as
$p_\theta(\vx|\vz_1, \vz_2) p_\lambda(\vz_1|\vz_2) p_\lambda(\vz_2)$ and
$q_\psi(\vz_1|\vx, \vz_2) q_\phi(\vz_2|\vx)$ respectively, with:
\begin{align}
    p_\lambda(\vz_2) & = p_T(\vz_2) \\
    p_\lambda(\vz_1|\vz_2) & = \Normal{\vz_1|\mu_\lambda(\vz_2), \diag{\sigma^2_\lambda(\vz_2)}} \\
    q_\phi(\vz_2|\vx) &= \Normal{\vz_2|\mu_\phi(\vx), \diag{\sigma^2_\phi(\vx)}} \\
    q_\psi(\vz_1|\vx, \vz_2) &= \Normal{\vz_1|\mu_\psi(\vx, \vz_2), \diag{\sigma^2_\psi(\vx, \vz_2)}}
\end{align}
The means $\mu_{\{\mu, \phi, \psi\}}$ and standard deviations $\sigma_{\{\mu, \phi, \psi\}}$ are parameterized by 
neural networks. We will apply resampling to the top-most prior distribution $p_\lambda(\vz_2)$.

\begin{table*}[htb]
    \centering
    \rowcolors{1}{white}{lightgray}
    \small
    \begin{tabular}{ l  >{$}c<{$} >{$}c<{$}  >{$}c<{$} >{$}c<{$}  >{$}c<{$}  >{$}c<{$} }
    \toprule
    \hiderowcolors &\multicolumn{2}{c}{\textbf{\VAE{} ($L=1$)}}& \multicolumn{2}{c}{\textbf{\HVAE} ($L=2$)} & \multicolumn{2}{c}{\textbf{\ConvHVAE} ($L=2$)}\\ 
    \textbf{\textsc{Dataset}}& \multicolumn{1}{c}{\small standard} & \multicolumn{1}{c}{\small \Lars{}} & \multicolumn{1}{c}{\small standard} & \multicolumn{1}{c}{\small \Lars{}}  & \multicolumn{1}{c}{\small standard} & \multicolumn{1}{c}{\small \Lars{}} \\ \showrowcolors \midrule
     staticMNIST & 89.11 & \mathbf{86.53} & 85.91& \mathbf{84.42}& 82.63& \mathbf{81.70}\\
    dynamicMNIST & \textcolor{black}{84.97} & \mathbf{\textcolor{black}{83.03}} &  82.66  & \mathbf{81.62} & 81.21 & \mathbf{80.30}\\
    Omniglot        &104.47 &\mathbf{ 102.63} &101.38 & \mathbf{100.40} & 97.76& \mathbf{97.08}\\
     FashionMNIST & \textcolor{black}{228.68} & \textcolor{black}{\mathbf{227.45}}& \textcolor{black}{227.40} & \textcolor{black}{\mathbf{226.68}} & \textcolor{black}{226.39}& \textcolor{black}{\mathbf{225.92}}\\
     \bottomrule
    \end{tabular}
     \vskip-1.25ex
    \caption{Test negative log likelihood (NLL; \emph{lower is better}) for different models with standard Normal prior (``standard'') and our proposed resampled prior (``\Lars{}''). $L$ is the number of stochastic layers in the model.}
    \label{tab:model_comparison}
\end{table*}
\begin{table*}[htb]
    \centering
    \rowcolors{1}{white}{lightgray}
    \small
    \begin{tabular}{l|>{$}c<{$}>{$}c<{$}>{$}c<{$}>{$}c<{$}|>{$}c<{$}>{$}c<{$}>{$}c<{$}>{$}c<{$}|>{$}c<{$}>{$}c<{$}>{$}c<{$}>{$}c<{$}}
    \toprule
    \hiderowcolors & \multicolumn{4}{c}{standard $\mathcal{N}(0,1)$} & \multicolumn{4}{c}{\text{\Lars{} \emph{post-hoc}}}& \multicolumn{4}{c}{\text{\Lars{} \emph{joint training}}}\\
    \textsc{Model} & \text{ELBO} & \text{KL}& \text{recons} &Z &\text{ELBO}  & \text{KL}&\text{recons}& Z& \text{ELBO}  &\text{KL}&\text{recons}& Z\\\showrowcolors\midrule
     dynamicMNIST  & 89.0  &25.8  & 63.2 & 1 & 87.3 &24.1&63.2&0.023&86.6 & 24.7& 61.9 & 0.015\\
     Omniglot & 110.8 &37.2&73.6&1& 108.8&35.2&73.6& 0.015& 108.4  &35.9& 72.5& 0.011\\
    \bottomrule
    \end{tabular}
     \vskip-1.25ex
    \caption{ELBO and its components (KL and reconstruction error) for VAEs with: a standard Normal prior, a post-hoc fitted \Lars{} prior, and a jointly trained \Lars{} prior.}
    \label{tab:posthoc_mnist_omniglot}
\end{table*}

\section{Experiments}

\paragraph{Datasets.} We perform unsupervised learning on a suite of standard datasets: static and dynamically binarized MNIST \citep{NADE_2011_Larochelle}, Omniglot \citep{Lake_Omniglot_2015}, and FashionMNIST \citep{FashionMNIST}; for more details, see \cref{app:datasets}. 

\paragraph{Architectures/Models.} We consider several architectures:
\begin{inlinelist}
    \item a VAE with single latent layer and an MLP encoder/decoder (referred to as \VAE);
    \item a VAE with two latent layers and an MLP encoder/decoder (\HVAE);
    \item a VAE with two latent layers and a convolutional encoder/decoder (\ConvHVAE).
\end{inlinelist}
For the acceptance function we use an MLP network with two layers of  $100$ \texttt{tanh} units and the \texttt{logistic} non-linearity for the output. Moreover, we perform resampling with truncation after $T=100$ attempts (\cref{eq:resampled_density_truncated}), as a good trade-off between approximation accuracy and sampling efficiency at generation time. We explore alternative architectures and truncations below. Full details of all architectures can be found in \cref{app:architectures}.

\paragraph{Training and evaluation.}
All models were trained using the Adam optimizer \citep{Kingma_adam} for $10^7$ iterations with the learning rate $3\cdot 10^{-4}$, which was decayed to $1\cdot 10^{-4}$ after $10^6$ iterations, and mini-batches of size $128$. Weights were initialized according to \citep{Glorot_2010_init}. Unless otherwise stated, we used warm-up (KL-annealing) for the first $10^5$ iterations \citep{Bowman_Warum_Up_2016}. We quantitatively evaluate all methods 
using the negative test log likelihood (NLL) estimated using 1000 importance samples \citep{burda2016importance}. We estimate $Z$ using $S$ MC samples (see \cref{eq:Z_MC}) with $S=10^{10}$ for evaluation after training (this only takes several minutes on a GPU) and $S=2^{10}$ during training, see \cref{app:sec:Z_MC} for further details. We repeated experiments for different random seeds with very similar results, so report only the means. We observed overfitting on static MNIST, so on that dataset we performed early stopping using the NLL on the validation set and halved the times for learning rate decay and warm-up.

\subsection{Quantitative Results}

First, we compare the resampled prior to a standard Normal prior on several architectures, see \cref{tab:model_comparison}. 
For all architectures and datasets considered, the resampled prior outperforms the standard prior.
In most cases, the improvement in the NLL exceeds $1$ nat and is notably larger for the simpler architectures.

Applied \emph{post-hoc} to a pretrained model (fixed encoder/decoder) the resampled prior almost reaches the performance of a jointly trained model, see \cref{tab:posthoc_mnist_omniglot}. This highlights that \Lars{} can also be effectively employed to improve trained models. In this case, the gain in ELBO is entirely due to a reduction in the KL as the reconstruction error is determined by the fixed encoder/decoder. In contrast, joint training reduces both, and reaches a better ELBO, though the reduction in KL is smaller than for post-hoc training.

\renewcommand*{\thefootnote}{\fnsymbol{footnote}}
\begin{table}[htb]
    \centering
    \rowcolors{1}{white}{lightgray}
    \small
    \begin{tabular}{l>{$}c<{$}}
    \toprule
    \hiderowcolors \textsc{Model} & \text{NLL} \\\showrowcolors\midrule
       VAE ($L=2$) + VGP \citep{Tran_Variational_GP_2015} & 81.32 \\
       LVAE ($L=5$) \citep{Ladder_VAE_Sonderby} & 81.74 \\
       HVAE\footnote{We use the same structure but a simpler architecture than \protect\citep{Jakub_Vamp_2018}.\label{footnote_vamp}} ($L=2$) + VampPrior \citep{Jakub_Vamp_2018} & \mathbf{81.24} \\
       \HVAE{} ($L=2$) + \Lars{} prior & 81.62 \\\midrule
       VAE + IAF \citep{KingmaIAF2016} & \textbf{79.10} \\
       \ConvHVAE{}\footref{footnote_vamp} + Vamp \citep{Jakub_Vamp_2018}& 79.75\\
       \ConvHVAE{} + \Lars{} prior & 80.30\\
    \bottomrule
    \end{tabular}
     \vskip-1.25ex
    \caption{Test NLL on dynamic MNIST for non-convolutional (above the line) and convolutional (below the line) models.}
    \label{tab:mnist_baselines}
\vskip2ex
    \centering
    \rowcolors{1}{white}{lightgray}
    \small
    \begin{tabular}{l>{$}c<{$}}
    \toprule
    \hiderowcolors \textsc{Model} & \text{NLL} \\\showrowcolors\midrule
      IWAE ($L=2$) \citep{burda2016importance} & 103.38\\
      LVAE ($L=5$) \citep{Ladder_VAE_Sonderby} & 102.11\\
      HVAE\footref{footnote_vamp} ($L=2$) + VampPrior \citep{Jakub_Vamp_2018} & 101.18\\
      \HVAE{} ($L=2$) + \Lars{} prior & \mathbf{100.40}\\\midrule
      ConvHVAE\footref{footnote_vamp} ($L=2$) + VampPrior \citep{Jakub_Vamp_2018}& 97.56 \\
      \ConvHVAE{} ($L=2$) + \Lars{} prior & 97.08 \\
    \bottomrule
    \end{tabular}
     \vskip-1.25ex
    \caption{Test NLL on dynamic Omniglot.}
    \label{tab:omniglot_baselines}
\end{table}
\renewcommand*{\thefootnote}{\arabic{footnote}}
\addtocounter{footnote}{-1}

In \cref{tab:omniglot_baselines,tab:mnist_baselines} we compare \Lars{} to other approaches on dynamic MNIST and Omniglot. \Lars{} is competitive with other approaches that make the prior or variational posterior more expressive when applied to similar architectures. We expect that our results can be further improved by using PixelCNN-style decoders, which have been used to obtained state-of-the-art results on dynamic MNIST (78.45 nats for PixelVAE with VampPrior \citep{Jakub_Vamp_2018}) and Omniglot (89.83 with VLAE \citep{Chen_variational_lossy_vae}).
The improvements of \Lars{} over the standard Normal prior are generally comparable to those obtained by the VampPrior \citep{Jakub_Vamp_2018}, though they tend to be slightly smaller.  
\Lars{} priors and VampPriors have somewhat complementary strengths. Sampling from VampPriors is more efficient as it amounts to sampling from a mixture model, which is non-iterative. On the other hand, the VampPriors that achieve the above results use 500 or 1000 pseudo-inputs and thus have hundreds of thousands of parameters, whereas the number of parameters in the \Lars{} priors is more than an order of magnitude lower. As \Lars{} priors operate on a low-dimensional latent space they are also considerably more efficient to train than VampPriors, which are connected to the input space through the pseudo-inputs.


\paragraph{Influence of network architecture.}
We investigated 
different network sizes for the acceptance function. As for the illustrative example, even a simple architecture can already notably improve over a standard prior, with successively more expressive networks further improving performance, see \cref{tab:ablation_a_architecture}. Notably, the average acceptance probability $Z$ is influenced much more by the truncation parameter than the network architecture. In practice, we choose $a=\MLP{100-100}$ as more expressive networks only showed a marginal improvement. While we suspect that substantially larger networks might lead to overfitting, we did not observe this for the networks considered in \cref{tab:ablation_a_architecture}. The relative size of the $a$ network w.r.t. encoder and decoder also determines the extra computational cost of \Lars{} over a standard prior. For example, the VAE ($L=1$) results in \cref{tab:model_comparison} require $25\%$ more multiply-adds per iteration; due to our unoptimized implementation, the difference in wall-clock time was closer to $40\%$. 

\begin{table}[tb]
    \centering
    \rowcolors{1}{white}{lightgray}
    \small
    \begin{tabular}{l>{$}c<{$}>{$}c<{$}}
    \toprule
    \hiderowcolors \textsc{Model} & \text{NLL} & Z\\\showrowcolors\midrule
     \VAE{} ($L=1$, MLP) & 84.97& 1\\
    \VAE{} + \Lars{}; $a=\MLP{10-10}$ & 84.08 & 0.02\\
    \VAE{} + \Lars{}; $a=\MLP{50-50}$ & 83.29& 0.016\\
    \VAE{} + \Lars{}; $a=\MLP{100-100}$ & 83.05& 0.015\\
    \VAE{} + \Lars{}; $a=\MLP{50-50-50}$ & 83.09& 0.015\\
    \VAE{} + \Lars{}; $a=\MLP{100-100-100}$ & 82.94& 0.014\\
    \bottomrule
    \end{tabular}
     \vskip-1.25ex
    \caption{Test NLL and $Z$ on dynamic MNIST. Different network architectures for $a(\vz)$ with $T=100$.}
    \label{tab:ablation_a_architecture}
\vskip2ex
    \centering
    \rowcolors{1}{white}{lightgray}
    \small
    \begin{tabular}{>{}l<{} >{$}c<{$}>{$}c<{$}>{$}c<{$}>{$}c<{$}}
    \toprule
    \hiderowcolors \textsc{Model} & \text{NLL} & Z & \mathrm{KL} & \text{recons}\\\showrowcolors\midrule
    \VAE{} ($L=1$, MLP) & 84.97 & 1 & 25.8 & 63.3\\\midrule
    \VAE{} + \Lars{}; $p_{T=2}$ & 84.60 & 0.19 & 25.4 & 63.2\\
    \VAE{} + \Lars{}; $p_{T=5}$ & 84.11 & 0.12 & 25.1 & 63.0\\
    \VAE{} + \Lars{}; $p_{T=10}$ & 83.71 & 0.08 & 25.0 & 62.6\\
    \VAE{} + \Lars{}; $p_{T=50}$ & 83.24 & 0.03 & 24.8 & 62.1\\
    \VAE{} + \Lars{}; $p_{T=100}$ & 83.05 & 0.014 & 24.7 & 61.9\\\midrule
    \VAE{} + \Lars{}; $p_\infty$ & 82.67 & 2 \cdot 10^{-6} &24.8&61.5\\
    \bottomrule
    \end{tabular}
     \vskip-1.25ex
    \caption{Influence of the truncation parameter on the test set of dynamic MNIST; $a=\MLP{100-100}$.
}
    \label{tab:ablation_truncation}
\end{table}
\paragraph{Influence of truncation.}
The truncation parameter $T$ tunes the trade-off between sampling efficiency and approximation quality. 
As shown in \cref{tab:ablation_truncation}, the NLL improves as the number of steps before truncating increases, whereas the value for $Z$ decreases accordingly. The gains in the NLL come both from the KL as well as the reconstruction term, though the relative contribution from the KL is larger.
We stress that for truncated sampling, $Z$ denotes the average acceptance rate \emph{for the first $T-1$ attempts} and does not take into account always accepting the $T^\text{th}$ proposal. As a reference point, a model that is \emph{trained} with indefinite (untruncated) resampling only accepts 2 out of $10^6$ samples and estimating $Z$ accurately thus requires a lot of samples. 

\paragraph{Combination with non-factorial priors.}
\begin{table}[htb]
    \centering
    \rowcolors{1}{white}{lightgray}
    \small
    \begin{tabular}{l>{$}c<{$}>{$}c<{$}}
    \toprule
    \hiderowcolors \textsc{Model} & \text{NLL} & Z\\\showrowcolors\midrule
    \VAE{} ($L=1$) + $\mathcal{N}(0,1)$ prior & 84.97 & 1\\
    \VAE{} + RealNVP prior & 81.86 & 1\\
    \VAE{} + MAF prior & 81.58 & 1\\
    \VAE{} + \Lars{} $\mathcal{N}(0,1)$ & 83.04 & 0.015\\
    \VAE{} + \Lars{} RealNVP & \mathbf{81.15} & 0.027\\
    \bottomrule
    \end{tabular}
     \vskip-1.25ex
    \caption{More expressive proposal and prior distributions. Test NLL on dynamic MNIST.}
    \label{tab:realnvp_proposal_mnist}
\end{table}
%
As stated before, \Lars{} is an orthogonal approach to specifying rich distributions and can be combined with other 
structured proposals such as non-factorial or autoregressive (flow) distributions. We investigated this using a RealNVP distribution as an alternative proposal on MNIST (\cref{tab:realnvp_proposal_mnist}) and Omniglot (\cref{tab:realnvp_proposal_omniglot}), see \cref{app:architectures} for details. The RealNVP prior by itself outperforms our resampling approach applied to a standard Normal prior. However, applying \Lars{} to a RealNVP proposal distribution, we obtain a further improvement of more than $0.5$ nats on both MNIST and Omniglot. A VAE trained with a more expressive Masked Autoregressive Flow (MAF) \citep{MAF_2017} as a prior outperformed the RealNVP prior but was still worse than a resampled RealNVP. Combining \Lars{} with an MAF proposal is impractical due to slow sampling of the MAF.


\subsection{Qualitative Results}
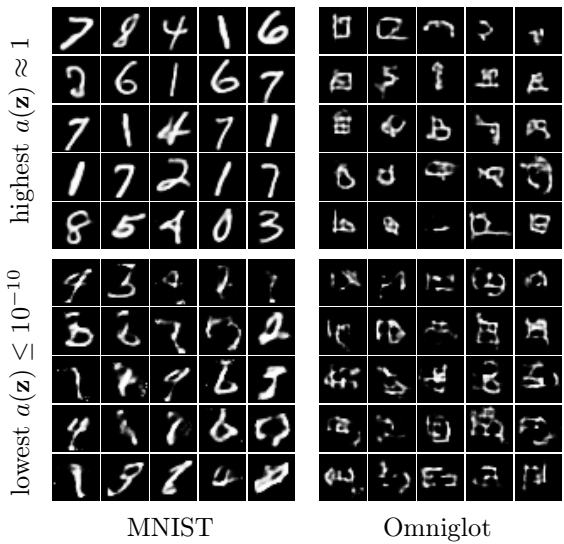
\begin{figure}[tb]
    \centering
    \begin{tikzpicture}
    \node (mnist_good) {\includegraphics[width=1.3in]{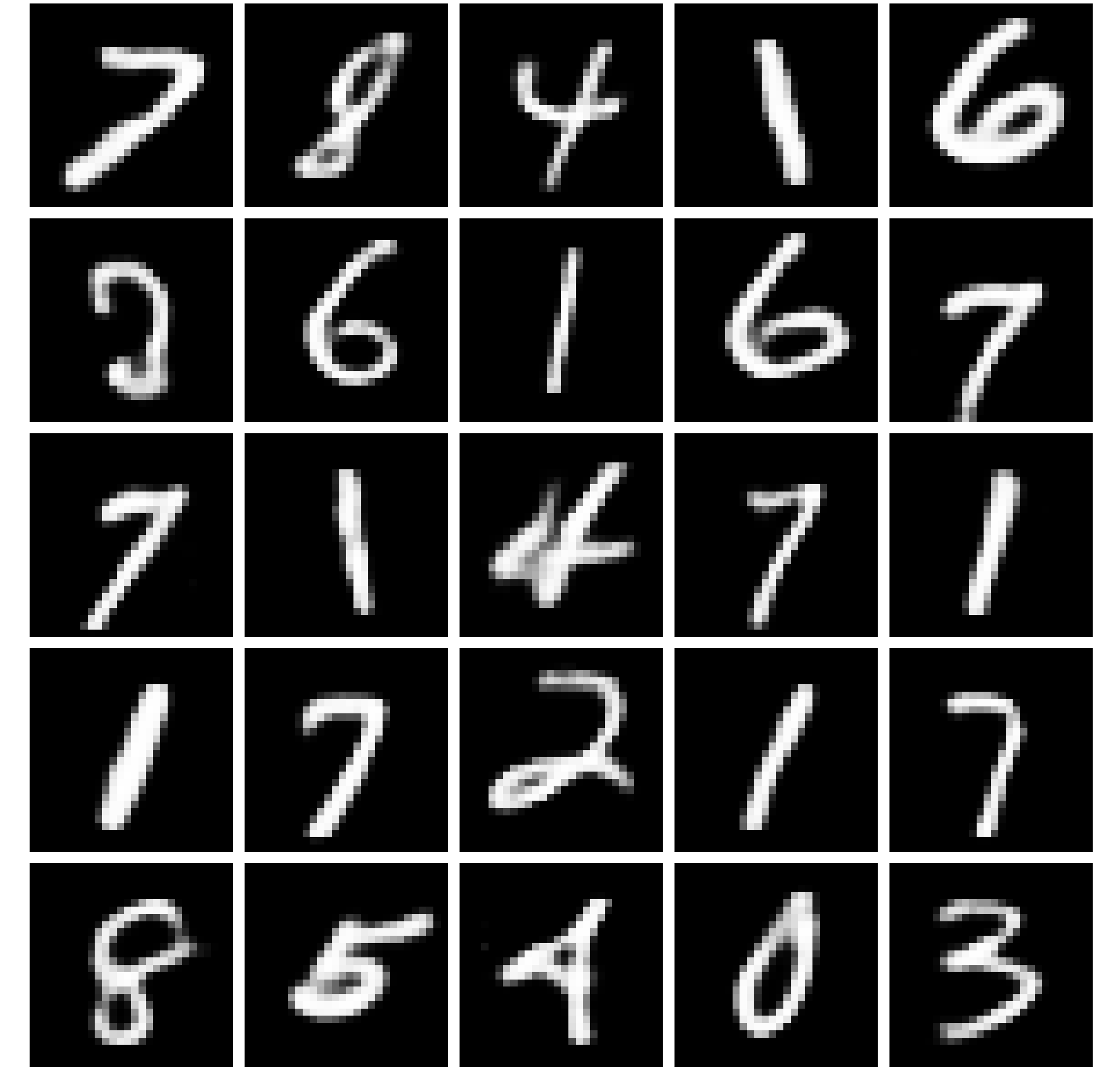}};
    \node[right=0. of mnist_good] (omniglot_good) {\includegraphics[width=1.3in]{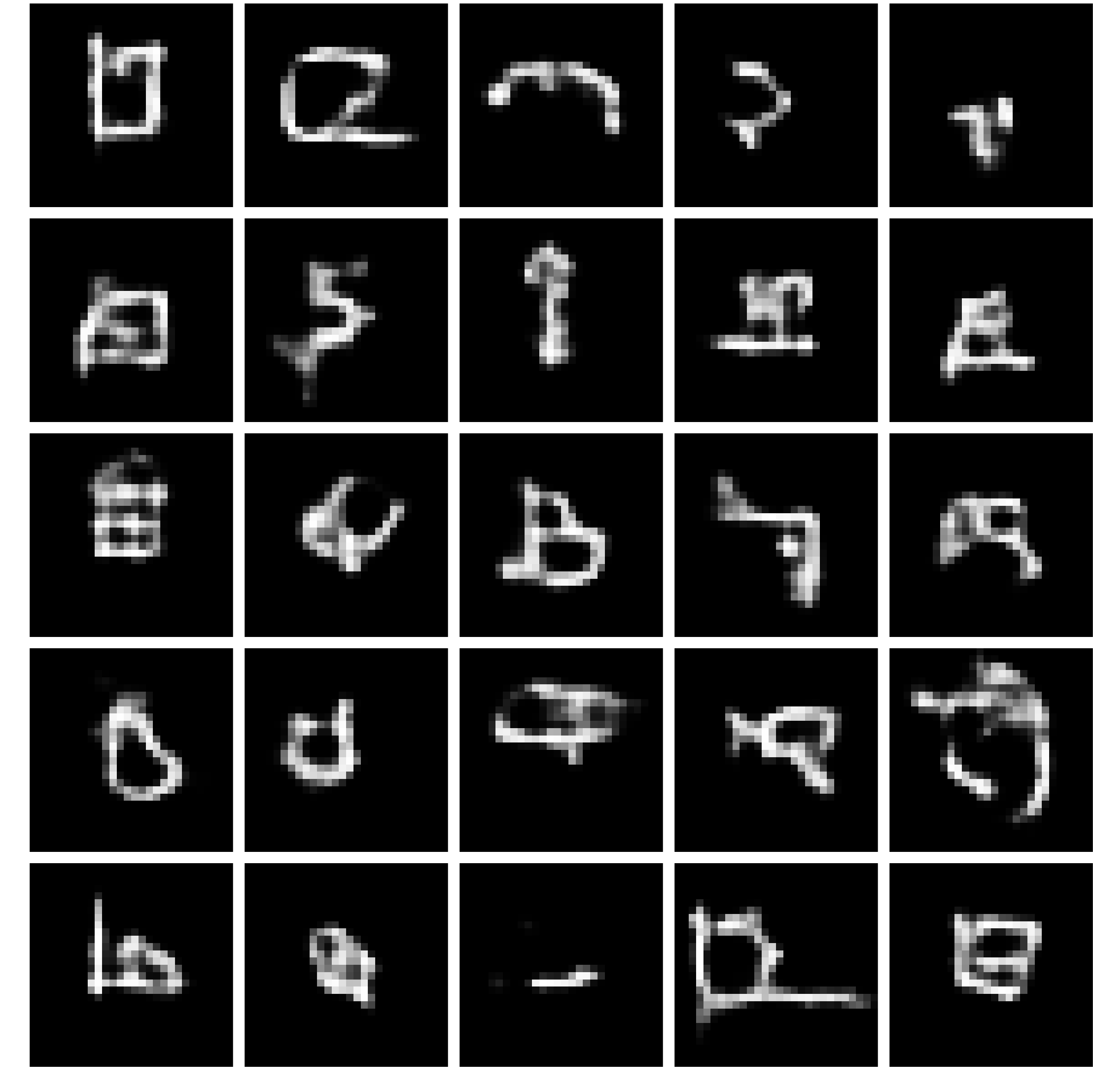}};
    \node[below=-.2 of mnist_good] (mnist_bad) {\includegraphics[width=1.3in]{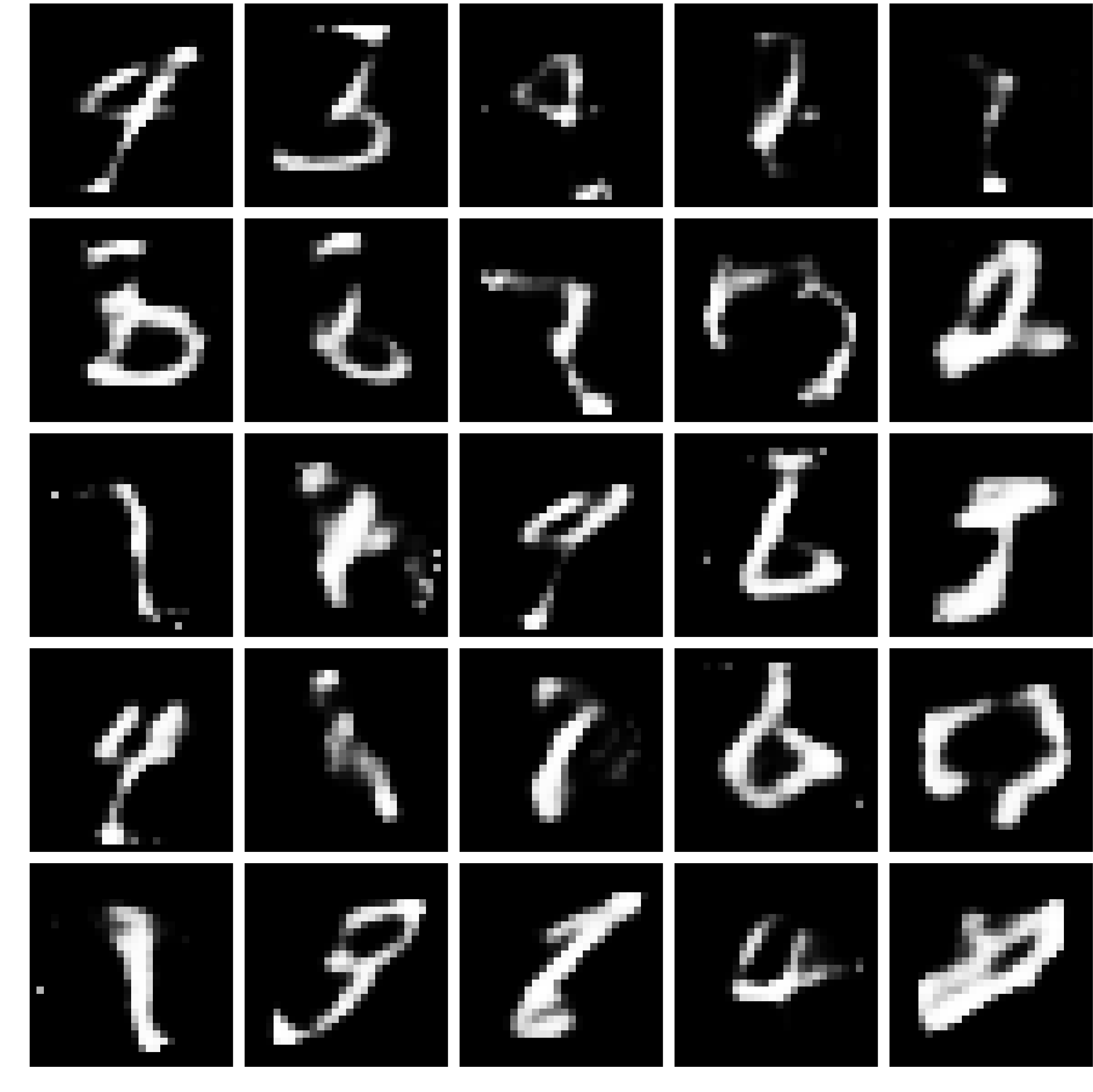} };
    \node[right=0. and 0. of mnist_bad] (omniglot_bad) {\includegraphics[width=1.3in]{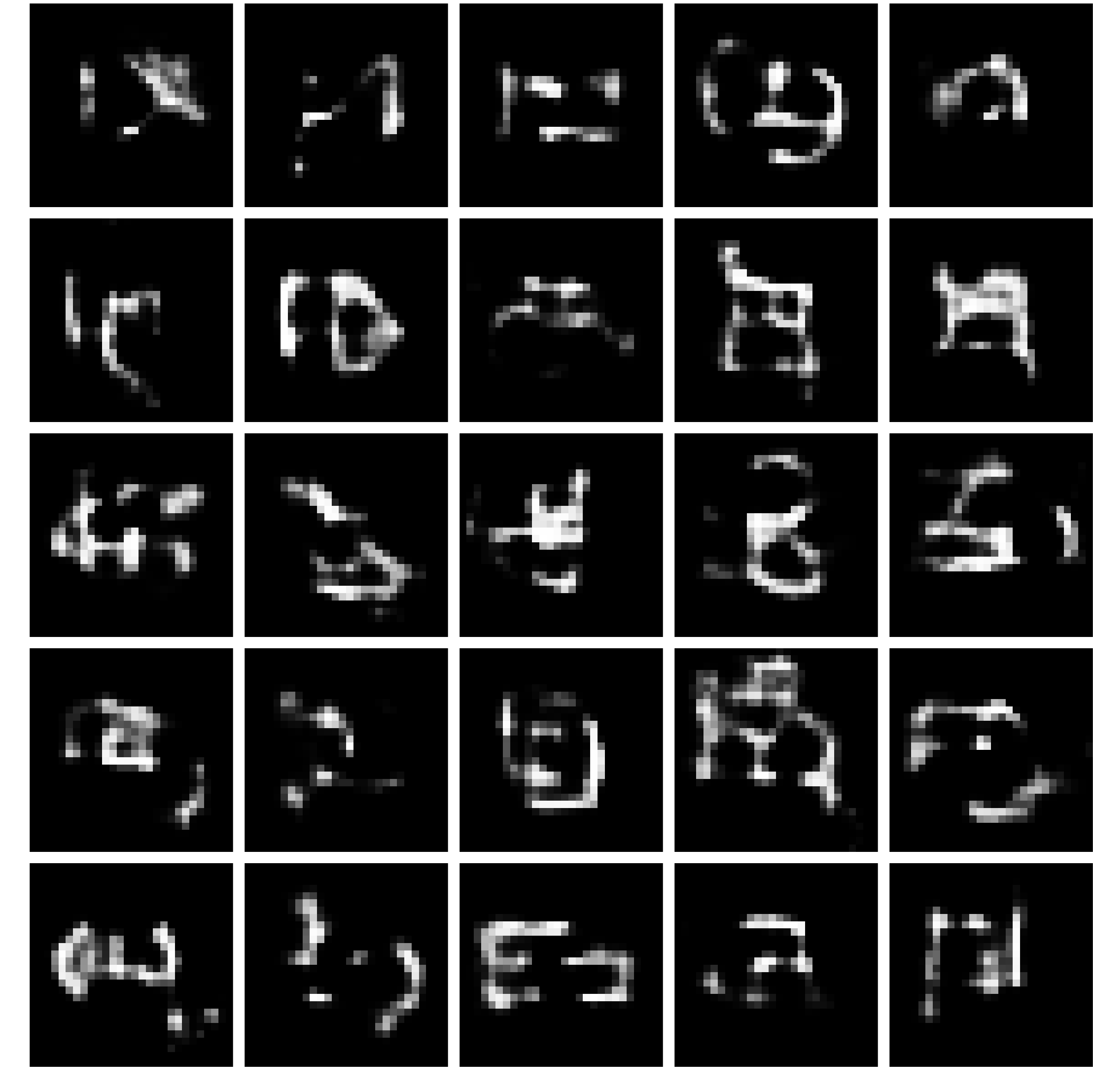}};
    \node[below=-0.1 of mnist_bad] {MNIST};
    \node[below=-0.1 of omniglot_bad] {Omniglot};
    \node[left=0.2 of mnist_good, rotate=90, anchor=center] {highest $a(\vz)\approx 1$};
    \node[left=0.2 of mnist_bad, rotate=90, anchor=center] {lowest $a(\vz) \leq 10^{-10}$};
    \end{tikzpicture}
    \vspace{-1.5ex}
    \caption{Ranked sample means \emph{from the proposal} of a VAE ($L=1$, MLP) with jointly trained resampled prior. We drew $10^4$ samples and show those with highest $a(\vz)$ (top) and lowest $a(\vz)$ (bottom).}
     \vskip-0.75ex
    \label{fig:samples_mlp}
\end{figure}
\paragraph{Samples.} For a qualitative analysis, we consider unconditional \emph{samples} from a VAE with a resampled prior. Our learned acceptance function assigns probability values $a(\vz)$ to all points in the latent space. This allows us to effectively \emph{rank} draws from the proposal and to compare 
particularly high- to particularly low-scoring ones.
We perform the same analysis twice, once for a resampled prior that is trained jointly with the VAE (\cref{fig:samples_mlp}) and once for a resampled prior that is trained post-hoc for a fixed VAE (\cref{fig:app:samples_mlp_posthoc}), both times with very similar results: Samples with $a(\vz)\approx 1$ are visually pleasing and show almost no artefacts, whereas samples with $a(\vz) \leq 10^{-10}$ look poor. When accepting/rejecting samples according to our pre-defined resampling scheme, these low-quality samples would most likely be rejected. Thus, our approach is able to 
\begin{inlinelist}
    \item automatically detect (rank) samples, and
    \item reject low-quality samples.
\end{inlinelist}
For more samples, see \cref{app:sec:samples_jointly_trained}.



%
\begin{table}[tb]
    \centering
    \rowcolors{1}{white}{lightgray}
    \small
    \begin{tabular}{l>{$}c<{$}>{$}c<{$}}
    \toprule
    \hiderowcolors \textsc{Model} & \text{NLL} & Z\\\showrowcolors\midrule
    \VAE{} ($L=1$) ($\mathcal{N}(0,1)$ prior) & 84.97 & 1\\
    \VAE{} (resampled $\mathcal{N}(0,1)$ prior) & \mathbf{83.04} & 0.015\\
    \VAE{} \text{\small resampled outputs} ($\mathcal{N}(0,1)$ prior) & 83.63 & 0.014\\
    \bottomrule
    \end{tabular}
     \vskip-1.25ex
    \caption{Test NLL on dynamic MNIST for a VAE with \emph{resampled outputs}.}
    \label{tab:output_vae}
\end{table}

\paragraph{Dimension-wise resampling.} Instead of resampling the entire vector $\vz$, a less wasteful sampling scheme would be to resample each dimension independently according to its own learned acceptance function $a_i(z_i)$ (or in blocks of several dimension). However, such a factorized approach did not improve over a standard (factorized) Normal prior in our case. We speculate that it might be more effective when the target density has strong factorial structure, as might be the case for $\beta$-VAEs \citep{Higgins_beta_VAE}; however, we leave this as future work.

\paragraph{VAE with 2D latent space.} In \cref{app:sec:2dlatent_space} we consider a resampled VAE with 2D latent space. While the NLL is not substantially different (\citep{ELBO_Surgery} show that the marginal KL gap is very small in this case), the resampled prior matches the aggregate posterior better and, interestingly, the jointly trained models give rise to a \emph{broader} aggregate posterior distribution, as the resampled prior upweights the tails of the proposal.


\subsection{Resampling \emph{discrete} outputs of a VAE}
\label{sec:output_vae}
So far, we have applied \Lars{} only to the continuous variables of the relatively low-dimensional latent space of a VAE. However, our approach is more general and can also be used to resample \emph{discrete and high-dimensional} random variables. We apply \Lars{} directly to the $784$-dimensional binarized (Bernoulli) outputs of a standard VAE in the pixel space. That is, we make the marginal likelihood richer by resampling it in the same way as the prior above:
 $\log p(\vx) = \log\frac{\pi(\vx)a(\vx)}{Z}$, 
where $\pi(\vx)$ is now the original marginal likelihood, $a(\vx)$ is the acceptance function in the pixel space, and $Z$ is the normalization constant. We can lower-bound the enriched marginal likelihood via its ELBO:
\begin{align}
    \Exp_{q(\vz|\vx)} \log\pi(\vx|\vz) - \kld*{q(\vz|\vx)\!}{\!p(\vz)} + \log\tfrac{a(\vx)}{Z}
\end{align}
In practice, we use the truncated sampling density with $T=100$ instead. $\vx$ corresponds to the observation under consideration, whereas $Z$ is estimated on decoded samples from the (regular) prior $p(\vz)$:
\begin{align}
    \textstyle Z = \int \pi(\vx| \vz) p(\vz) a(\vx) \d \vx \d \vz \approx \frac{1}{S} \sum_s^S a(\vx_{s}),
\end{align}
with $\vx_{s} \sim \pi(\vx|\vz_s)$  and  $\vz_s \sim p(\vz)$. We used the same MLP encoder/decoder as in our previous experiments, but increased the size of the $a(\vz)$ network to match the architecture of the encoder. 
As $\vx$ is a discrete random variable, we cannot easily compute the contribution of $\log Z$ to the gradient for the encoder and decoder parameters due to the inapplicability of the reparameterization trick. For simplicity, we ignore this contribution, which effectively leads to post-hoc fitting of $a(\vx)$, see \cref{app:sec:train_resampled_output} for a discussion.

Resampling the outputs of a VAE is slower than resampling the prior for several reasons:
\begin{inlinelist}
    \item we need to decode all prior samples $\vz_s$,
    \item a larger $a(\vx)$ has to act on a higher dimensional space, and
    \item estimation of $Z$ typically requires more samples to be accurate (we used $S=10^{11}$).
\end{inlinelist}
These are also the reasons why we apply \Lars{} in the lower-dimensional latent space to enrich the prior in all other experiments in this paper.

Still, \Lars{} works fairly well in this case. In terms of NLL, the VAE with resampled outputs outperforms the original VAE by over $1$ nat (\cref{tab:output_vae}) and is only about 0.6 nats worse than the VAE with resampled prior. The learned acceptance function can be used to rank the binary output samples, see \cref{fig:app:samples_output_vae_mnist}.

\section{Related Work}


\paragraph{Expressive prior distribution for VAEs.}
Recently, several methods of improving VAEs by making their priors more expressive have been proposed.
The VampPrior \citep{Jakub_Vamp_2018} is parameterized as a mixture of variational posteriors obtained by applying the encoder to a fixed number of learned pseudo-observations. While sampling from the VampPrior is fast, evaluating its density can be relatively expensive as the number of pseudo-observations used tends to be in the hundreds and the encoder needs to be applied on each one. A mixture of Gaussian prior \citep{nalisnick2016approximate,dilokthanakul2016deep} provides a simpler alternative to the VampPrior, but  is harder to optimize and does not perform as well \citep{Jakub_Vamp_2018}.
Autoregressive models parameterized with neural networks allow specifying very expressive priors \citep[e.g.][]{gregor2015draw,gulrajani2016pixelvae,MAF_2017} that support fast density evaluation but sampling from them is slow, as it is an inherently sequential process. While sampling in \Lars{} also appears  sequential, it can be easily parallelized by considering a large number of candidates in parallel. \citet{Chen_variational_lossy_vae} used autoregressive flows to define priors which are both fairly fast to evaluate and sample from.
Flow-based approaches require invertible transformations with fast-to-compute Jacobians; these requirements limit the power of any single transformation and thus necessitate chaining a number of them together. In contrast, \Lars{} places no restriction on the parameterization of the acceptance function and can work well even with fairly small neural networks. On the other hand, evaluating models trained with \Lars{} requires estimating the normalizing constant using a large number of MC samples.
As we showed above however, \Lars{} and flows can be combined to give rise to models with better objective value as well as higher sampling efficiency.
\paragraph{Rejection Sampling (RS).} 
Variational Rejection Sampling \citep{Grover_rejection_2018} uses RS to make the variational posterior closer to the exact posterior; as it relies on evaluating the target density, it is not applicable to our setting.
\citet{Naesseth_RS_reparameterization_gradients} derive reparameterization gradients for some not directly reparameterizable distributions by using RS to correct for sampling from a reparameterizable proposal instead of the distribution of interest. Similarly to VRS, this method requires being able to evaluate the target density.
    
\paragraph{Density Ratio Estimation (DRE).} DRE \citep{sugiyama2012density} 
estimates a ratio of densities by training a classifier to discriminate between samples from them. 
It has been used to estimate the KL term in the ELBO in order to train VAEs with implicit priors  \citep{Mescheder2017adversarial}, but the approach tends to overestimate the ELBO \citep{Rosca_Distribution_Matching_2018},  making model comparison difficult. The acceptance probability function learned by \Lars{} can be interpreted as 
estimating a (rescaled) density ratio between the aggregate posterior and the proposal, which is done end-to-end as a part of the generative process.
Note that we do not aim to estimate the density ratio as accurately as possible. Instead, we aim to learn a compact model for $a(\vz)$, so that the resulting $p(\vz)$ strikes a good balance between approximating $q(\vz)$ well and generalizing to new observations. Moreover, as the resampled prior has an explicit density, we can perform reliable model comparison by estimating the normalizing constant using a large number of samples.

\paragraph{Products of Experts.} The density induced by \Lars{} can be seen as a special instance of the product-of-experts (PoE,  \citep{hinton_product_experts}) architecture with two experts. However, while sampling from PoE models is generally difficult as it requires MCMC algorithms, our approach yields exact independent samples because of its close relationship to classical rejection sampling.

\section{Conclusion}

We introduced Learned Accept/Reject Sampling (\Lars{}), a powerful method for approximating a target density $q$ given a proposal density and samples from the target. \Lars{} uses a learned acceptance function to reweight the proposal, and can be applied to both continuous and discrete variables. We employed \Lars{} to define a resampled prior for VAEs and showed that it outperforms a standard Normal prior by a considerable margin on several architectures and datasets. It is competitive with other approaches that enrich the posterior or prior and can be efficiently combined with the ones that support efficient sampling, such as flows. \Lars{} can also be applied post hoc to already trained models and its learned acceptance function effectively ranks samples from the proposal by visual quality. 

We addressed two challenges of our approach, potentially low sampling efficiency and the requirement to estimate the normalization constant, and demonstrated that inference remains tractable and that a truncated resampling scheme provides an interpretable way to trade off sampling efficiency against approximation quality.

\Lars{} is not limited to variational inference and exploring its potential in other contexts is an interesting direction for future research. Its successful application to resampling the discrete outputs of a VAE is a first step in this direction. We also envision generalizing our approach to more than one acceptance function, e.g.~resampling dimensions independently, or using a different resampling scheme for the first $T$ and for the following attempts.

\FloatBarrier

\subsubsection*{Acknowledgements}
We thank Ulrich Paquet, Michael Figurnov, Guillaume Desjardins, M\'{e}lanie Rey, J\"{o}rg Bornschein, Hyunjik Kim, and George Tucker for their helpful comments and suggestions.


\subsubsection*{References}
\printbibliography[heading=none]
\clearpage

\end{refsection}
\onecolumn

\appendix
\renewcommand\thefigure{\thesection.\arabic{figure}} 
\setcounter{figure}{0}
\renewcommand\thetable{\thesection.\arabic{table}} 
\setcounter{table}{0}
\renewcommand\theequation{\thesection.\arabic{equation}} 
\setcounter{equation}{0}

\begin{refsection}

\begin{center}
    \LARGE{\textbf{Supplementary Material for ``Resampled Priors for Variational Autoencoders''}}
\end{center}

\section{Derivations}
\subsection{Resampling \emph{ad infinitum}}
\label{app:resampling_indefinite}
Here, we derive the density $p(\vz)$ for our rejection sampler that samples from a proposal distribution $\pi(\vz)$ and accepts a sample with probability given by $a(\vz)$. 

By definition, the probability of sampling $\vz$ from the proposal is given by $\pi(\vz)$, so the probability of sampling and then accepting a particular $\vz$ is given by the product $\pi(\vz) a(\vz)$. The probability of sampling and then rejecting  $\vz$ is given by the complementary probability $\pi(\vz) (1-a(\vz))$.

Thus, the expected probability of accepting a sample is given by $\int \pi(\vz) a(\vz) \d \vz$, whereas the expected probability of rejecting a sample is given by the complement $1 - \int \pi(\vz) a(\vz) \d \vz$.

With this setup we can now derive the density $p(\vz)$ for sampling and accepting a particular sample $\vz$. First, we consider the case, where we keep resampling until we finally accept a sample, if necessary indefinitely; that is, we derive the density $p_\infty(\vz)$ (\cref{eq:resampled_density_indefinite} in the main paper). It is given by the sum over an infinite series of events: We could sample $\vz$ in the first draw and accept it; or we could reject the sample in the first draw and subsequently sample and accept $\vz$; and so on. As all draws are independent, we can compute the sum as a geometric series:
\begin{align}
    p_\infty(\vz) & = \sum_{t=1}^\infty \Prob\left(\text{accept sample $\vz$ at step $t$|rejected previous $t-1$ samples}\right) \Prob\left(\text{reject $t-1$ samples}\right) \\
    & = \sum_{t=1}^\infty a(\vz) \pi(\vz) \left(1-\int \pi(\vz)a(\vz)\right)^{t-1} \d \vz\\
    & = \pi(\vz) a(\vz) \sum_{t=0}^\infty (1-Z)^t  \qquad\qquad Z = \int \pi(\vz)a(\vz)\,\d \vz\\
    & = \pi(\vz) a(\vz) \frac{1}{1-(1-Z)} \\
    & = \frac{\pi(\vz)a(\vz)}{Z}.
\end{align}
Note that going from the second to third line we have redefined the index of summation to obtain the standard formula for the sum of a geometric series: 
\begin{align}
    \sum_{n=0}^\infty x^n = \frac{1}{1-x} && \mathrm{for }\,\, |x| < 1
\end{align}
Thus, the log probability is given by:
\begin{align}
    \label{eq:app:resampled_indefinite}
    \log p(\vz) & = \log\pi(\vz) + \log a(\vz) - \log Z
\end{align}
As $0 \leq a(\vz) \leq 1$ we find that $0 \leq Z \leq 1$.

\paragraph{Average number of resampling steps until acceptance.}
We start by noting that the number of resampling steps performed until a sample is accepted follows the geometric distribution with success probability $Z$:
\begin{equation}
    \Prob(\text{resampling time}=t) = Z(1-Z)^{t-1}    
\end{equation}
This is due to the fact that each proposed candidate sample is accepted with probability $Z$, such decisions are independent, and the process continues until a sample is accepted. As the expected value of a geometric random variable is simply the reciprocal of the success probability, the expected number of resampling steps is $\left\langle t \right\rangle_{p_\infty} = 1/Z$. 




\subsection{Truncated Resampling}
\label{app:resampling_truncated}
Next, we consider a different resampling scheme that we refer to as \emph{truncated resampling} and derive its density. By truncation after the $T^\mathrm{th}$ step we mean that if we reject the first $T-1$ samples we accept the next ($T^\mathrm{th}$) sample with probability $1$ regardless of the value of $a(\vz)$.

Again, we can derive the density in closed form, this time by utilizing the formula for the truncated geometric series, 
\begin{align}
    \sum_{n=0}^N x^n = \frac{1-x^{N+1}}{1-x} && \mathrm{for }\,\, |x|< 1
\end{align}

\begin{align}
    p_T(\vz) & = \sum_{t=1}^T \Prob\left(\text{accept sample $\vz$ at step $t$|rejected previous $t-1$ samples}\right) \Prob\left(\text{reject $t-1$ samples}\right) \\
    & = \sum_{t=1}^{T-1}\Prob\left(\text{accept sample $\vz$ at step $t$|rejected previous $t-1$ samples}\right) \Prob\left(\text{reject $t-1$ samples}\right) \\
    & \qquad + \Prob\left(\text{accept sample $\vz$ at step $T$|rejected previous $T-1$ samples}\right) \Prob\left(\text{reject $T-1$ samples}\right) \\
    & = a(\vz)\pi(\vz) \sum_{t=0}^{T-2} (1-Z)^t + \pi(\vz) (1-Z)^{T-1} \\
    & = a(\vz) \pi(\vz) \frac{1-(1-Z)^{T-1}}{Z} + \pi(\vz) (1-Z)^{T-1}.
\end{align}
Again, note that we have shifted the index of summation going from the second to third line. Moreover, note that $a(\vz)$ does not occur in the second term in the third line as we accept the $T^\mathrm{th}$ sample with probability $1$ regardless of the actual value of $a(\vz)$.

The obtained probability density is normalized, that is, $\int p_T(\vz) \,\d \vz = 1$, because $\int \pi(\vz)a(\vz) \,\d \vz=Z$ and $\int  \pi(\vz) \,\d \vz=1$.
Thus, the log probability is given by:
\begin{align}
    \label{eq:app:resampled_truncated}
    \log p_T(\vz) & = \log\pi(\vz) + \log \left[ a(\vz) \frac{1-(1-Z)^{T-1}}{Z} + (1-Z)^{T-1} \right]
\end{align}

As discussed in the main paper, $p_T$ is a mixture of the untruncated density $p_\infty$ and the proposal density $\pi$. We can consider the two limiting cases of $T=1$ (accept every sample) and $T\rightarrow\infty$ (resample indefinitely, see \cref{app:resampling_indefinite}) and recover the expected results:
\begin{align}
    \log p_{T=1}(\vz) & = \log \pi(\vz) \\
    \lim_{T\rightarrow \infty} \log p_T(\vz) & = \log \frac{\pi(\vz) a(\vz)}{Z}.
\end{align}
That is, if we accept the first candidate sample ($T=1$), we obtain the proposal, whereas if we sample \emph{ad infinitum}, we converge to the result from above, \cref{eq:app:resampled_indefinite}.

\paragraph{Average number of resampling steps until acceptance.}
%
We can derive the average number of resampling steps for truncated resampling by using the result for indefinite resampling, $\left\langle t \right\rangle_{p_\infty} = 1/Z$, and noting that the geometric distribution is memoryless:
\begin{align}
    \left\langle t \right\rangle_{p_T} & = \left\langle t \right\rangle_{p_\infty} - \left\langle t \right\rangle_{p_{T..\infty}} + \left\langle t \right\rangle_{\text{accept $T$th sample}}\\
    & = \frac{1}{Z} - \left(T-1+\frac{1}{Z}\right) (1-Z)^{T-1} + T (1-Z)^{T-1} \\
    & = \frac{1-(1-Z)^{T}}{Z}
\end{align}
where we used the memoryless property of the geometric series for the second term. Also note that as the probability to accept the $T^\mathrm{th}$ sample given that we rejected all previous $T-1$ samples is $1$ instead of $Z$, the third term does not include a factor of $Z$.

From this result we can derive the following limits and special cases, which agree with the intuitive behaviour for truncated resampling; specifically, if we reject all samples ($a(\vz)\approx 0$, thus $Z\rightarrow 0^+$), the average number of resampling steps achieves the maximum possible value ($T$), whereas if we accept every sample, it goes down to $1$:
\begin{align}
    \left\langle t \right\rangle_{p_{T=1}} & = 1 \\
    \left\langle t \right\rangle_{p_{T=2}} & = 2-Z \leq 2 \\
    \left\langle t \right\rangle_{p_{T\rightarrow\infty}} & = \frac{1}{Z} \\
    \lim_{Z\rightarrow 0^+} \left\langle t \right\rangle_{p_T} &= T \\
    \lim_{Z\rightarrow 1^-} \left\langle t \right\rangle_{p_T} &= 1
\end{align}

\subsection{Illustration of standard/classical Rejection Sampling}
\begin{figure}[htb]
    \centering
    \includestandalone[mode=buildnew]{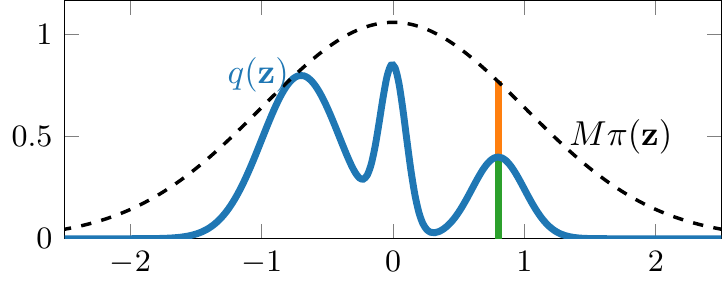}
    \caption[]{In rejection sampling we can draw samples from a complicated target distribution (\protect\tikz[baseline=(current bounding box.base),inner sep=0pt]{\draw[line width=1pt, solid, C1] (0pt,3pt) --(15pt,3pt);}) by sampling from a simpler proposal (\protect\tikz[baseline=(current bounding box.base),inner sep=0pt]{\draw[line width=1pt, dashed] (0pt,3pt) --(15pt,3pt);}) and accepting samples with probability $a(\vz) = \frac{q(\vz)}{M \pi(\vz)}$. The green and orange line are proportional to the relative accept and rejection probability, respectively.} 
    \label{fig:app:rejection_sampler}
\end{figure}

\subsection{Estimation of the normalization constant $Z$}
\label{app:sec:Z_MC}
Here we give details about the estimation of the normalization constant $Z$, which is needed both for updating the parameters of the resampled prior $p_T(\vz)$ as well as for evaluating trained models.

\paragraph{Model evaluation.} For evaluation of the trained model, we estimate $Z$ with a large number of Monte Carlo (MC) samples from the proposal:
\begin{equation}
\label{app:eq:Z_MC_estimate}
    Z=\frac{1}{S}\sum_s^S a(\vz_s) \qquad \text{with} \,\,\vz_s \sim \pi(\vz).
\end{equation}
In practice, we use $S=10^{10}$ samples, and evaluating $Z$ takes only several minutes on a GPU. The fast evaluation time is due to the relatively low dimensionality of the latent space, so drawing the samples and evaluating their acceptance function values $a(\vz)$ is fast and can be parallelized by using large batch sizes ($10^5$). For symmetric proposals, we also use antithetic sampling, that is, if we draw $\vz$, then we also include $-\vz$, which also corresponds to a valid/exact sample from the proposal. Antithetic sampling can reduce variance (see e.g. Section 9.3 of \citep{kroese2013handbook}).

\begin{figure}[htb]
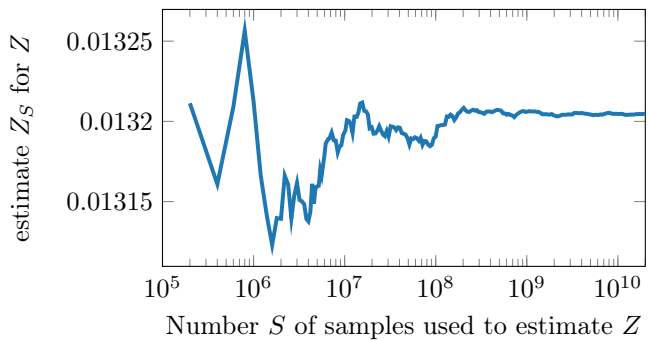

    \centering
    \includestandalone{figures/mnist_mlp_Z_trace}
    \caption{Estimation of $Z$ by MC sampling from the proposal. For less than $10^7$ samples the estimate is quite noisy but it stabilizes after about $10^9$ samples.}
    \label{fig:app:Z_trace}
\end{figure}

In \cref{fig:app:Z_trace} we show how the Monte Carlo estimates of the normalization $Z$ typically evolves with the number of MC samples. We plot the estimate of $Z$ as a function of the number of samples used to estimate it, $S$. That is
\begin{equation}
    Z_S=\frac{1}{S}\sum_i^S a(\vz_i) \qquad \text{with} \,\,\vz_s \sim \pi(\vz)
\end{equation}

\paragraph{Model training.} During training we would like to draw as few samples from the proposal as possible in order not to slow down training. In principle, we could use the same Monte Carlo estimate as introduced above (\cref{app:eq:Z_MC_estimate}). However, when using too few samples, the estimate for $Z$ can be very noisy, see \cref{fig:app:Z_trace}, and the values for $\log Z$ are substantially biased. In practice, we use two techniques to deal with these and related issues:
\begin{inlinelist}
    \item smoothing the $Z$ estimate by using exponentially moving averages in the forward pass, and
    \item including the sample from $q(\vz|\vx)$, on which we evaluate the resampled prior to compute the KL in the objective function, in the estimator. 
\end{inlinelist}
We now describe these techniques in more detail.

During training we evaluate $\kld*{q(\vz|\vx)}{p(\vz)}$ in a stochastic fashion, that is, we evaluate $\log q(\vz_r|\vx_r) - \log p(\vz_r)$ for each data point $\vx_r$ in the minibatch and average the results. Here, $\vz_r \sim q(\vz|\vx_r)$ is a sample from the variational posterior that corresponds to data point $\vx_r$. Thus, we need to evaluate the resampled prior $p(\vz_r)$ on all samples $\vz_r$ in that batch. The mean KL contribution of the prior terms is then given by:
\begin{align}
    \frac{1}{R} \sum_{r=1}^R \log p_T(\vz_r) = \frac{1}{R}\sum_r^R (\log\pi(\vz_r) + \log a(\vz_r)) - \log Z
    \label{app:eq:logp}
\end{align}
where, so far, $Z$ is shared between all data points in the minibatch. To reduce the variability of $Z$ as well as the bias of $\log Z$, we smooth $Z$ using exponentially moving averages. That is, given the previous moving average from iteration $i$, $\left\langle Z\right\rangle_i$, as well as the current estimate of $Z$ computed from $S$ MC samples, $Z_i$, we obtain the new moving average as:
\begin{align}
    \left\langle Z\right\rangle_{i+1} = (1-\epsilon) \left\langle Z\right\rangle_{i} + \epsilon Z_i 
\end{align}
where $\epsilon$ controls how much $Z$ is smoothed. In practice, we use $\epsilon = 0.1$ or $\epsilon=0.01$. However, we can only use the moving average in the forward pass, as it is prohibitively expensive (and unnecessary) to backpropagate through all the terms in the sum. Therefore, we only want to backpropagate through the last term ($Z_i$), which needs to be rescaled to account for $\epsilon$. Before we show how we do this, we introduce our second way to reduce variance: including the sample from $q(\vz|\vx_r)$ in the estimator. %

For this, we introduce a datapoint dependent estimate $Z_r$, and replace the $\log Z$ term in \cref{app:eq:logp} by:
\begin{align}
    \log Z & \rightarrow \frac{1}{R}\sum_r^R \log Z_r \label{app:eq:logZ_r} \\
    Z_r & = \frac{1}{S+1}\left[\sum_s^S a(\vz_s) + \frac{\pi(\vz_r)}{q(\vz_r|\vx_r)}a(\vz_r)\right] && \vz_s \sim \pi(\vz); \quad \vz_r \sim q(\vz|\vx_r) \\
    & = \frac{1}{S+1}\left[ S Z_S + \frac{\pi(\vz_r)}{q(\vz_r|\vx_r)}a(\vz_r)\right]
\end{align}
where the sample $\vz_r$ from the variational posterior is reweighted by the ratio of prior and variational posterior, similar to \citep{botev17acomplementary}. Note that we do not want to backpropagate through this fraction, as this would alter the gradients for the encoder and the proposal. We are now in a position to write down the algorithm that computes the different $Z_r$ values to use in the objective, see \cref{app:alg:Z}.

\IncMargin{1em}
\begin{algorithm}
\DontPrintSemicolon
\SetAlgoLined
    \SetKwInOut{Input}{input} \SetKwInOut{Output}{output}
    \Input{Minibatch of samples $\{\vz_r\}_r^R$ from $\{q(\vz|\vx_r)\}_r^R$; $S$ samples $\{\vz_s\}_s^S$ from $\pi$; previous moving average $\left\langle Z\right\rangle_i$}
    \Output{Updated moving average $\left\langle Z\right\rangle_{i+1}$; $\log Z$ estimate for current minibatch $\{\vx_r\}_r^R$}
    \nosemic $Z_S \leftarrow \frac{1}{S} \sum_s^S a(\vz_s)$\;
    \For{$r \leftarrow 1$ \KwTo $R$}{
        \nosemic $Z_{r,\mathrm{curr}} \leftarrow \frac{1}{S+1}\left[S Z_S + {}\right.$ \texttt{stop\_grad}$\left.\left(\frac{\pi(\vz_r)}{q(\vz_r)}\right)a(\vz_r)\right]$ \;
        \nosemic $Z_{r,\mathrm{smooth}} \leftarrow (1-\epsilon) \left\langle Z\right\rangle_i + \epsilon Z_{r,\mathrm{curr}}$\;
        \nosemic $Z_r \leftarrow Z_{r,\mathrm{curr}} + {}$\texttt{stop\_grad}$(Z_{r,\mathrm{smooth}} - Z_{r,\mathrm{curr}})$\;
    }
    \nosemic $\left\langle Z\right\rangle_{i+1} \leftarrow \frac{1}{R}\sum_r^R $ \texttt{stop\_grad}$\left(Z_{r,\mathrm{smooth}}\right)$\;
    \nosemic $\log Z \leftarrow \frac{1}{R}\sum_r^R \log Z_r$
    \caption{Estimation of $\log Z$ (\protect\cref{app:eq:logZ_r}) in the objective during training\label{app:alg:Z}}
    \label{alg:logZ}
\end{algorithm}
\DecMargin{1em}
Note that in the forward pass, line 5 evaluates to $Z_{r,\mathrm{smooth}}$, whereas in the backward direction (when computing gradients), it evaluates to $Z_{r,\mathrm{curr}}$. Thus, we use the smoothed version in the forward pass and the current estimate for the gradients.

While \cref{app:alg:Z} estimates $\log Z$ used in the density of the indefinite/untruncated resampling scheme, we can easily adapt it to compute the truncated density as well, as we compute the individual $Z_{r}$, which can be used instead.

\clearpage

\section{Experimental Details}

We implemented all of our models using \texttt{TensorFlow} \citep{tensorflow2015-whitepaper} and \texttt{TensorFlow Probability} \citep{tensorflow-distributions}.

\subsection{Datasets}
\label{app:datasets}

The MNIST dataset \citep{LeNet1998} contains 50,000 training, 10,000 validation and 10,000 test images of the digits 0 to 9. Each image has a size of $28\times 28$ pixels. We use both a dynamically binarized version of this dataset as well as the statically binarized version introduced by \citet{NADE_2011_Larochelle}. 

Omniglot \citep{Lake_Omniglot_2015} contains 1,623 handwritten characters from 50 different alphabets, with differently drawn examples per character. The images are split into 24,345 training and 8,070 test images. Following \citet{Jakub_Vamp_2018} we take 1,345 training images as validation set. Each image has a size of $28 \times 28$ pixels and we applied dynamic binarization to them.

FashionMNIST \citep{FashionMNIST} is a recently proposed plug-in replacement for MNIST with 60,000 train/validation and 10,000 test images split across 10 classes. Each image has size of $28 \times 28$ pixels and we applied dynamic binarization to them.

\subsection{Network architectures}
\label{app:architectures}

In general, we decided to use simple standard architectures. We observed that more complicated networks can overfit quite drastically, especially on staticMNIST, which is not dynamically binarized.

\paragraph{Notation.} For all networks, we specify the input size and then consecutively the output sizes of the individual layers separated by a ``${}-{}$''. Potentially, outputs are reshaped to convert between convolutional layers (abbreviated by ``CNN'') and fully connected layers (abbreviated by ``MLP''). When we nest networks, e.g. by writing $p(\vx | \vz_1, \vz_2) =\MLP{[\MLP{d_\vz-300}, \MLP{d_\vz-300}] - 300 - 28\times 28}$, we mean that first the two inputs/conditioning variables, $\vz_1$ and $\vz_2$ in this case, are transformed by neural networks, here an $\MLP{d_\vz-300}$, and their concatenated outputs (indicated by $[ \cdot , \cdot ]$) are subsequently used as an input to another network, in this case with hidden layer of $300$ units and reshaped output $28\times 28$.

\paragraph{\VAE{} ($L=1$).}

For the single latent layer VAE, we used an MLP with two hidden layers of 300 units each for both the encoder and the decoder. The latent space was chosen to be $d_\vz=50$ dimensional and the nonlinearity for the hidden layers was \texttt{tanh}; the likelihood was chosen to be Bernoulli. The encoder parameterizes the mean $\mu$ and the log standard deviation $\log \sigma$ of the diagonal Normal variational posterior.
\begin{align*}
    q(\vz | \vx) & = \mathcal{N}(\vz; \mu_\vz(\vx), \sigma_\vz(\vx)) \\
    p(\vx | \vz) & = \mathrm{Bernoulli}(\vx; \mu_\vx(\vz))
\end{align*}
which are given by:
\begin{align*}
    \mu_\vz(\vx) & = \MLP{28\times 28 - 300 - 300 - d_\vz} \\
    \log\sigma_\vz(\vx) & = \MLP{28\times 28 - 300 - 300 - d_\vz} \\
    \mu_\vx(\vz) & = \MLP{d_\vz - 300 - 300 - 28\times 28}
\end{align*}
where the networks for $\mu_z(\vx)$ and $\log\sigma_z(\vx)$ are shared up to the last layer. In the following, we will abbreviate this as follows:
\begin{align*}
    q(\vz | \vx) & = \MLP{28\times 28 - 300 - 300 - d_\vz} \\
    p(\vx | \vz) & = \MLP{d_\vz - 300 - 300 - 28\times 28}
\end{align*}

\paragraph{\HVAE{} ($L=2$).}

For the hierarchical VAE with two latent layers, we used two different architectures, one based on MLPs and the other on convolutional layers. Both were inspired by the architectural choices of \citet{Jakub_Vamp_2018}, however, we chose to use simpler models than them with fewer layers and regular nonlinearities instead of gated units to avoid overfitting.

The MLP was structured very similarly to the single layer case:
\begin{align*}
    q(\vz_2 | \vx) & = \MLP{28\times 28 - 300 - 300 - d_\vz} \\
    q(\vz_1 | \vx, \vz_2) & = \MLP{[\MLP{28\times 28 - 300}, \MLP{d_\vz - 300}] - 300 - d_\vz}\\
    p(\vz_1 | \vz_2) &= \MLP{d_\vz - 300 - 300 - d_\vz}\\
    p(\vx | \vz_1, \vz_2) & = \MLP{[\MLP{d_\vz-300}, \MLP{d_\vz-300}] - 300 - 28\times 28}
\end{align*}
We again use \texttt{tanh} nonlinearities in hidden layers of the MLP and a Bernoulli likelihood model.

\paragraph{\ConvHVAE{} ($L=2$).}
The overall model structure is similar to the \HVAE{} but instead of MLPs we use CNNs with strided convolutions if necessary. To avoid imbalanced up/down-sampling we chose a kernel size of $4$ which works well with strided up/down-convolutions. As nonlinearities we used \texttt{ReLU} activations after convolutional layers and \texttt{tanh} nonlinearities after MLP layers. 
\begin{align*}
    q(\vz_2 | \vx) & = \MLP{\CNN{28\times 28\times 1 - 14\times 14 \times 32 - 7 \times 7 \times 32 - 4 \times 4 \times 32}-d_\vz} \\
    q(\vz_1 | \vx, \vz_2) & = \MLP{[\CNN{28\times 28\times 1 - 14\times 14 \times 32 - 7 \times 7 \times 32 - 4 \times 4 \times 32}, \MLP{d_\vz - 4 \times 4 \times 32}] - 300 - d_\vz}\\
    p(\vz_1 | \vz_2) &= \MLP{d_\vz - 300 - 300 - d_\vz}\\
   p(\vx | \vz_1, \vz_2) & = \CNN{\MLP{[\MLP{d_\vz - 300}, \MLP{d_\vz - 300}] - 4\times 4\times 32} - 7\times 7\times 32 - 14\times 14\times 32 - 32\times 32 \times 1  }
\end{align*}

\paragraph{Acceptance function $a(\vz)$.}
For $a(\vz)$ we use a very simple MLP architecture,
\begin{align*}
    a(\vz) & = \MLP{d_\vz - 100 - 100 - 1},
\end{align*}
again with \texttt{tanh} nonlinearities in the hidden layers and a \texttt{logistic} nonlinearity on the output to ensure that the final value is in the range $[0,1]$.

\paragraph{RealNVP.}
\label{all:realnvp}
For the RealNVP prior/proposal we employed the reference implementation from \texttt{TensorFlow Probability} \citep{tensorflow-distributions}. We used a stack of $4$ RealNVPs with two hidden MLP layers of $100$ units each and performed reordering permutations in-between individual RealNVPs, as the RealNVP only transforms half of the variables.

\paragraph{Masked Autoregressive Flows.}
For the Masked Autoregressive Flows \citep{MAF_2017} (MAF) prior/proposal we use a stack of 5 MAFs with \MLP{100-100} each and again use the reference implementation from \texttt{TensorFlow Probability} \citep{tensorflow-distributions}. Random permutations are employed between individual MAF blocks \citep{MAF_2017}.

\subsection{Training procedure for a VAE with resampled prior}
\label{app:sec:vae_training}

\Lars{} constitutes another method to specify rich priors, which are parameterized through the acceptance function $a_\lambda(\vz)$; we use subscript $\lambda$ in this section to highlight the learnable parameters of the acceptance function. The acceptance function only modifies the prior of the model with the encoder and decoder remaining unchanged. Thus, the only part of the training objective, which changes compared to training of a normal VAE, is the evaluation of prior contribution to the KL term, $\log p(\vz)$. This log density is evaluated on samples from the encoder distribution $q_\varphi(\vz|\vx)$. Both for truncated and untruncated sampling, evaluation of $\log p(\vz)$ entails evaluating 
\begin{inlinelist}
    \item the proposal density $\pi(\vz)$
    \item the acceptance function $a_\lambda(\vz)$
    \item the normalizer $Z$
\end{inlinelist}
and their logarithms. Evaluation of the proposal and acceptance function is straight forward and evaluation of the normalizer and its logarithm are detailed in \cref{alg:logZ}. 

As discussed in the main paper, we never need to perform accept/reject sampling during training but only need to evaluate the prior density. To update the normalizer during training, we require a modest number of samples from the proposal; however, we never actually perform the accept/reject step and never need to decode these samples into image space using the decoder model. 

The full training procedure for a VAE as well as the changes necessary due the resampled prior are detailed in \cref{alg:vae_training}. We detail the pseudo code for resampled priors with untruncated/indefinite resampling but an extension to the truncated case is straight forward.

\IncMargin{1em}
\begin{algorithm}
\DontPrintSemicolon
\SetAlgoLined
    \SetKwInOut{Input}{input}\SetKwInOut{Output}{output}
    \Input{dataset of images $\mathcal{D_\text{train}}=\{\vx_n\}_n^N$}
    \Output{trained model parameters $\varphi, \theta$ (encoder/decoder) \textcolor{red}{and $\lambda, \log Z$ (resampled prior)}}
    initialize parameters of encoder $\varphi$ and decoder $\theta$ \;
    \textcolor{red}{initialize parameters of the resampled prior $\lambda$ and the moving average $\left\langle Z\right\rangle$}\;
    \For{$it \leftarrow 1$ \KwTo $N_{it}$}{
        sample a minibatch of data $\{\vx_r\}_r^R$\;
        sample latents $\{\vz_r\}_r^R$ from the encoder distribution $\{q_\varphi(\vz|\vx_r)\}_r^R$ (using reparameterization)\;
        evaluate the reconstruction term: $\text{recon} \leftarrow \frac{1}{R}\sum_r^R \log p_\theta(\vx|\vz_r)$\;
        \textcolor{red}{update moving average $\left\langle Z \right\rangle$ and compute $\log Z$ using \cref{alg:logZ}}\;
        evaluate the KL term: $\text{kl} \leftarrow \frac{1}{R}\sum_r^R \left[\log q_\varphi(\vz_r|\vx_r) - \log \pi(\vz_r) \textcolor{red}{{}-\log a_\lambda(\vz_r)}\right] \textcolor{red}{{} + \log Z}$\;
        evaluate the objective: $\text{recon}-\text{kl}$\;
        backpropagate gradients into all parameters $\varphi, \theta$ \textcolor{red}{and $\lambda$}\;
    }
    compute final normalizer for evaluation: $Z \leftarrow \frac{1}{S_\text{eval}} \sum_{s}^{S\text{eval}}a(\vz_s)$ with $\vz_s \sim \pi(\vz)$\;
    \caption{Pseudo code for training of a VAE \textcolor{red}{with resampled prior (for untruncated resampling)}}
    \label{alg:vae_training}
\end{algorithm}
\DecMargin{1em}

\FloatBarrier

\subsection{Training a VAE with resampled outputs}
\label{app:sec:train_resampled_output}
Here, we elaborate on the training procedure for a VAE with \emph{resampled discrete outputs} presented in \cref{sec:output_vae}. In particular, we explain why we essentially perform post-hoc fitting for the acceptance function $a(\vx)$. 

As stated in the main text, the ELBO for the resampled objective is given by:
\begin{align}
    \Exp_{q_\varphi(\vz|\vx)} \log\pi_\theta(\vx|\vz) - \kld*{q_\varphi(\vz|\vx)\!}{\!p(\vz)} + \log\tfrac{a_\lambda(\vx)}{Z},
    \label{app:eq:elbo_resampled}
\end{align}
where $\vx$ corresponds to the observation under consideration and we have reintroduced the subscripts $\varphi, \theta, \lambda$ to indicate the learnable parameters of the encoder, decoder, and acceptance function, respectively. The normalizer $Z$ is estimated on decoded samples from the (regular) prior $p(\vz)$:
\begin{align}
    \textstyle Z = \int \pi_\theta(\vx| \vz) p(\vz) a_\lambda(\vx) \d \vx \d \vz = \Exp_{\pi_\theta(\vx| \vz) p(\vz)} [a_\lambda(\vx)]  \approx \frac{1}{S} \sum_s^S a_\lambda(\vx_{s}),
    \label{app:eq:resampled_output_Z}
\end{align}
with $\vx_{s} \sim \pi(\vx|\vz_s)$  and  $\vz_s \sim p(\vz)$. Note that because the images in the MNIST dataset are binary, both the observations $\vx$ as well as the model samples $\vx_s$ are discrete variables. This does not make learning the parameters $\lambda$ of the acceptance function $a_\lambda(\vx)$ more difficult than in the continuous case, as it does not involve propagating gradients \emph{through} $\vx$.

However, the situation is different for the parameters $\theta$ of the decoder, as the distribution of the model samples $\vx_s$ and, thus, our estimate of $Z$ depends on them. Therefore we need to compute the gradient of the sample-based estimate $Z$ w.r.t.~the decoder parameters $\theta$; this would be be easy to do by using the reparameterization trick if $\vx_s$ were continuous. Unfortunately, in our case, the samples $\vx_s$ are discrete and thus non-reparameterizible, which means that we have to use a more general gradient estimator such as REINFORCE \citep{Williams1992_reinforce}. Unfortunately, estimators of this type usually have much higher variance than reparameterization gradients; so for simplicity we choose to ignore this contribution to the gradients of the decoder parameters. Computationally, this amounts to estimating $Z$ using 
\begin{equation}
    Z \approx \frac{1}{S}\sum_s^S a_\lambda(\texttt{stop\_grad}(\vx_s)).
\end{equation}
Thus, the encoder and decoder parameters are trained effectively by optimizing the first two terms of \cref{app:eq:elbo_resampled}, whereas the parameters of the acceptance function are trained by optimizing the last term of \cref{app:eq:elbo_resampled}. This is essentially equivalent to the post-hoc setting considered in \cref{tab:posthoc_mnist_omniglot} because the acceptance function has no effect on training the rest of the model.

\clearpage 
\section{Additional results}

\subsection{Illustrative examples in 2D}
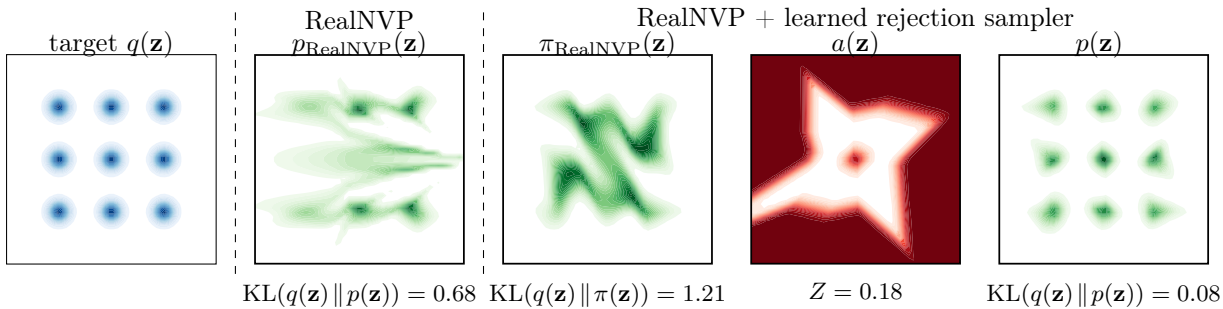
\begin{figure*}[htb]
    \centering
    \begin{tikzpicture}
    \node (target) {\includegraphics[height=2.8cm]{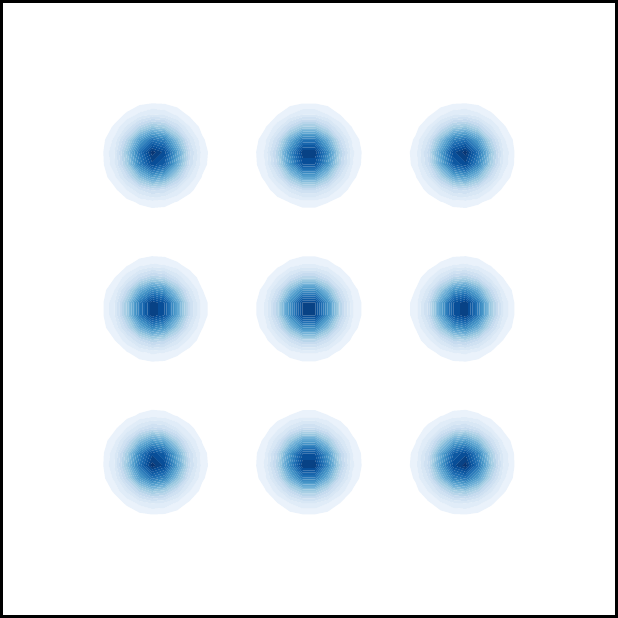}};
    \node[right=0.25cm of target] (p_high) {\includegraphics[height=2.8cm]{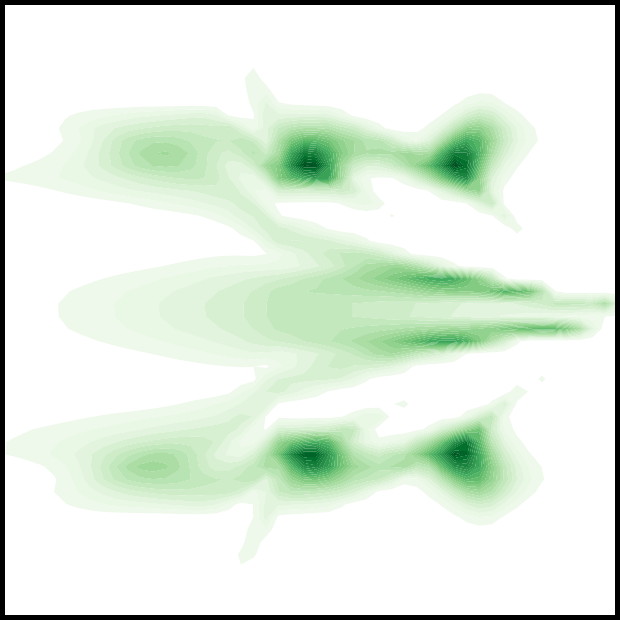}};
    \node[right=0.25cm of p_high] (a_high) {\includegraphics[height=2.8cm]{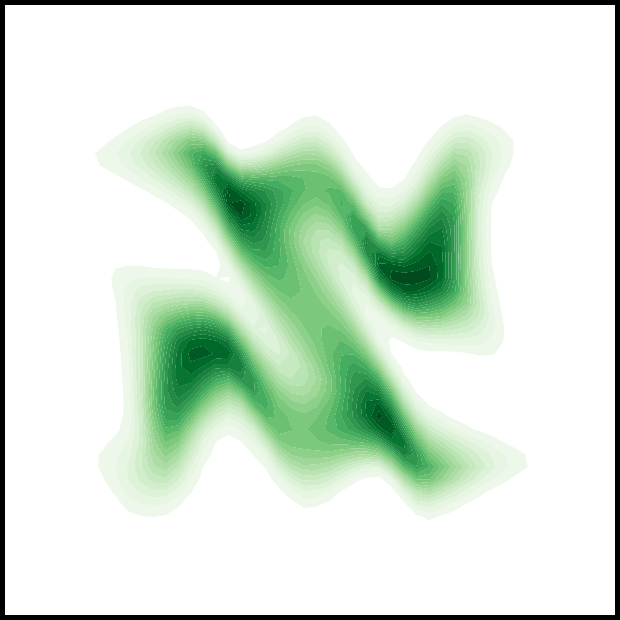}};
    \node[right=0.25cm of a_high] (p_low) {\includegraphics[height=2.8cm]{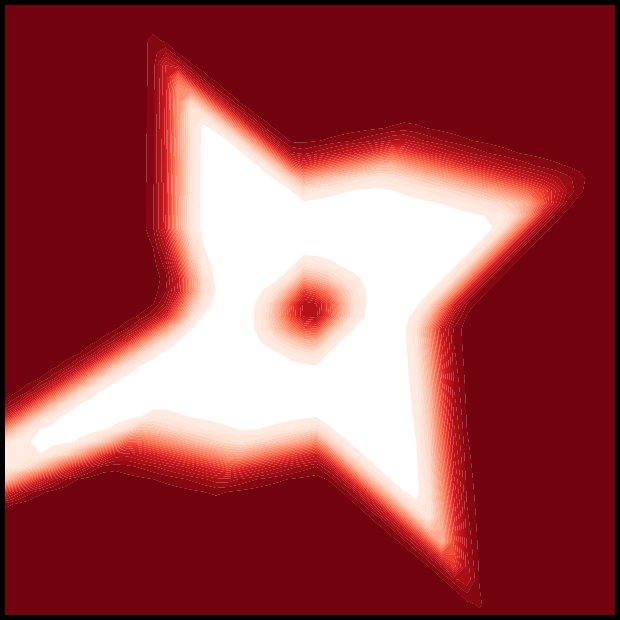}};
    \node[right=0.25cm of p_low] (a_low) {\includegraphics[height=2.8cm]{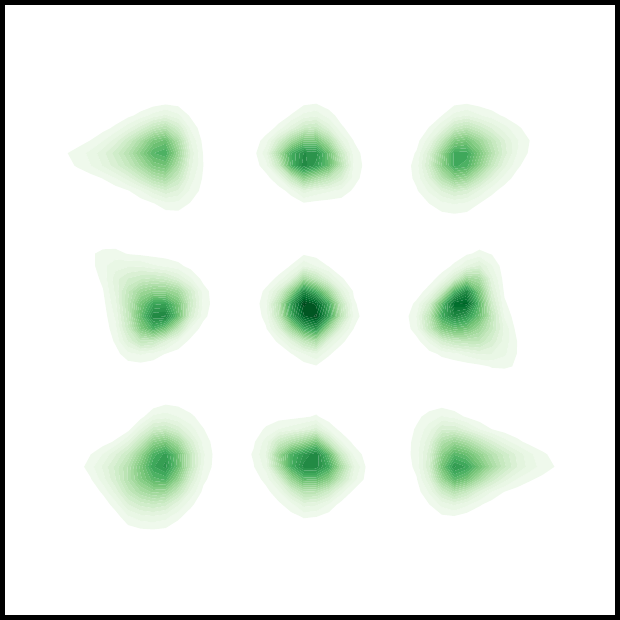}};
    \draw[dashed] ($(p_high.south)!0.5!(target.south)$) -- ($(p_high.north)!0.5!(target.north) + (0, 0.5)$);
    \draw[dashed] ($(p_high.south)!0.5!(a_high.south)$) -- ($(p_high.north)!0.5!(a_high.north) + (0, 0.5)$);
    \node[anchor=center] at (target.north) {target $q(\vz)$\strut};
    \node[anchor=center] (p_nvp_text) at (p_high.north) {$p_\mathrm{RealNVP}(\vz)$\strut};
    \node[anchor=north] at (p_high.south) {\small $\kld*{q(\vz)\!}{\!p(\vz)}=0.68$};
    \node[anchor=center] at (a_high.north) {$\pi_\mathrm{RealNVP}(\vz)$\strut};
    \node[anchor=north] at (a_high.south) {\small $\kld*{q(\vz)\!}{\!\pi(\vz)}=1.21$};
    \node[anchor=center] (a_text) at (p_low.north) {$a(\vz)$\strut};
    \node[anchor=north] at (p_low.south) {\small $Z=0.18$};
    \node[anchor=center] at (a_low.north) {$p(\vz)$ \strut};
    \node[anchor=north] at (a_low.south) {\small $\kld*{q(\vz)\!}{\!p(\vz)}=0.08$};
    \node[above, anchor=center] at (p_nvp_text.north) {RealNVP};
    \node[above, anchor=center] at (a_text.north) {RealNVP + learned rejection sampler};
    \end{tikzpicture}
    \caption[]{Training with a RealNVP proposal. The target is approximated either by a RealNVP alone (left) or a RealNVP in combination with a learned rejection sampler (right). Darker values correspond to higher numbers.}
    \label{fig:app:toy_realnvp}
\end{figure*}

Here, we show larger versions of the images in \cref{fig:toy} and also show the learned density of a RealNVP when used in isolation as a prior, $p_\mathrm{RealNVP}$, see \cref{fig:app:toy_realnvp}. We find that the RealNVP by itself is not able to isolate all the modes, as has already been observed by \citet{NAF_2018}, who observe a similar shortcoming for Masked Autoregressive Flows (MAFs).

\begin{figure*}[htb]
    \centering
    \begin{tikzpicture}
    \node (target) {\includegraphics[height=2.8cm]{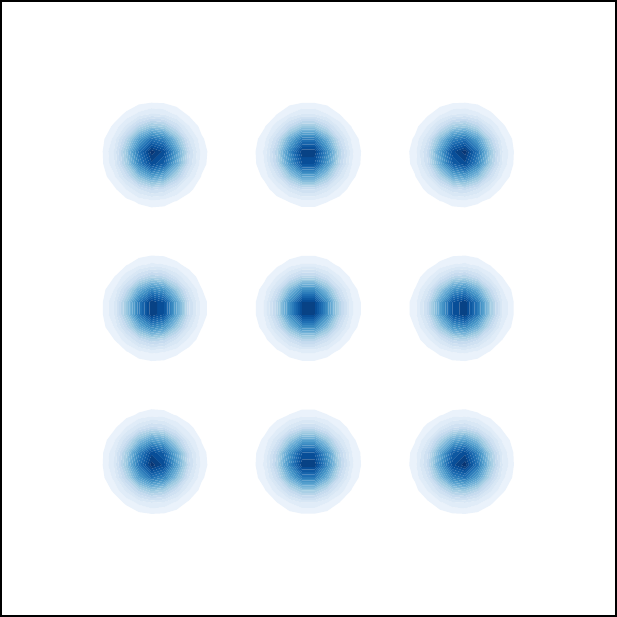}};
    \node[right=0.25cm of target] (p_high) {\includegraphics[height=2.8cm]{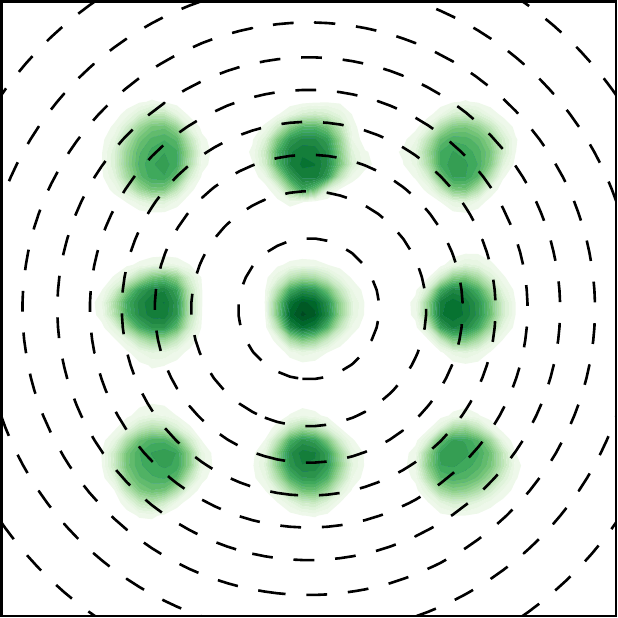}};
    \node[right=0.25cm of p_high] (a_high) {\includegraphics[height=2.8cm]{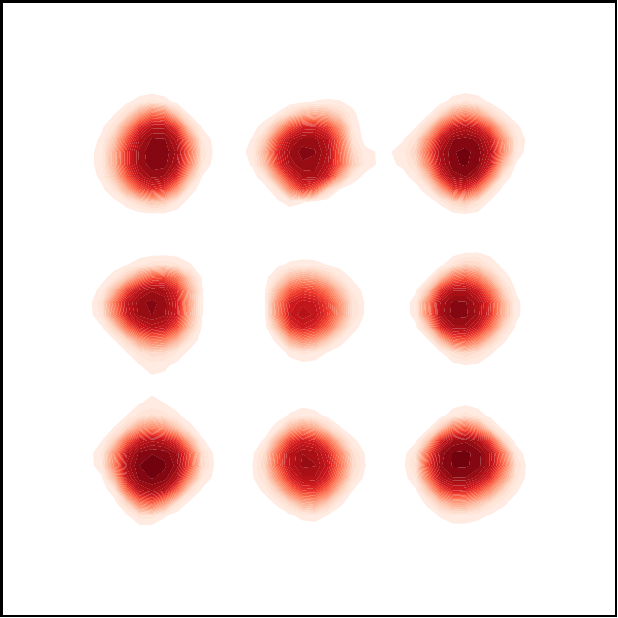}};
    \node[right=0.25cm of a_high] (p_low) {\includegraphics[height=2.8cm]{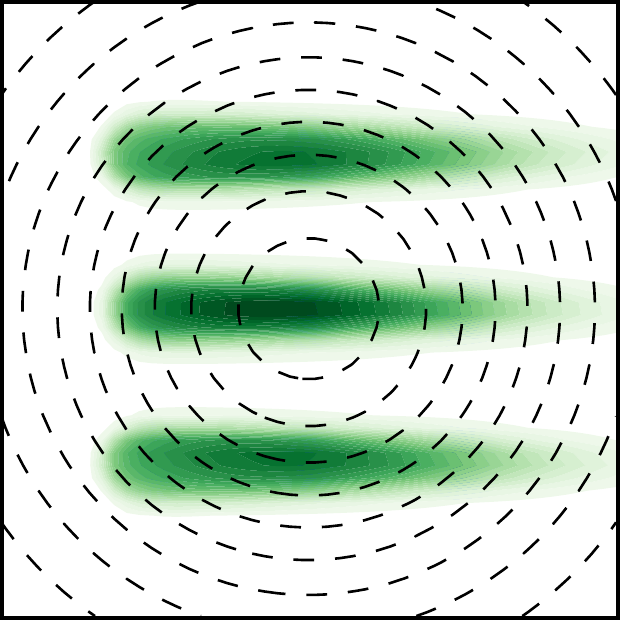}};
    \node[right=0.25cm of p_low] (a_low) {\includegraphics[height=2.8cm]{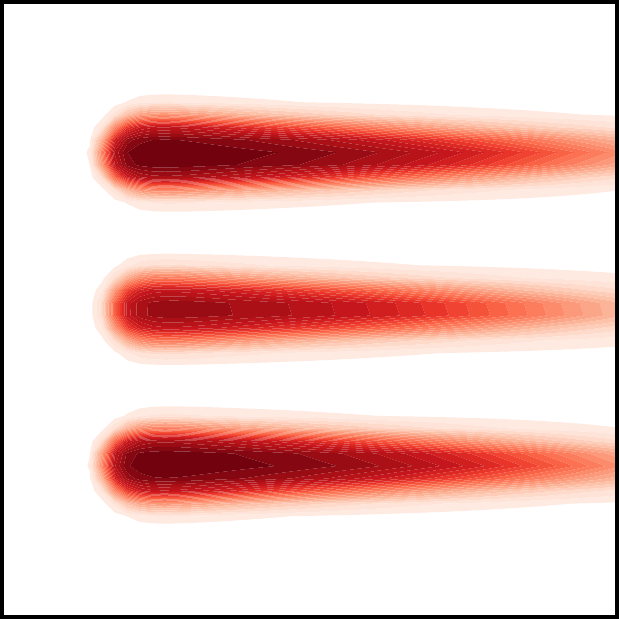}};
    \draw[dashed] ($(p_high.south)!0.5!(target.south)$) -- ($(p_high.north)!0.5!(target.north) + (0, 0.5)$);
    \draw[dashed] ($(a_high.south)!0.5!(p_low.south)$) -- ($(a_high.north)!0.5!(p_low.north) + (0, 0.5)$);
    \node[anchor=center] at (target.north) {target $q(\vz)$\strut};
    \node[anchor=center] (p_high_text) at (p_high.north) {$p(\vz)$\strut};
    \node[anchor=north] at (p_high.south) {\small $\kld*{q(\vz)\!}{\!p(\vz)}=0.06$};
    \node[anchor=center] (a_high_text) at (a_high.north) {$a(\vz)$\strut};
    \node[anchor=north] at (a_high.south) {\small $Z=0.09$};
    \node[anchor=center] (p_low_text) at (p_low.north) {$p(\vz)$\strut};
    \node[anchor=north] at (p_low.south) {\small $\kld*{q(\vz)\!}{\!p(\vz)}=0.66$};
    \node[anchor=center] (a_low_text) at (a_low.north) {$a(\vz)$\strut};
    \node[anchor=north] at (a_low.south) {\small $Z=0.17$};
    \node[above, anchor=center] at ($(a_high_text.north)!0.5!(p_high_text.north)$) {high capacity $a(\vz)$: MLP [2-10-10-1]};
    \node[above, anchor=center] at ($(a_low_text.north)!0.5!(p_low_text.north)$) {low capacity $a(\vz)$: MLP [2-10-1]};
    \end{tikzpicture}
    \caption[]{Learned acceptance functions $a(\vz)$ (red) that approximate a fixed target $q$ (blue) by reweighting a $\mathcal{N}(0,1)$ proposal (\protect\tikz[baseline=(current bounding box.base),inner sep=0pt]{\draw[line width=1pt, dashed] (0pt,3pt) --(15pt,3pt);}) to obtain an approximate density $p(\vz)$ (green). 
    Darker values correspond to higher numbers.}
    \label{fig:app:toy}
\end{figure*}

\FloatBarrier

\subsection{\Lars{} in combination with a non-factorial prior on Omniglot}
In \cref{tab:realnvp_proposal_omniglot} we show results for combining a RealNVP proposal with \Lars{}. Similarly to MNIST (\cref{tab:realnvp_proposal_mnist} in the main paper), we find that a RealNVP used as a prior by itself outperforms a combination of VAE with resampled prior. However, if we combine the RealNVP proposal with \Lars{}, we obtain a further improvement.

\begin{table}[htb]
    \centering
    \rowcolors{1}{white}{lightgray}
    \begin{tabular}{l>{$}c<{$}>{$}c<{$}}
    \toprule
    \hiderowcolors \textsc{Model} & \text{NLL} & Z\\\showrowcolors\midrule
    \VAE{} ($L=1$) + $\mathcal{N}(0,1)$ prior & 104.53 & 1\\
    \VAE{} + RealNVP prior & 101.50 & 1\\
    \VAE{} + resampled $\mathcal{N}(0,1)$ & 102.68 & 0.012\\
    \VAE{} + resampled RealNVP & 100.56 & 0.018\\
    \bottomrule
    \end{tabular}
    \caption[short]{Test NLL on dynamic Omniglot. Equivalent table to \cref{tab:realnvp_proposal_mnist} in the main paper.}
    \label{tab:realnvp_proposal_omniglot}
\end{table}

\clearpage 
\subsection{Samples from jointly trained VAE models}
\label{app:sec:samples_jointly_trained}

\begin{figure}[htb]
    \centering
    \begin{tikzpicture}
    \node (mnist_best) {\includegraphics[width=1.5in]{figures/samples/mnist_mlp/1729353_wid5_mean_best.pdf}};
    \node[right=0. of mnist_best] (omniglot_best) { \includegraphics[width=1.5in]{figures/samples/omniglot_mlp/1734458_wid5_mean_best.pdf}};
    \node[right=0. of omniglot_best] { \includegraphics[width=1.5in]{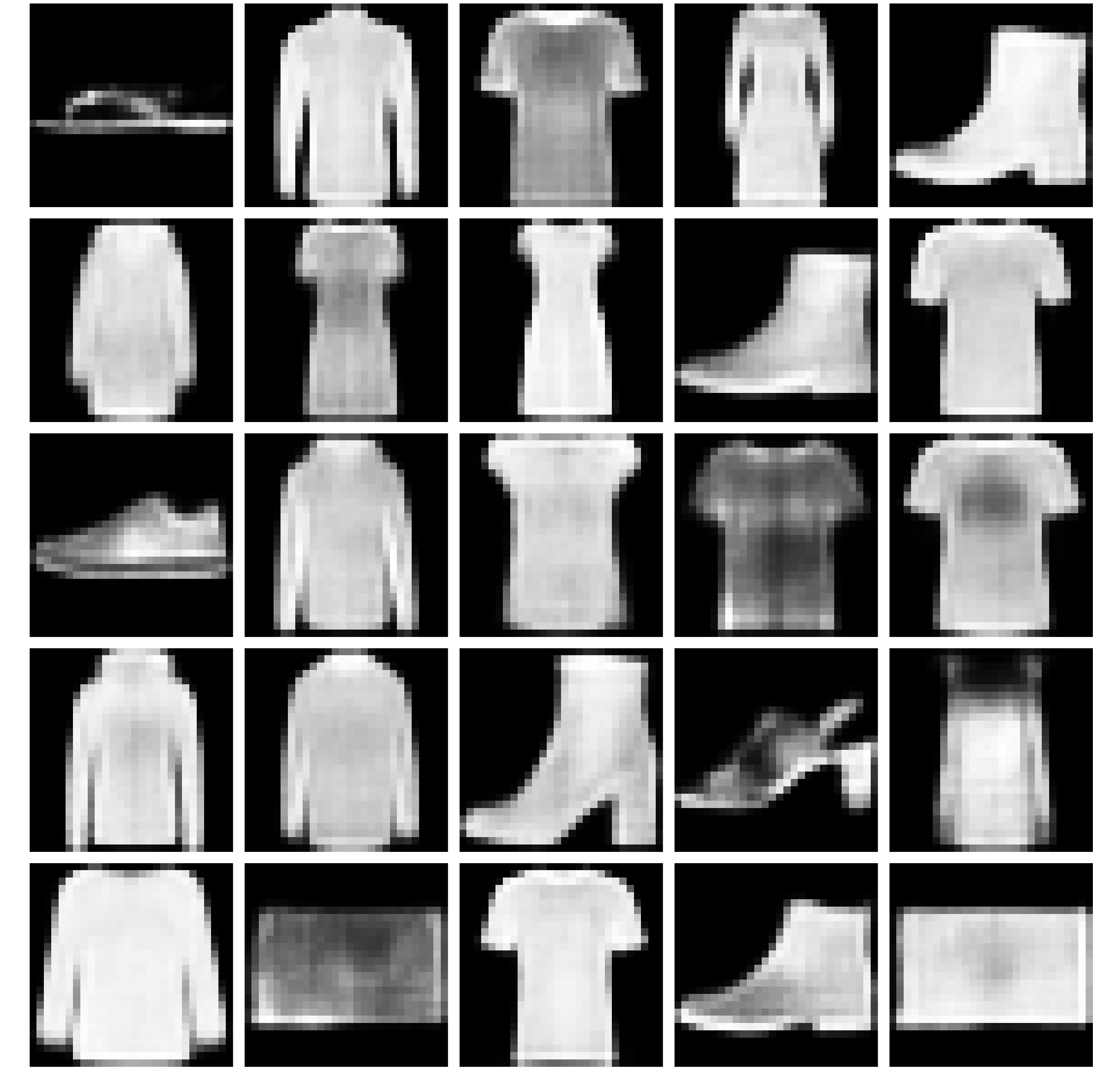}};
    \node[below=0. of mnist_best] (mnist_random) {\includegraphics[width=1.5in]{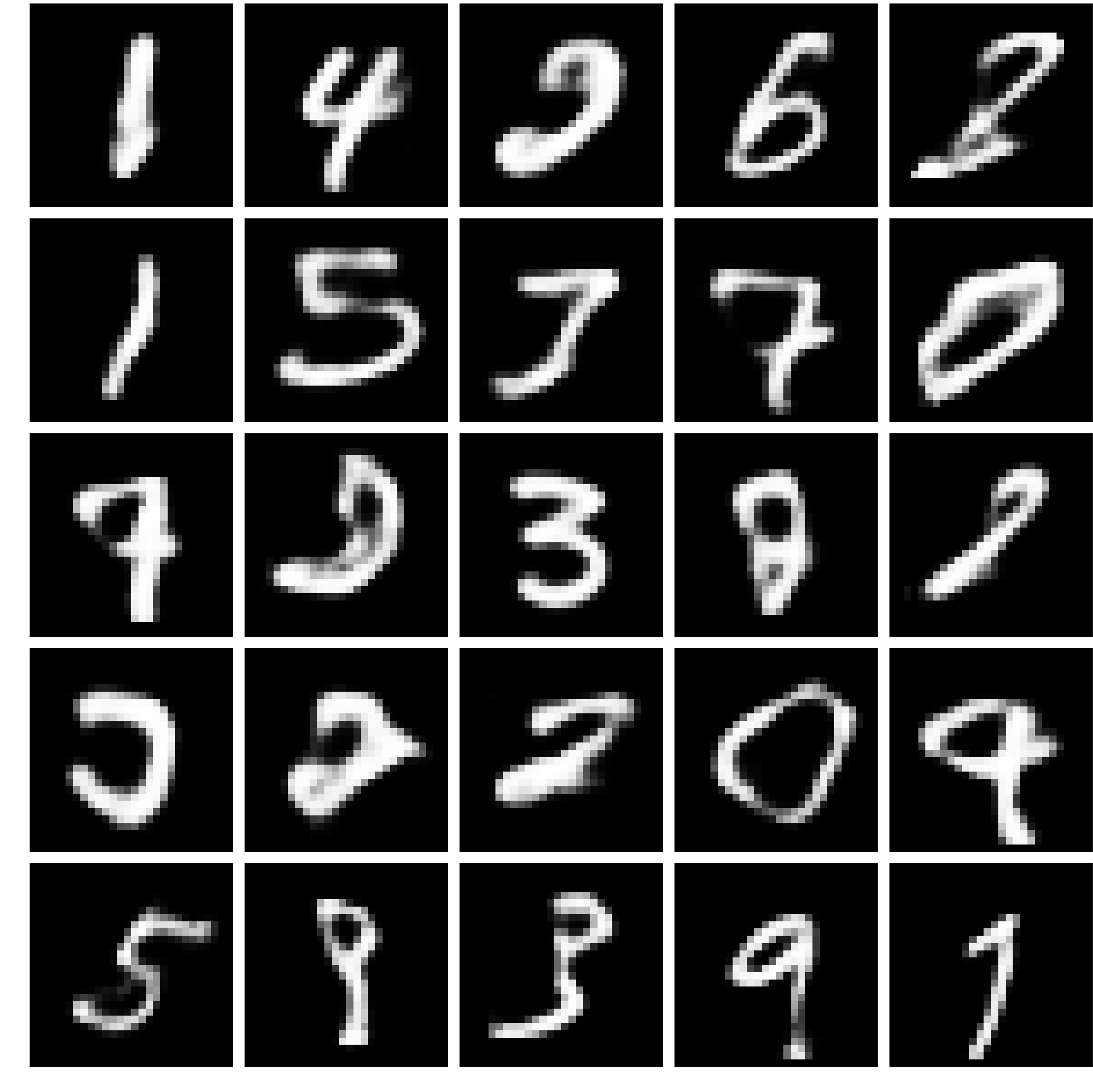}};
    \node[right=0. of mnist_random] (omniglot_random) { \includegraphics[width=1.5in]{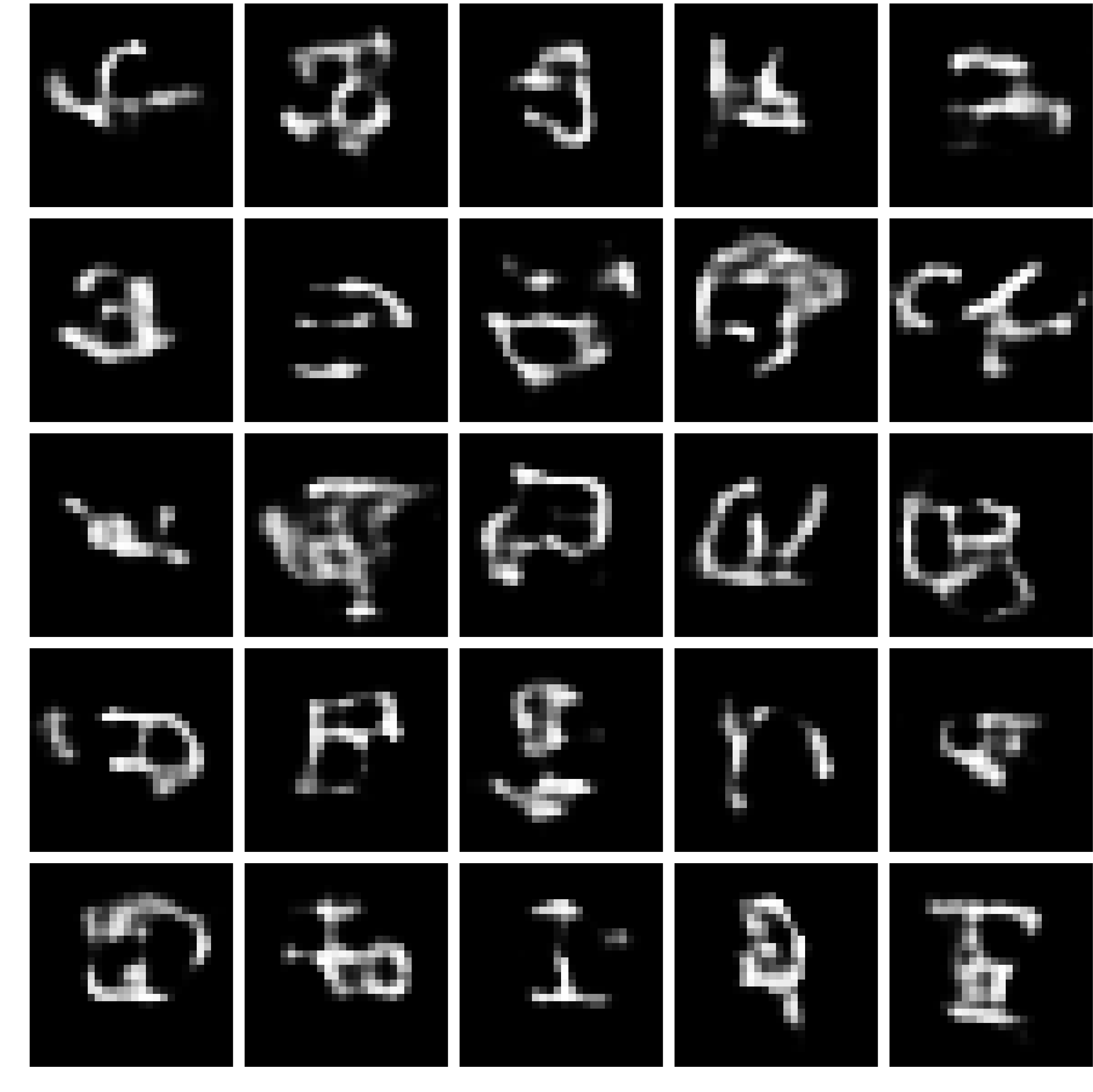}};
    \node[right=0. of omniglot_random] { \includegraphics[width=1.5in]{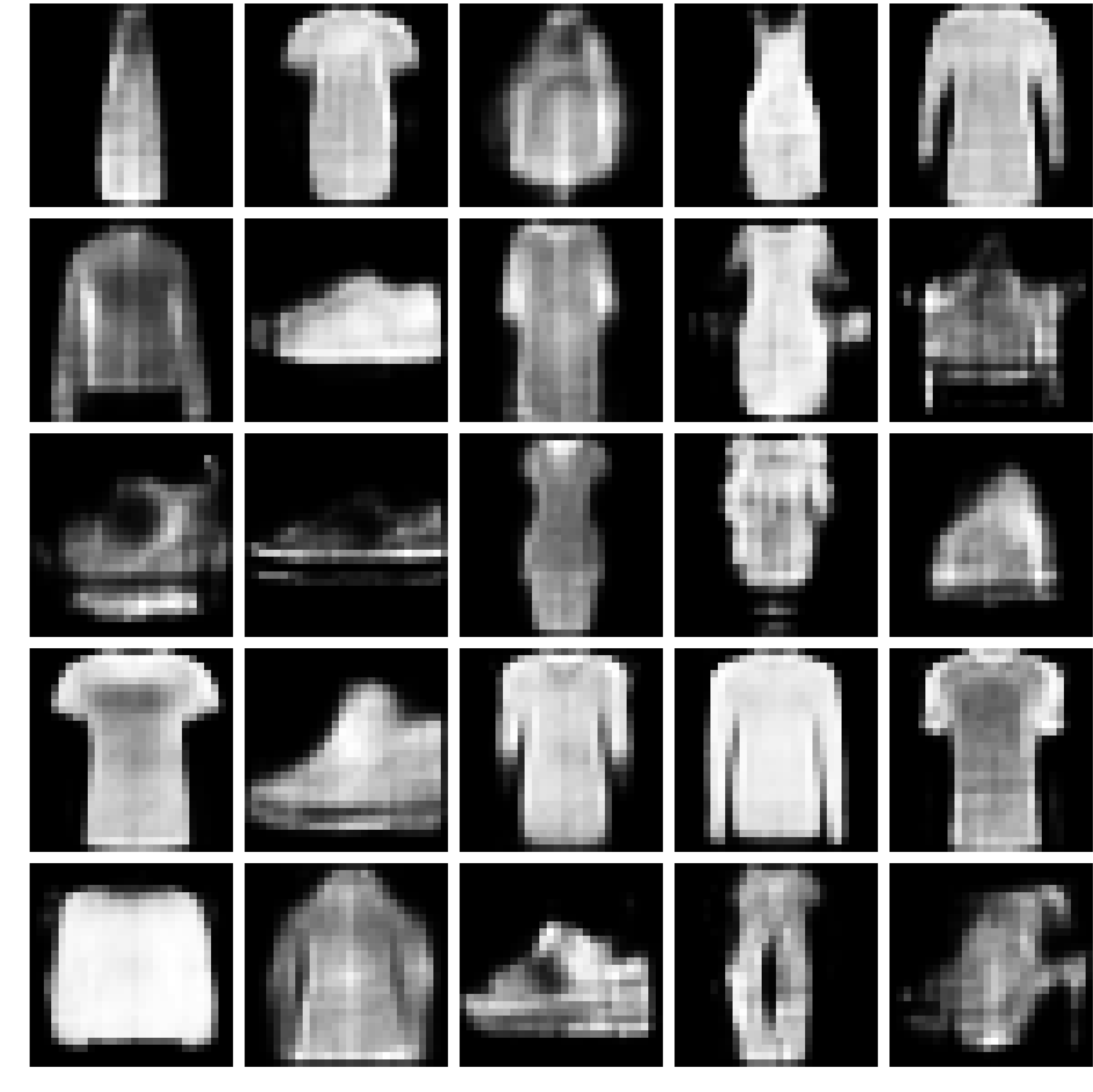}};
    \node[below=0. of mnist_random] (mnist_worst) {\includegraphics[width=1.5in]{figures/samples/mnist_mlp/1729353_wid5_mean_worst.pdf}};
    \node[right=0. of mnist_worst] (omniglot_worst) { \includegraphics[width=1.5in]{figures/samples/omniglot_mlp/1734458_wid5_mean_worst.pdf}};
    \node[right=0. of omniglot_worst] (fashion_worst) { \includegraphics[width=1.5in]{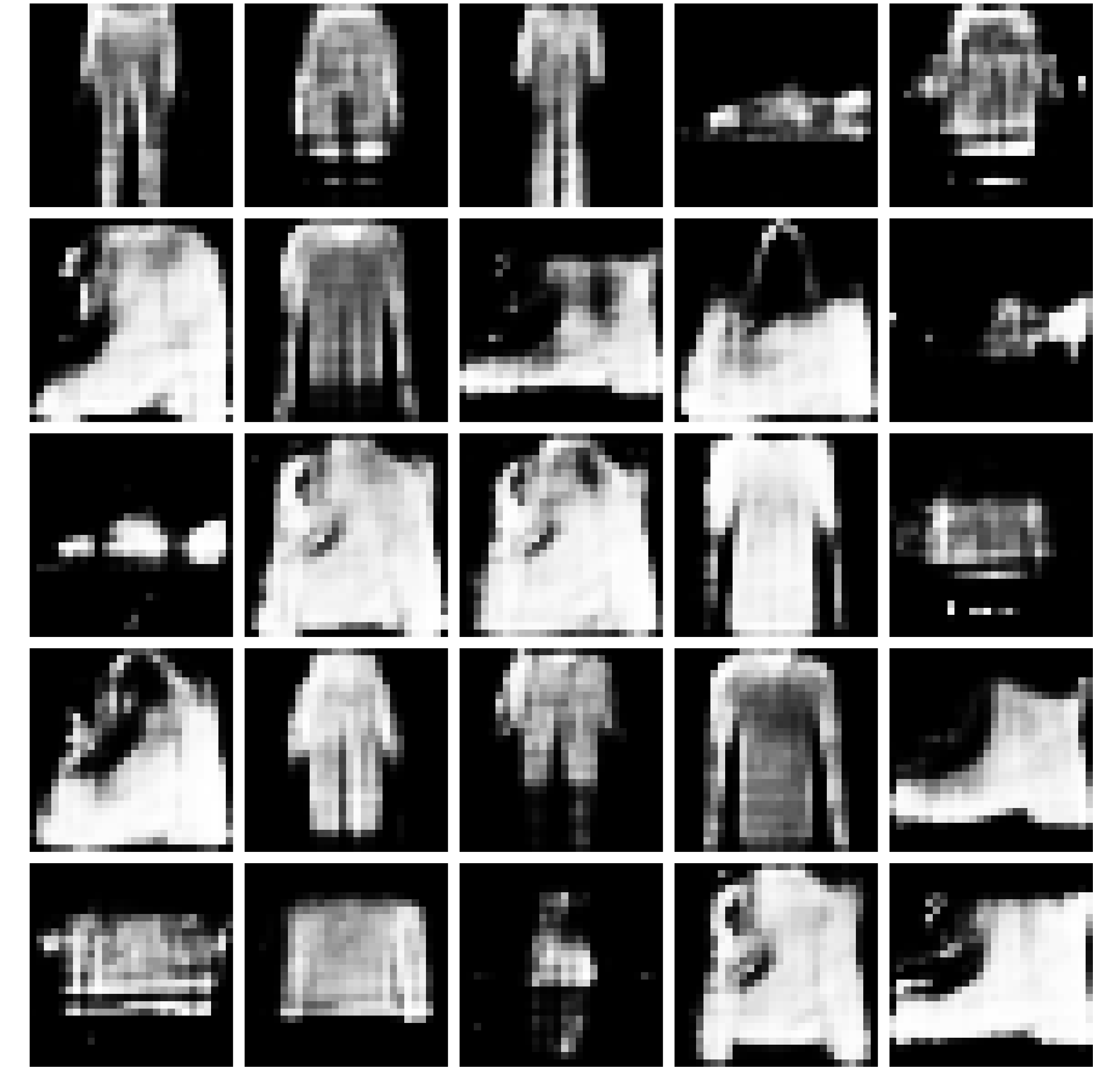}};
    \node[below=0. of mnist_worst] {MNIST};
    \node[below=0. of omniglot_worst] {Omniglot};
    \node[below=0. of fashion_worst] {FashionMNIST};
    \node[left=0.2 of mnist_best, rotate=90, anchor=center] {highest $a(\vz)$};
    \node[left=0.2 of mnist_random, rotate=90, anchor=center] {\makebox[1.5in][c]{random samples from proposal}};
    \node[left=0.2 of mnist_worst, rotate=90, anchor=center] {lowest $a(\vz)$};
    \end{tikzpicture}
    \caption{Comparison of sample means from a VAE with MLP encoder/decoder and jointly trained resampled prior. Samples shown are out of $10^4$ samples drawn. \textit{top:} highest $a(\vz)$; \textit{middle:} random samples from the proposal; \textit{bottom:} lowest $a(\vz)$. All samples from the simple VAE model with MLP encoder/decoder.}
    \label{fig:app:samples_mlp}
\end{figure}

\begin{figure}[htb]
    \centering
    \includestandalone{figures/mnist_mlp_samples_ranked}
    \includegraphics[width=\textwidth]{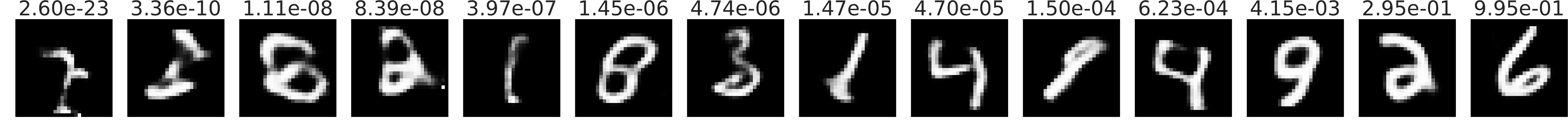}
    \caption{Samples from the proposal of a jointly trained VAE model with resampled prior on MNIST and their corresponding acceptance function values. \textit{top:} Distribution of the $a(\vz)$ values (index sorted by $a(\vz)$); note the log scale for $a(\vz)$. The average acceptance probability for this model is $Z \approx 0.014$. \textit{bottom:} Representative samples and their $a(\vz)$ values. The $a(\vz)$ value correlates with sample quality.}
    \label{fig:app:samples_mlp_ranked}
\end{figure}

We generated $10^4$ samples from the proposal $\pi(\vz) = \mathcal{N}(\vz; 0,1)$ of a jointly trained model and sorted them by their acceptance value $a(\vz)$. \cref{fig:app:samples_mlp} shows samples selected by their acceptance probability as assigned by the acceptance function $a(\vz)$. The top row shows samples with highest values (close to $1$), which would almost certainly be accepted by the resampled prior; visually, these samples are very good. The middle row shows 25 random samples from the proposal, some of which look very good while others show visible artefacts. The corresponding acceptance values typically reflect this, see also \cref{fig:app:samples_mlp_ranked}, in which we also show the $a(\vz)$ distribution for all $10^4$ samples from the proposal sorted by their $a$ value. The bottom row shows samples with the lowest acceptance probability (typically below $a(\vz) \approx 10^{-10}$), which would almost surely be rejected by the resampled prior. Visually, these samples are very poor. 

\subsection{Samples from applying \Lars{} to a pretrained VAE model}
\label{app:sec:samples_post-hoc}
\begin{figure}[htb]
    \centering
    \begin{tikzpicture}
    \node (mnist_best) {\includegraphics[width=1.5in]{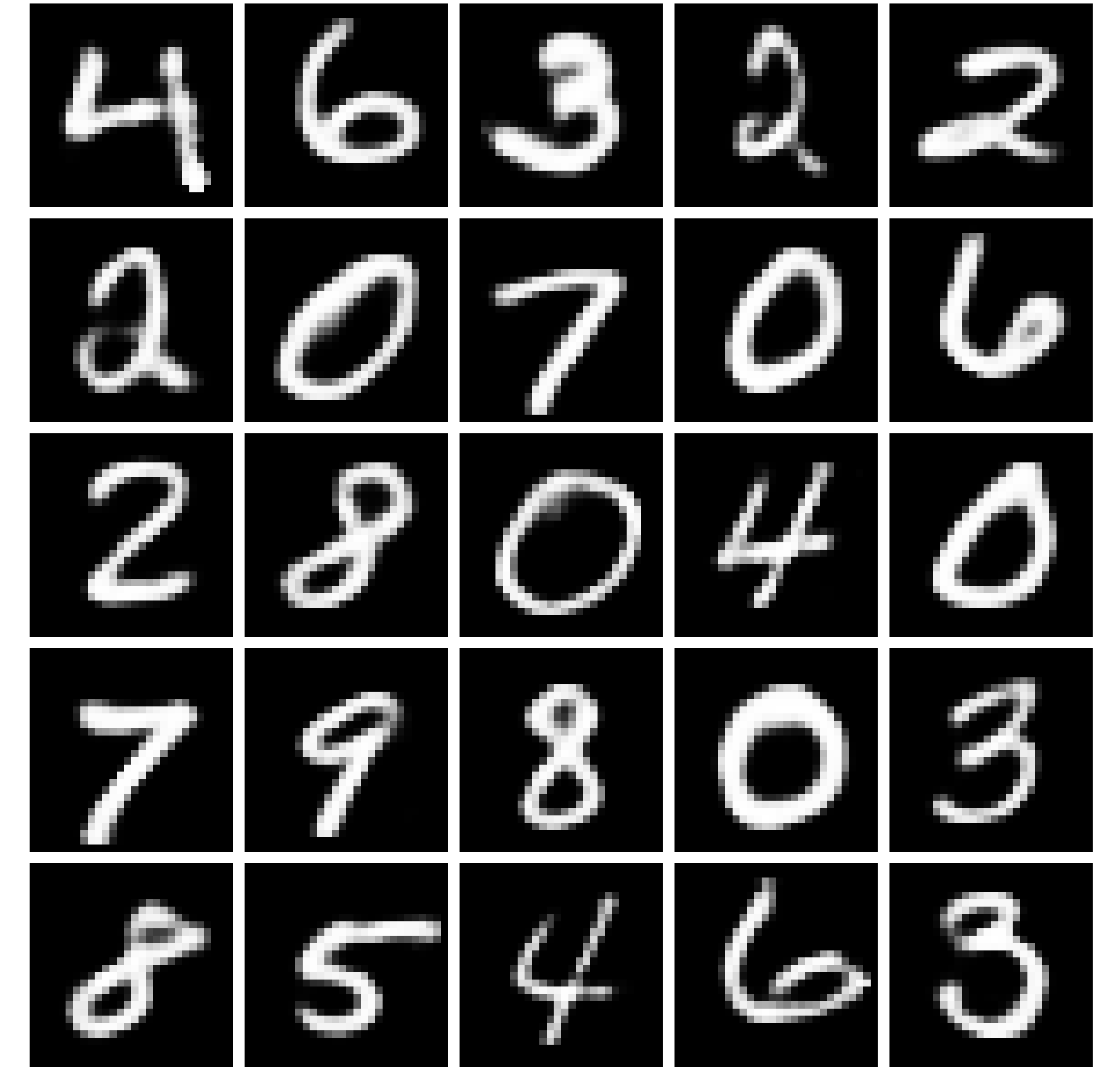} };
    \node[right=0. of mnist_best] (omniglot_best) { \includegraphics[width=1.5in]{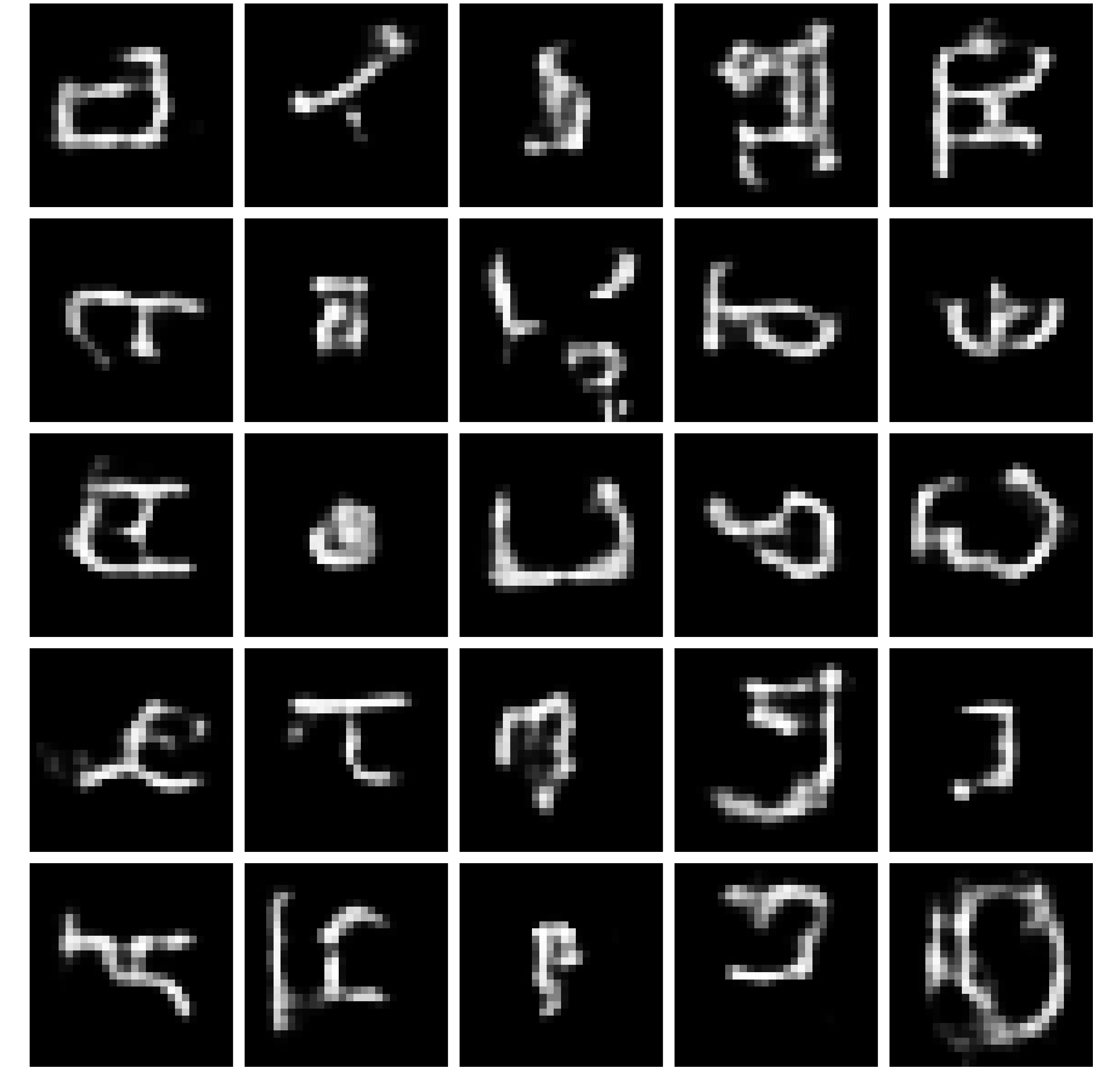}};
    \node[below=0. of mnist_best] (mnist_random) {\includegraphics[width=1.5in]{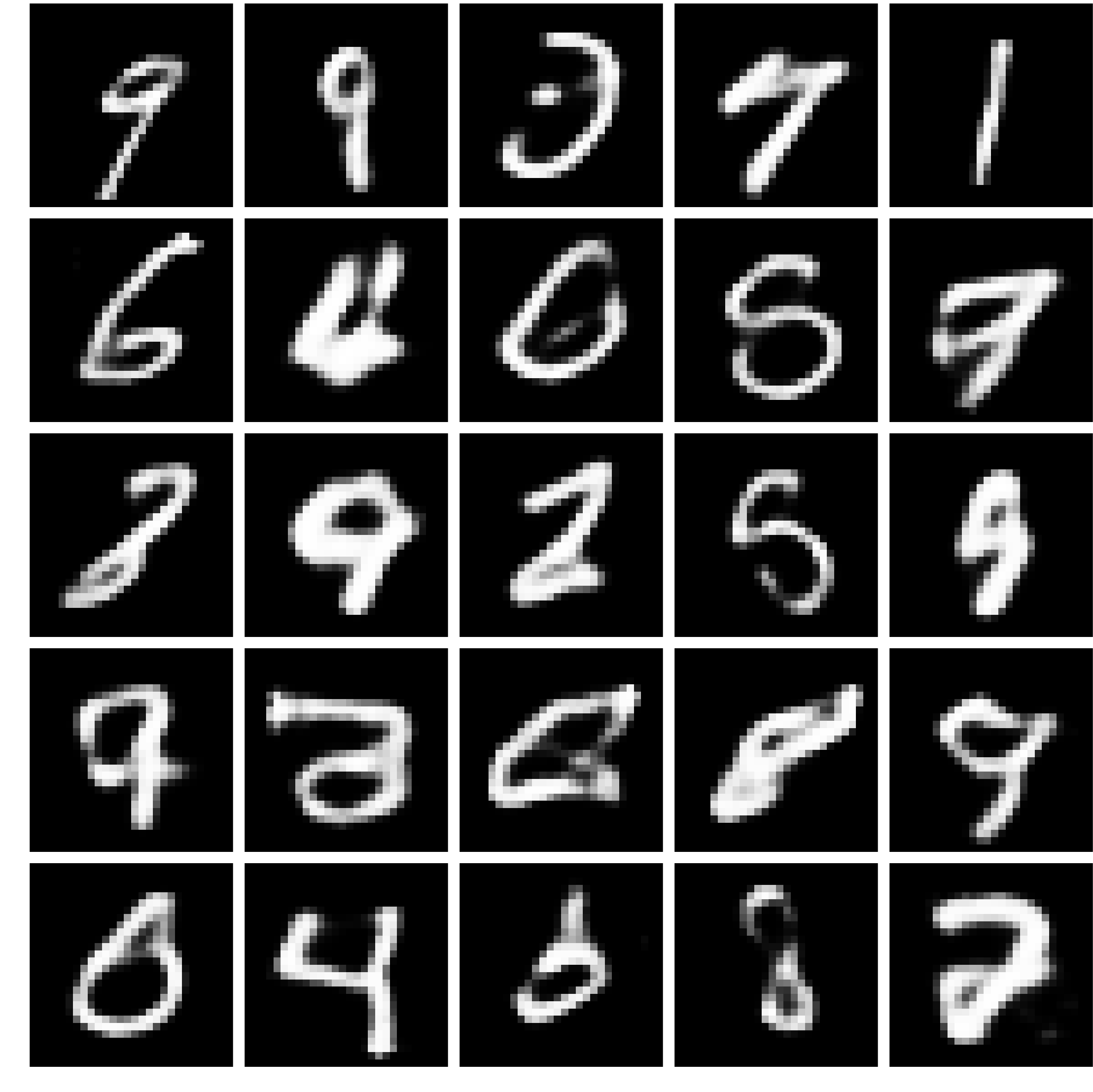} };
    \node[right=0. of mnist_random] (omniglot_random) { \includegraphics[width=1.5in]{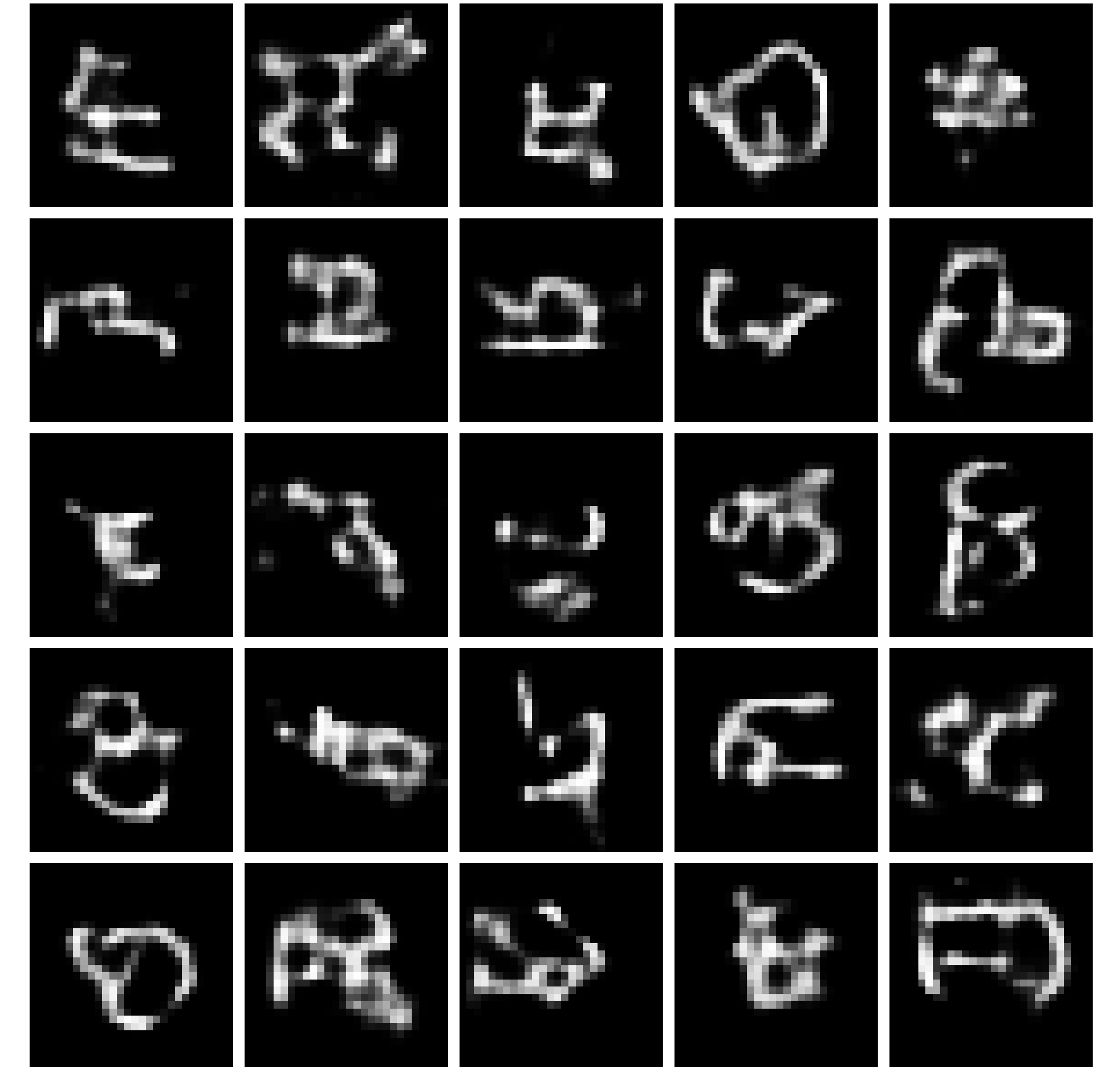}};
    \node[below=0. of mnist_random] (mnist_worst) {\includegraphics[width=1.5in]{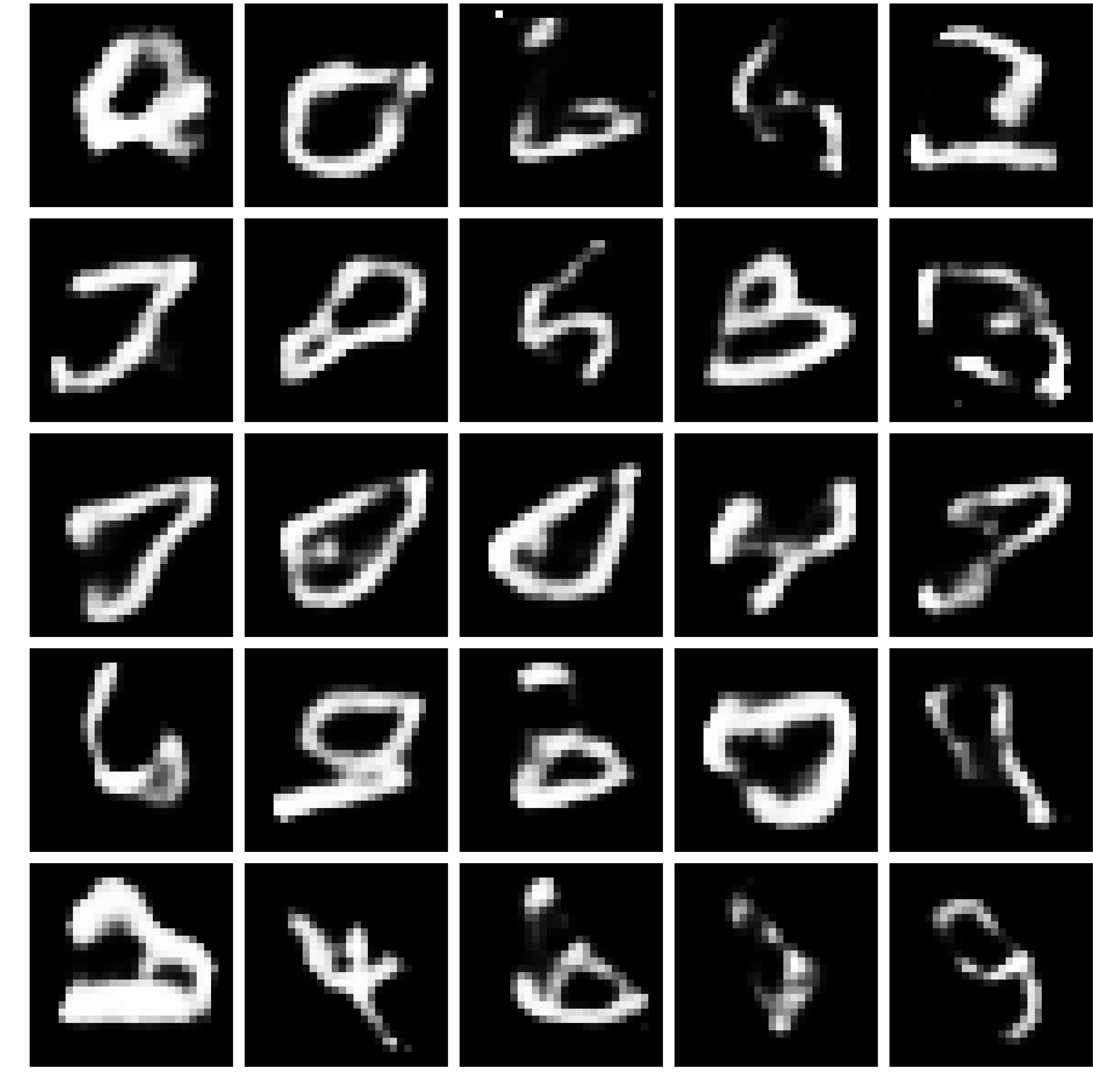} };
    \node[right=0. of mnist_worst] (omniglot_worst) {\includegraphics[width=1.5in]{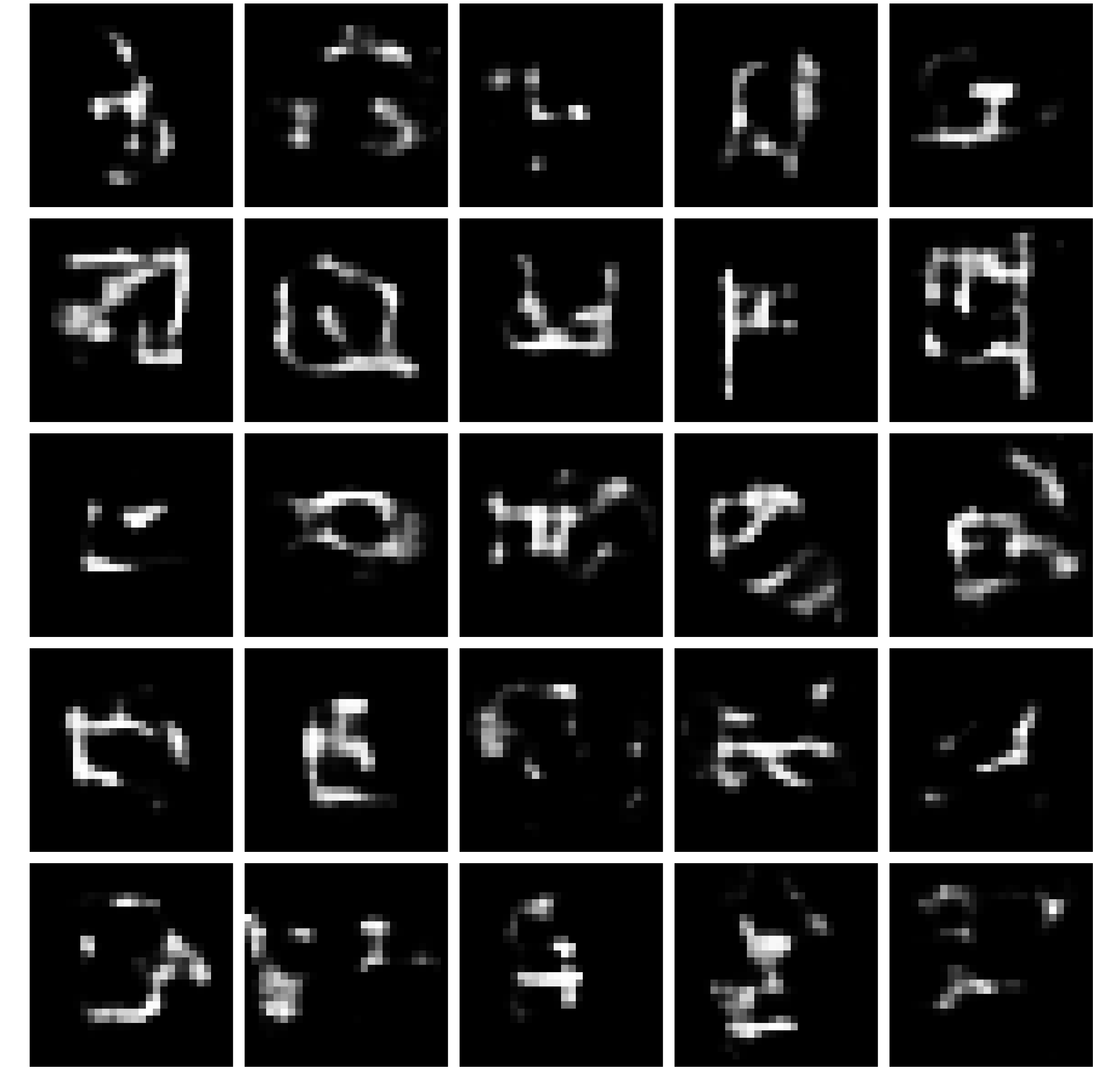}};
    \node[below=0. of mnist_worst] {MNIST};
    \node[below=0. of omniglot_worst] {Omniglot};
    \node[left=0.2 of mnist_best, rotate=90, anchor=center] {highest $a(\vz)$};
    \node[left=0.2 of mnist_random, rotate=90, anchor=center] {\makebox[1.5in][c]{random samples from proposal}};
    \node[left=0.2 of mnist_worst, rotate=90, anchor=center] {lowest $a(\vz)$};
    \end{tikzpicture}
    \caption{Comparison of sample means from a VAE with pretrained MLP encoder/decoder to which we applied a resampled \Lars{} prior post-hoc. Samples shown are out of $10^4$ samples drawn. \textit{top:} samples with highest $a(\vz)$; \textit{middle:} random samples from the proposal; \textit{bottom:} samples with lowest $a(\vz)$.}
    \label{fig:app:samples_mlp_posthoc}
\end{figure}
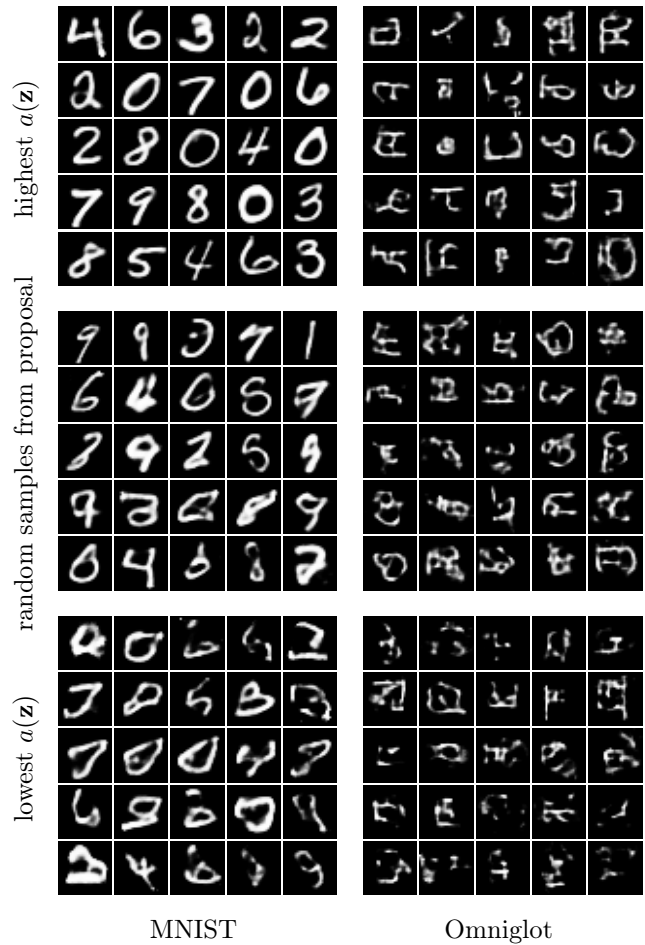

Similarly to \cref{app:sec:samples_jointly_trained}, we now show samples from a VAE that has been trained with the usual standard Normal prior and to which we applied the resampled prior post-hoc, see \cref{fig:app:samples_mlp_posthoc}. We fixed the encoder and decoder and only trained the acceptance function on the usual ELBO objective. 

\FloatBarrier

\subsection{Samples from a VAE with \Lars{} output distribution}
\label{app:sec:samples_output_vae}
\begin{figure}[htb]
    \centering
    \includegraphics[width=2in]{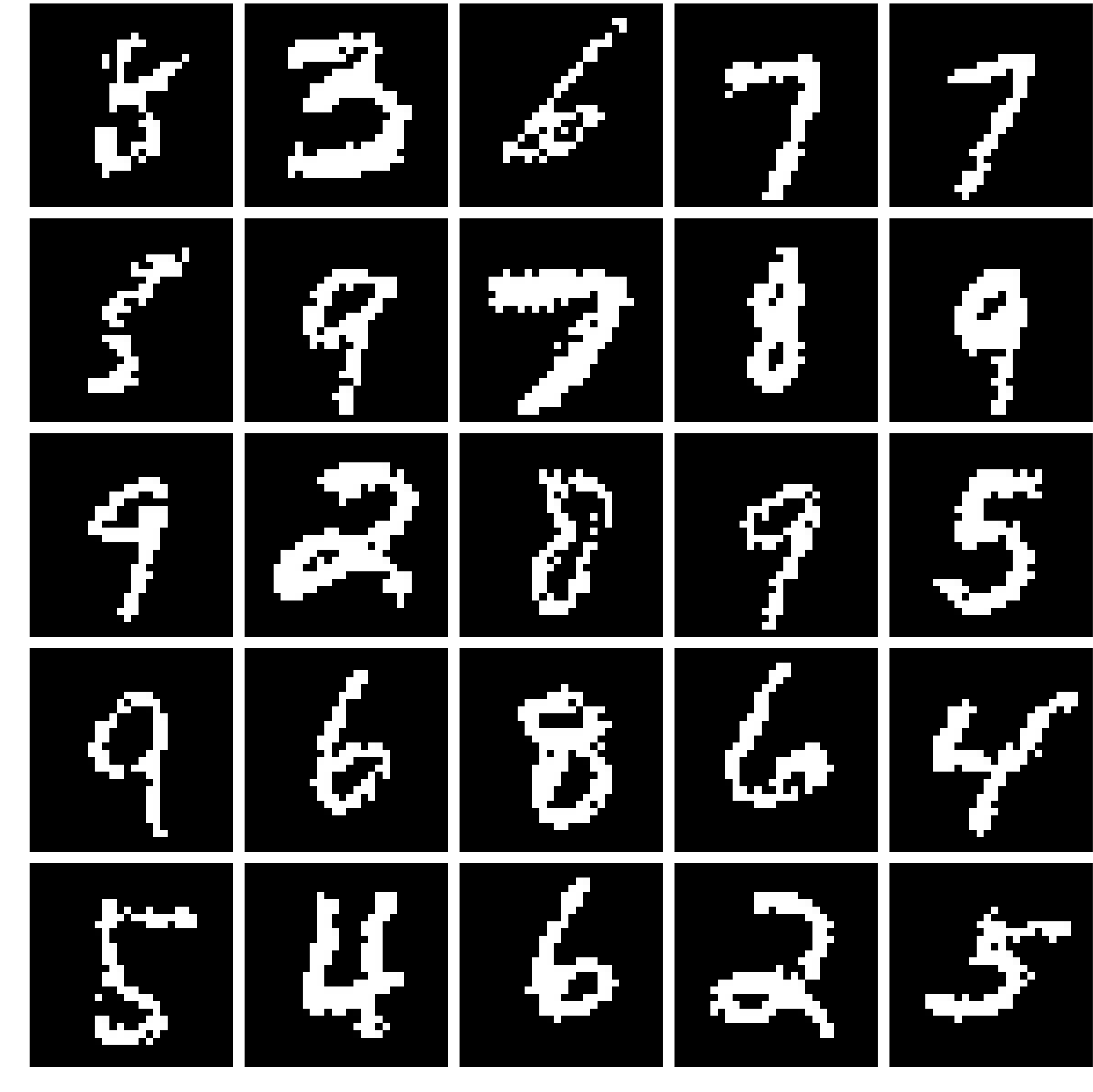}
    \includegraphics[width=2in]{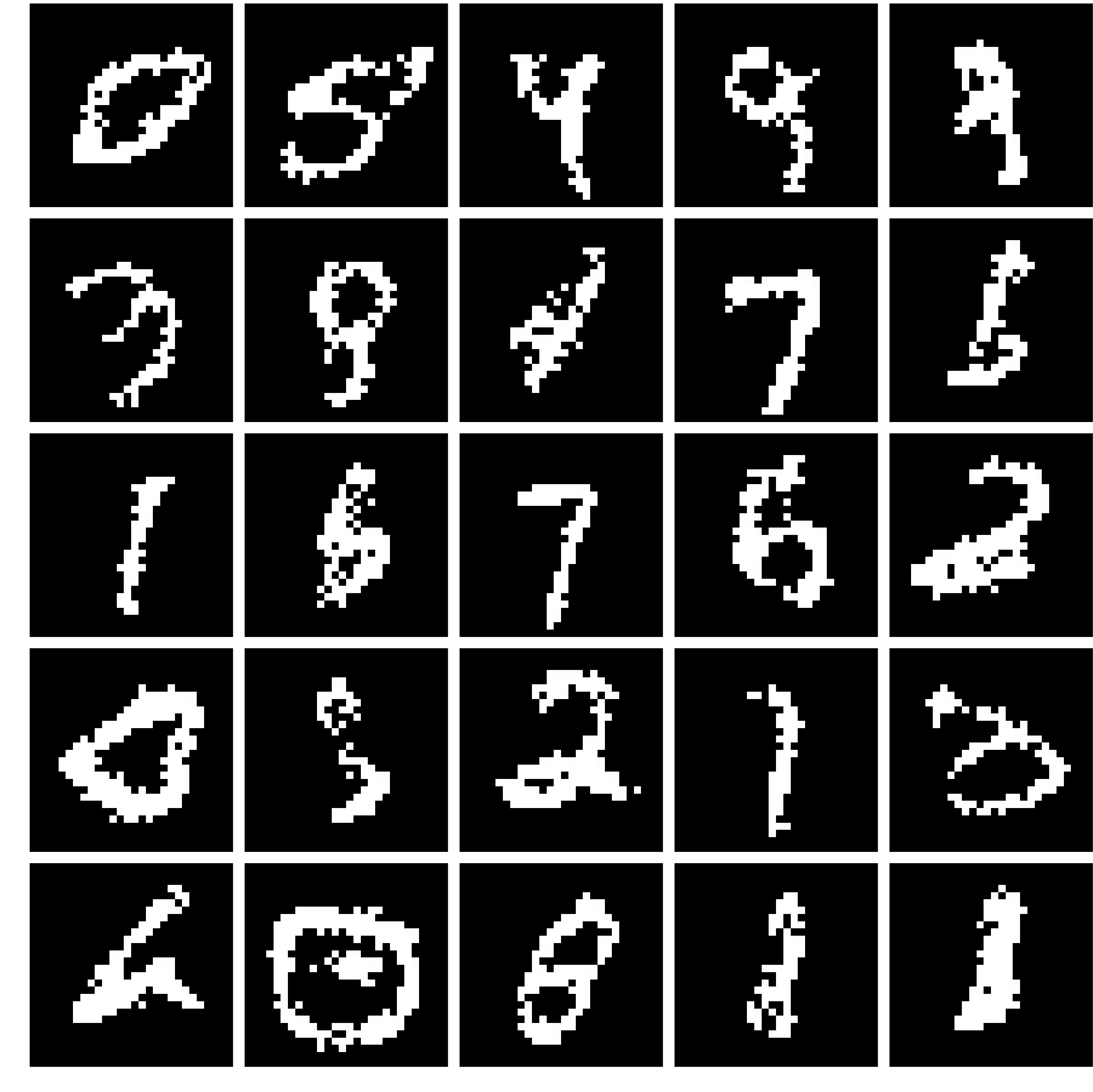}    \includegraphics[width=2in]{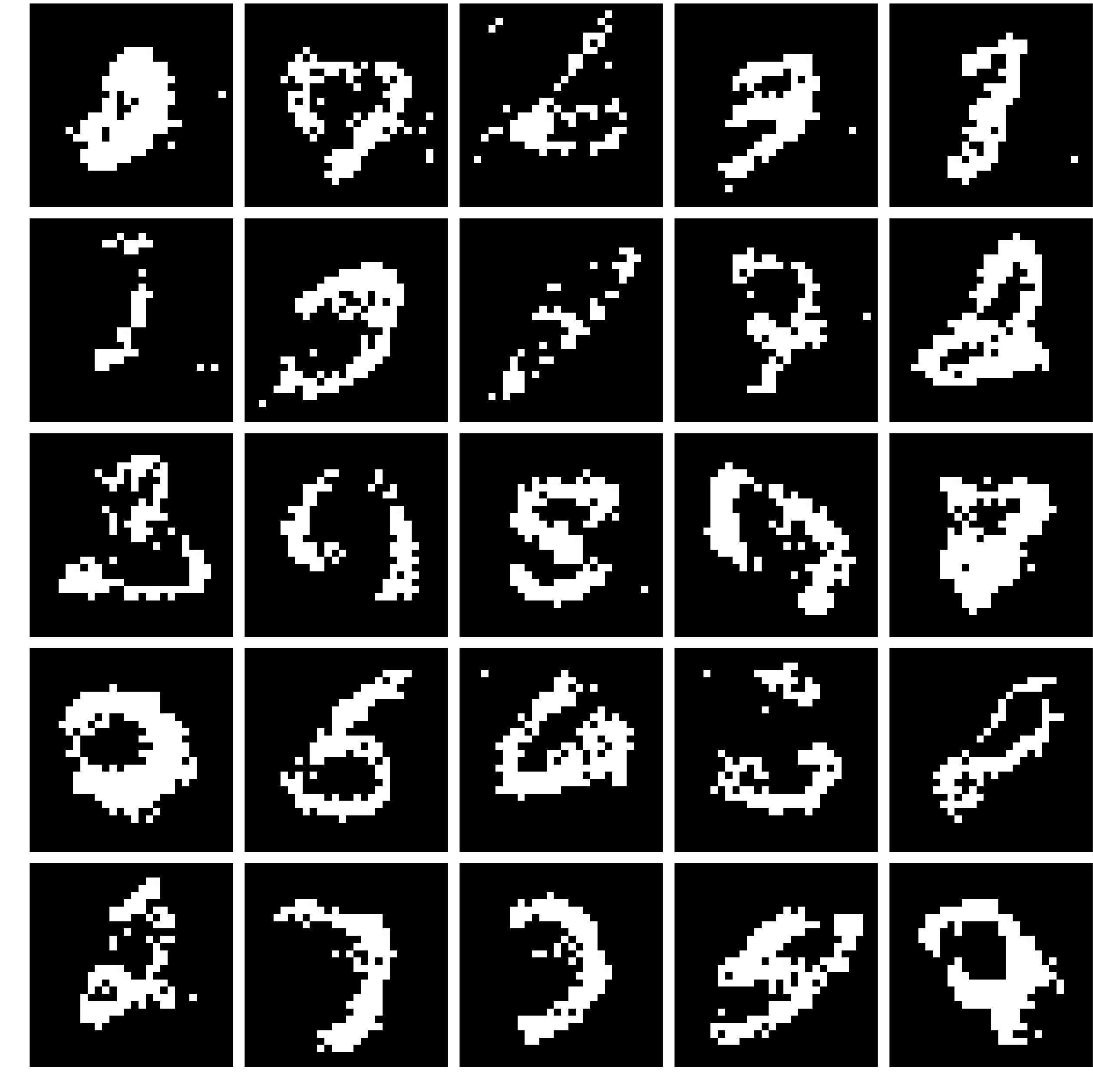}\\
    \hfill \makebox[2in][c]{Best samples $a(\vx) \approx 1$}\makebox[2in][c]{Random samples}\makebox[2in][c]{Worst samples $a(\vx) \leq 10^{-10}$}\hfill  \hspace{0.1cm} \\
    \vspace{0.5cm}
    \includegraphics[width=\textwidth]{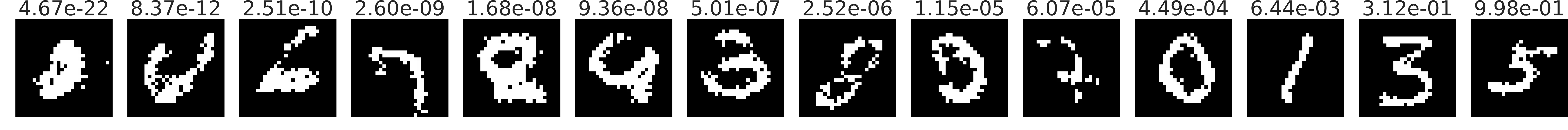}
    \caption{Samples from a VAE with a jointly trained acceptance function on the \emph{output}, that is, with a resampled marginal likelihood. \textit{top:} Samples grouped by the acceptance function (best, random, and worst); \textit{bottom:} Samples and their acceptance function values.}
    \label{fig:app:samples_output_vae_mnist}
\end{figure}

In \cref{fig:app:samples_output_vae_mnist} we show ranked samples for a VAE with resampled marginal log likelihood, that is, we apply \Lars{} on the \emph{discrete} output space rather than in the continuous latent space. We find that even in this high dimensional space, our approach works well, and $a(\vx)$ is able to reliably rank images.

\subsection{VAE with resampled prior and 2D latent space}
\label{app:sec:2dlatent_space}

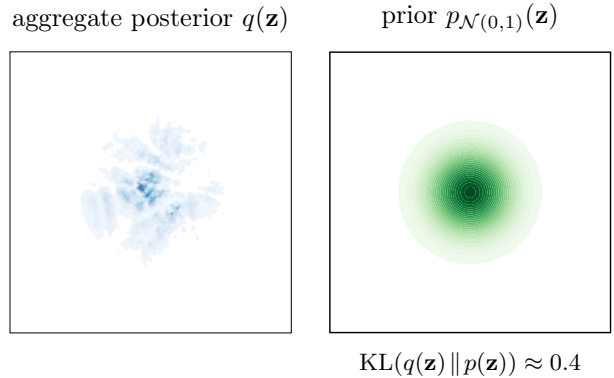
\begin{figure}[htb]
    \centering
    \begin{tikzpicture}
    \node (q) {\includegraphics[height=3.75cm]{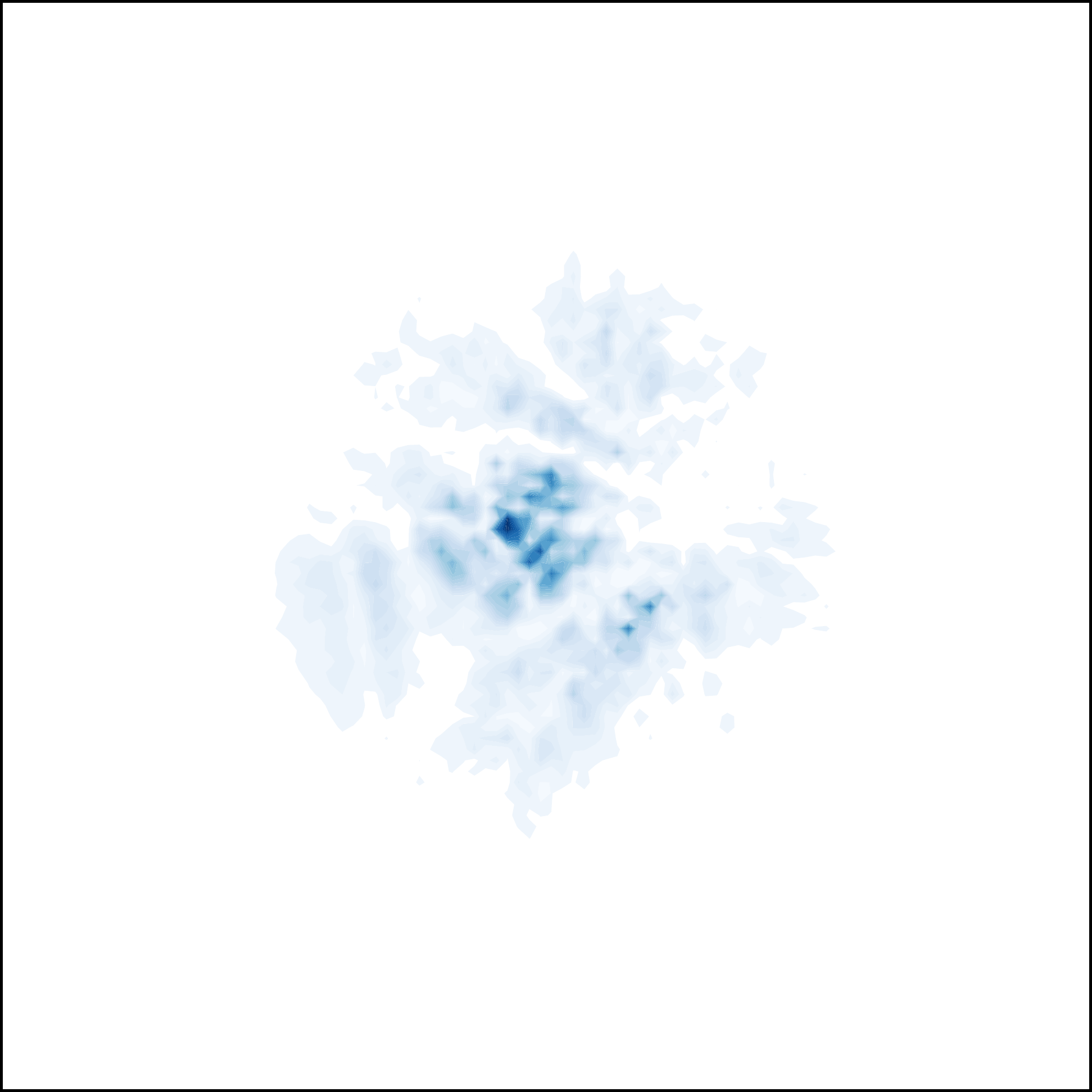}};
    \node[right=0.25cm of q] (p) {\includegraphics[height=3.75cm]{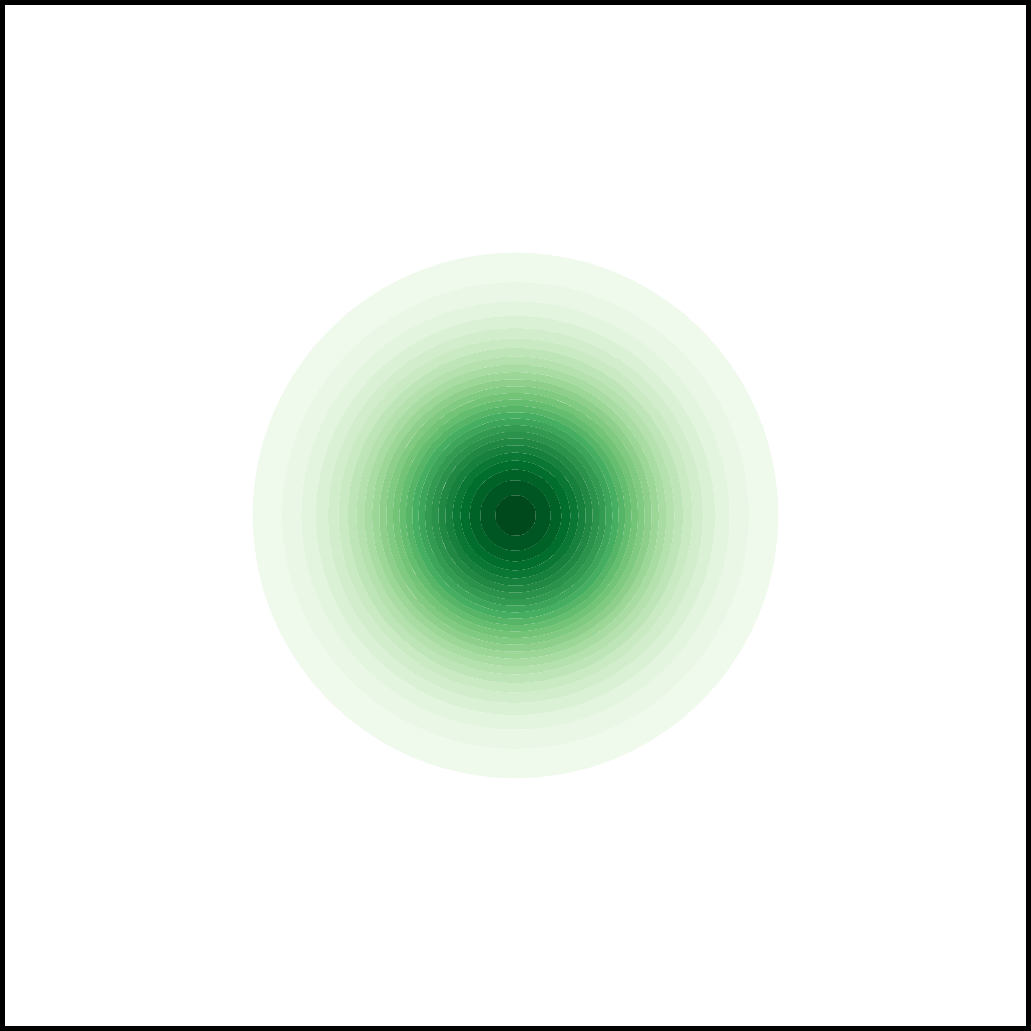}};
    \node[anchor=south] at (q.north) {aggregate posterior $q(\vz)$};
    \node[anchor=south] at (p.north) {prior $p_{\mathcal{N}(0,1)}(\vz)$};
    \node[anchor=north] at (p.south) {\small $\kld*{q(\vz)\!}{\!p(\vz)}\approx 0.4$};
    \end{tikzpicture}
    \caption[]{Aggregate posterior and fixed standard Normal prior for a regular VAE trained with the standard Normal prior. Dynamic MNIST}
    \label{fig:app:2dlatent_regular}
\end{figure}

\begin{figure}[htb]
    \centering
    \begin{tikzpicture}
    \node (q) {\includegraphics[height=3.75cm]{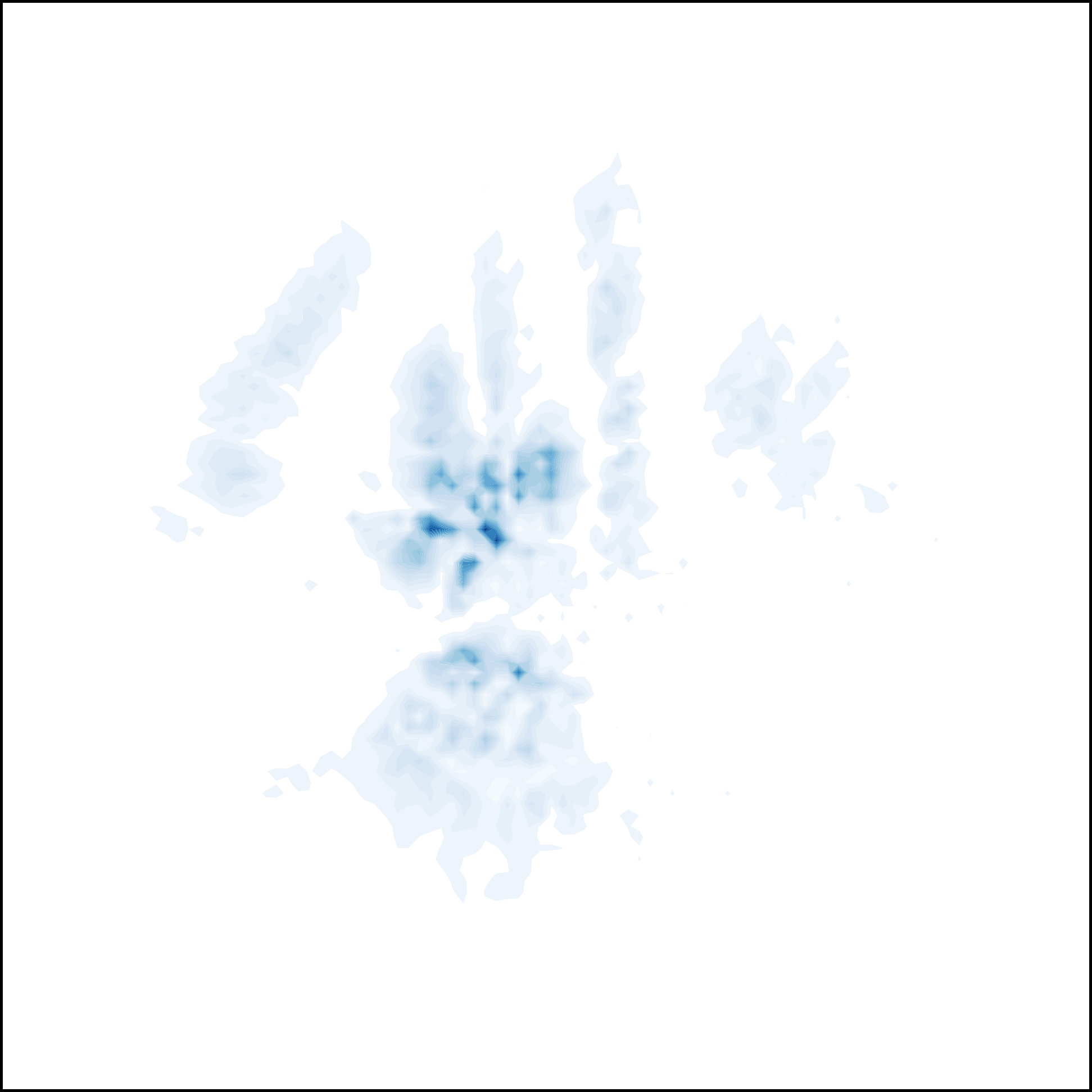}};
    \node[right=0.25cm of q] (pi) {\includegraphics[height=3.75cm]{figures/2dlatent/2dlatent_reject_pi.pdf}};
    \node[right=0.25cm of pi] (a) {\includegraphics[height=3.75cm]{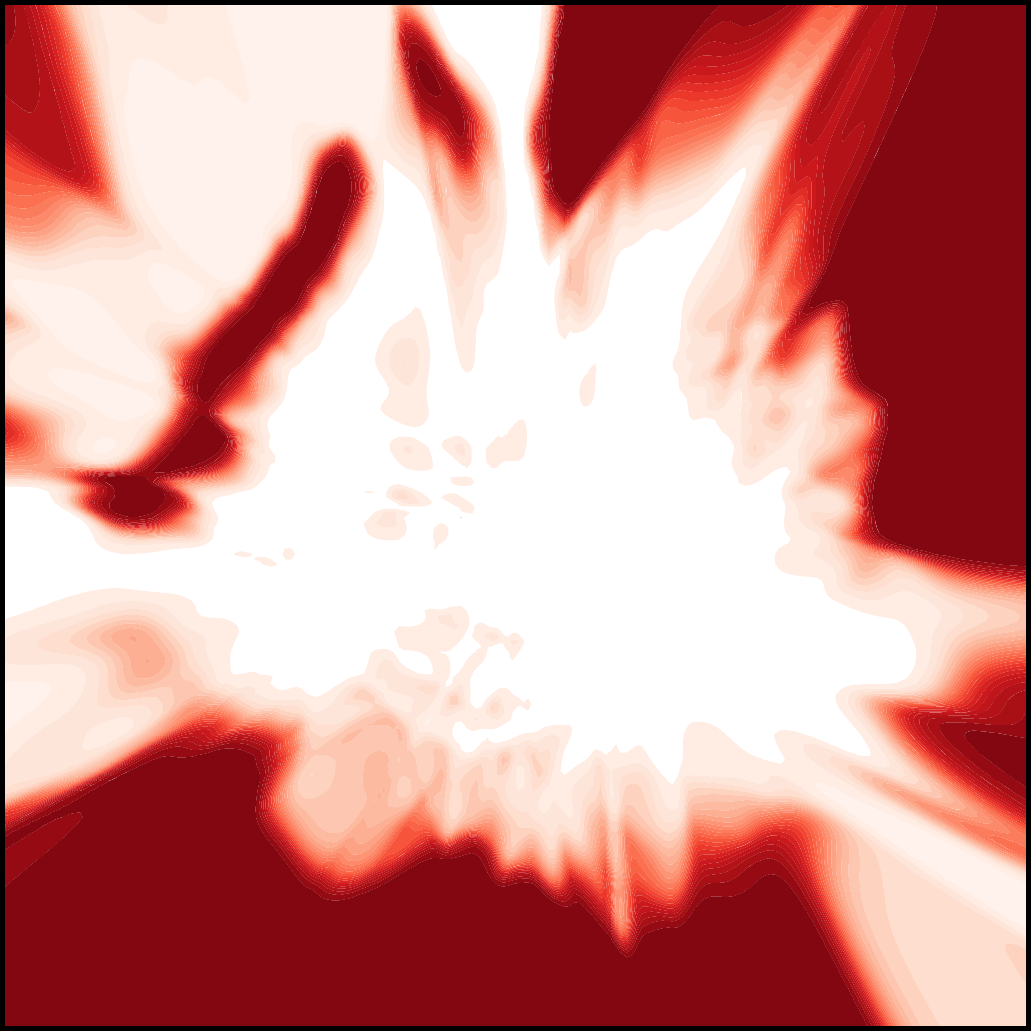}};
    \node[right=0.25cm of a] (p) {\includegraphics[height=3.75cm]{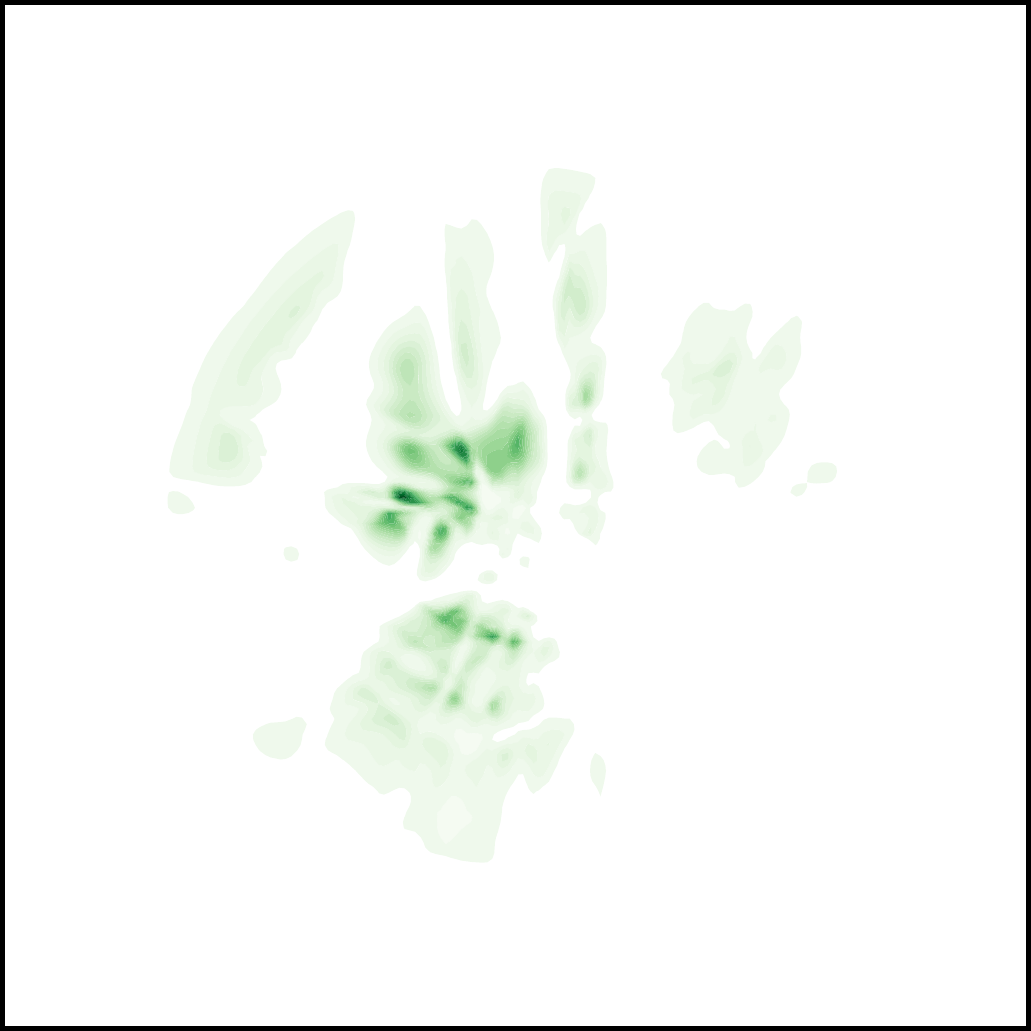}};
    \node[anchor=south] at (q.north) {aggregate posterior $q(\vz)$};
    \node[anchor=south] at (pi.north) {proposal $\pi(\vz)=\mathcal{N}(0,1)$};
    \node[anchor=north] at (pi.south) {\small $\kld*{q(\vz)\!}{\!\pi(\vz)}\approx 1.14$};
    \node[anchor=south] (a_text) at (a.north) {$a(\vz)$ \MLP{2-100-100-1}};
    \node[anchor=north] at (a.south) {\small $Z\approx 0.03$};
    \node[anchor=south] at (p.north) {\Lars{} prior $p_{T=100}(\vz)$};
    \node[anchor=north] at (p.south) {\small $\kld*{q(\vz)\!}{\!p(\vz)}\approx 0.15$};
    \end{tikzpicture}
    \caption[]{Aggregate posterior and resampled prior for a VAE with \Lars{} prior and \textbf{high capacity} network for $a$: $a=\MLP{2-100-100-1}$. Dynamic MNIST}
    \label{fig:app:2dlatent_reject}
\end{figure}

\begin{figure}[htb]
    \centering
    \begin{tikzpicture}
    \node (q) {\includegraphics[height=3.75cm]{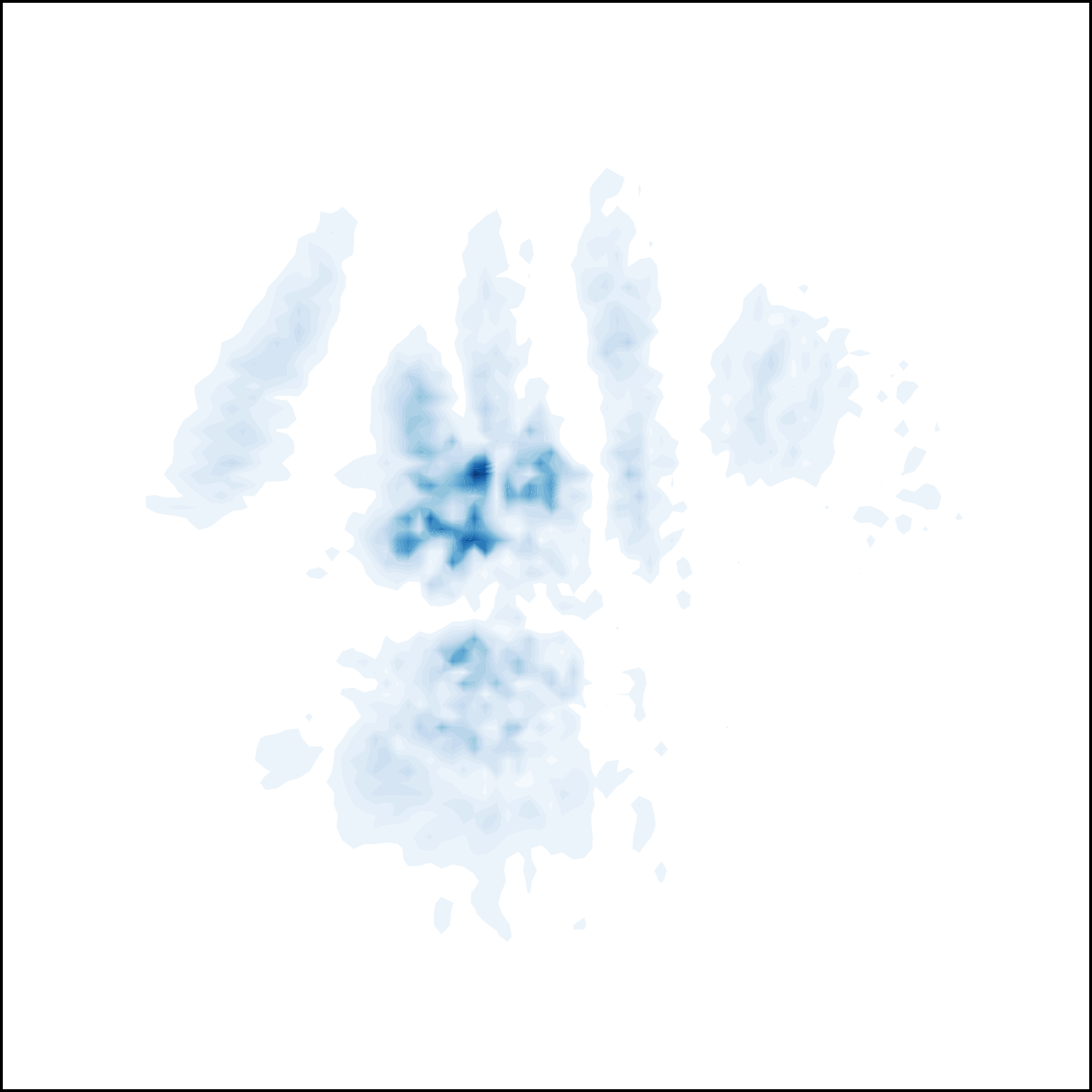}};
    \node[right=0.25cm of q] (pi) {\includegraphics[height=3.75cm]{figures/2dlatent/2dlatent_reject_pi.pdf}};
    \node[right=0.25cm of pi] (a) {\includegraphics[height=3.75cm]{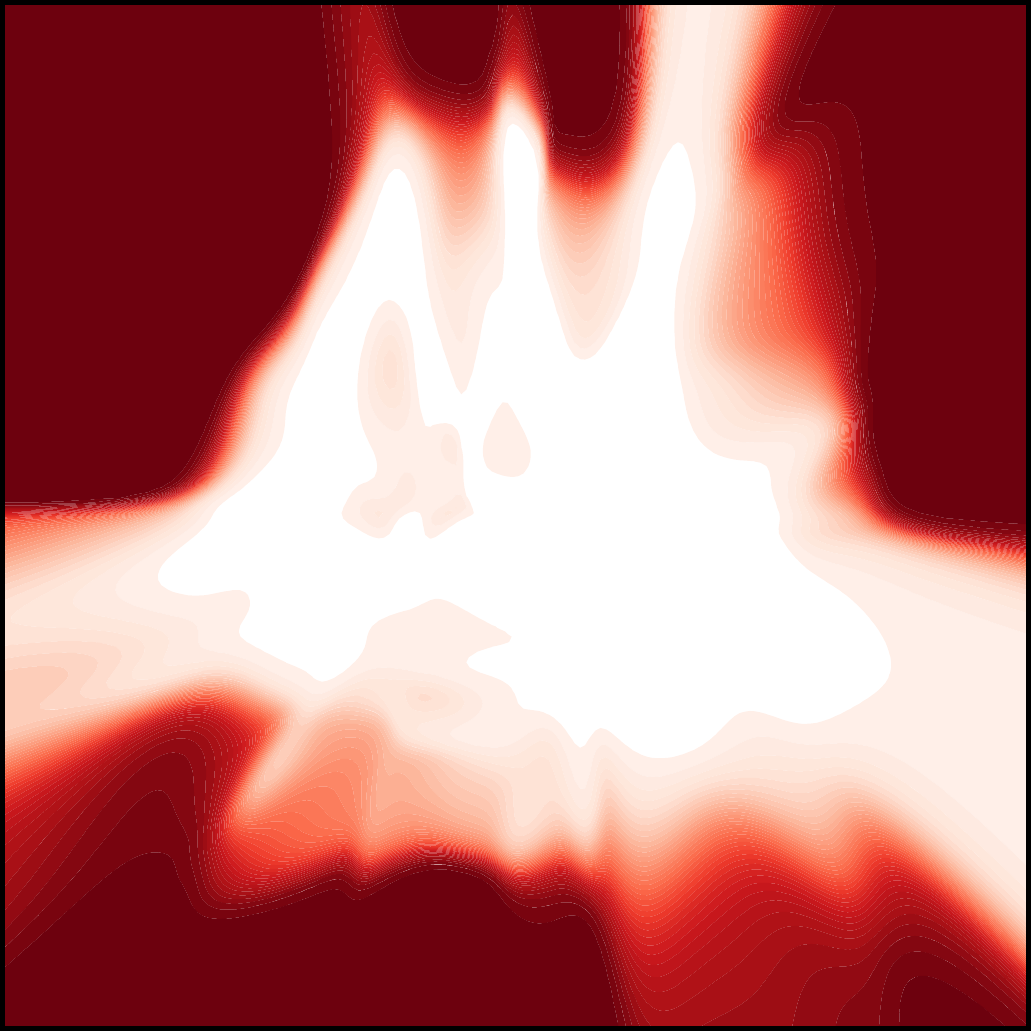}};
    \node[right=0.25cm of a] (p) {\includegraphics[height=3.75cm]{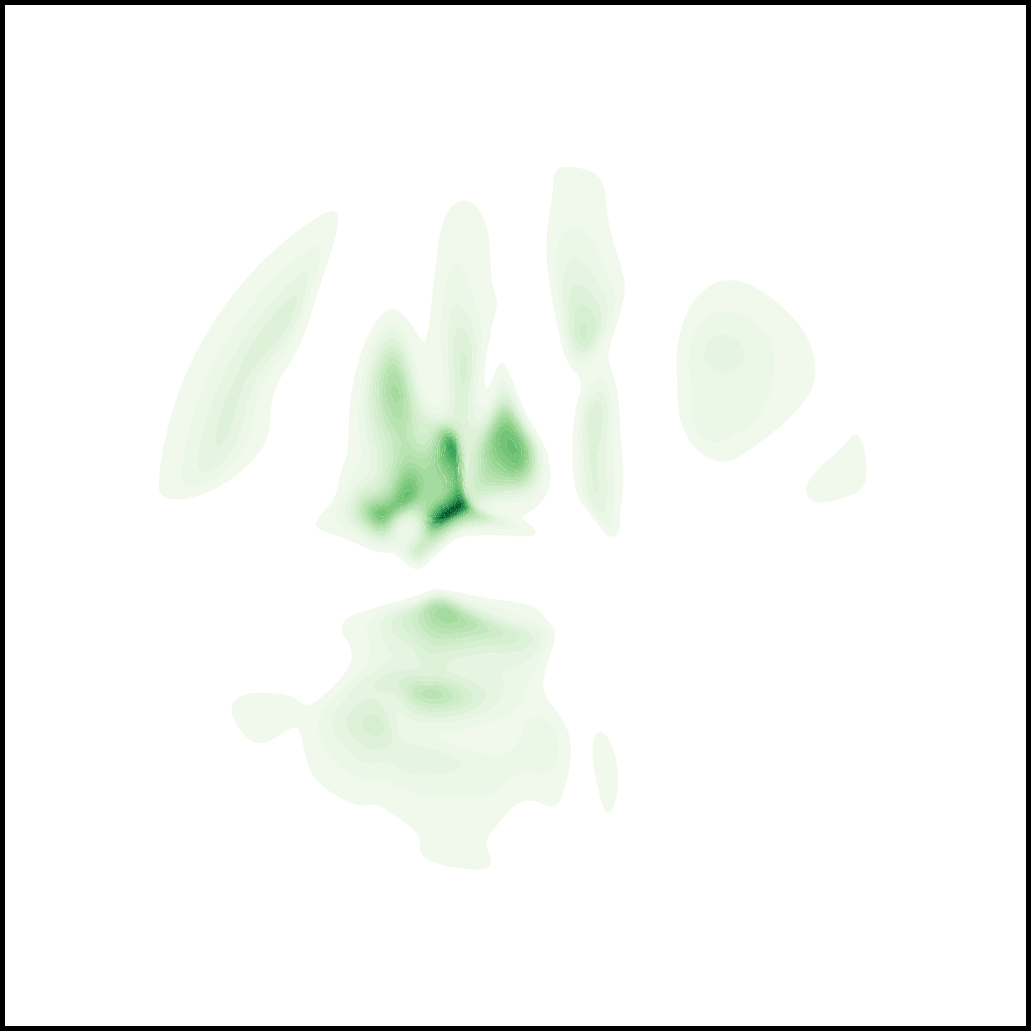}};
    \node[anchor=south] at (q.north) {aggregate posterior $q(\vz)$};
    \node[anchor=south] at (pi.north) {proposal $\pi(\vz)=\mathcal{N}(0,1)$};
    \node[anchor=north] at (pi.south) {\small $\kld*{q(\vz)\!}{\!\pi(\vz)}\approx 1.15$};
    \node[anchor=south] (a_text) at (a.north) {$a(\vz)$ \MLP{2-20-20-1}};
    \node[anchor=north] at (a.south) {\small $Z\approx 0.02$};
    \node[anchor=south] at (p.north) {\Lars{} prior $p_{T=100}(\vz)$};
    \node[anchor=north] at (p.south) {\small $\kld*{q(\vz)\!}{\!p(\vz)}\approx 0.20$};
    \end{tikzpicture}
    \caption[]{Aggregate posterior and resampled prior for a VAE with \Lars{} prior and \textbf{low capacity} network for $a$: $a=\MLP{2-20-20-1}$. Dynamic MNIST}
    \label{fig:app:2dlatent_reject_small_a}
\end{figure}

\begin{figure}[htb]
    \centering
    \begin{tikzpicture}
    \node (q) {\includegraphics[height=3.75cm]{figures/2dlatent/2d_latent_regular_qz.pdf}};
    \node[right=0.25cm of q] (pi) {\includegraphics[height=3.75cm]{figures/2dlatent/2dlatent_reject_pi.pdf}};
    \node[right=0.25cm of pi] (a) {\includegraphics[height=3.75cm]{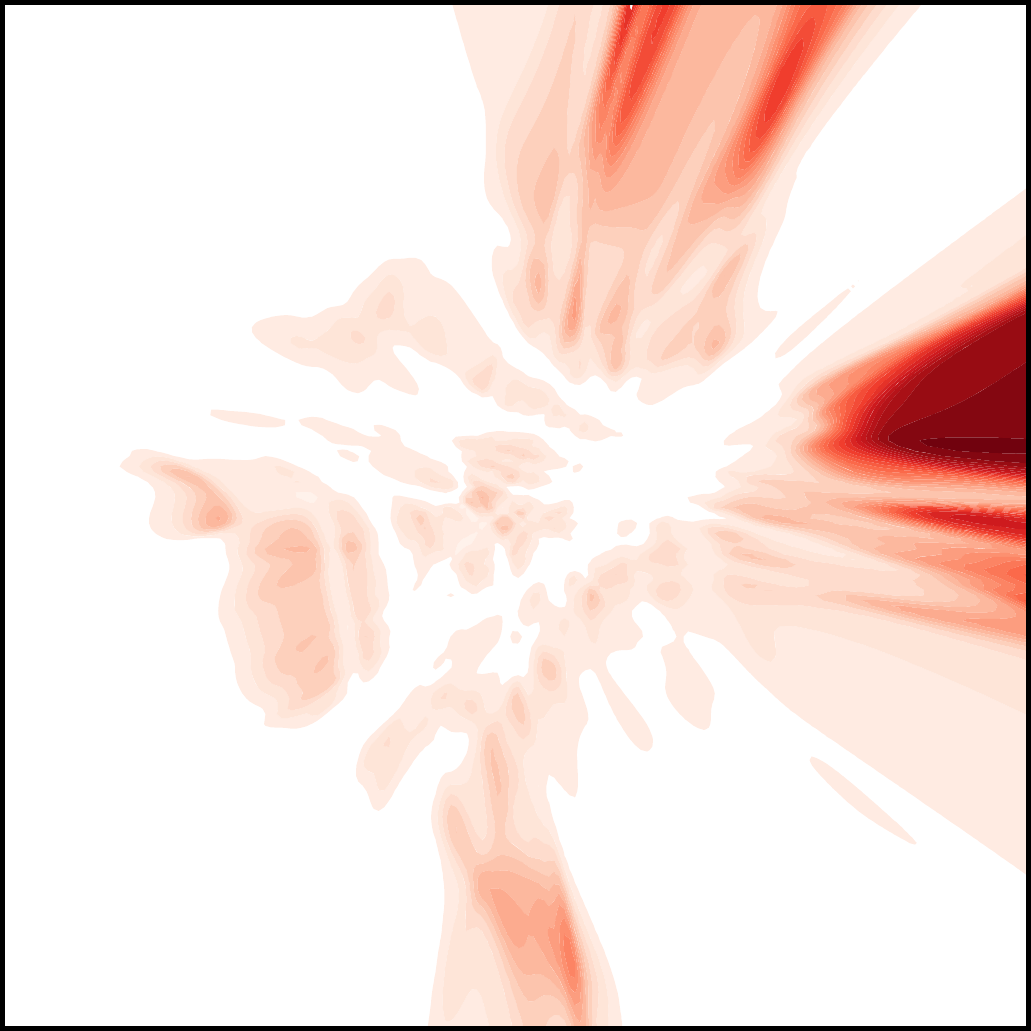}};
    \node[right=0.25cm of a] (p) {\includegraphics[height=3.75cm]{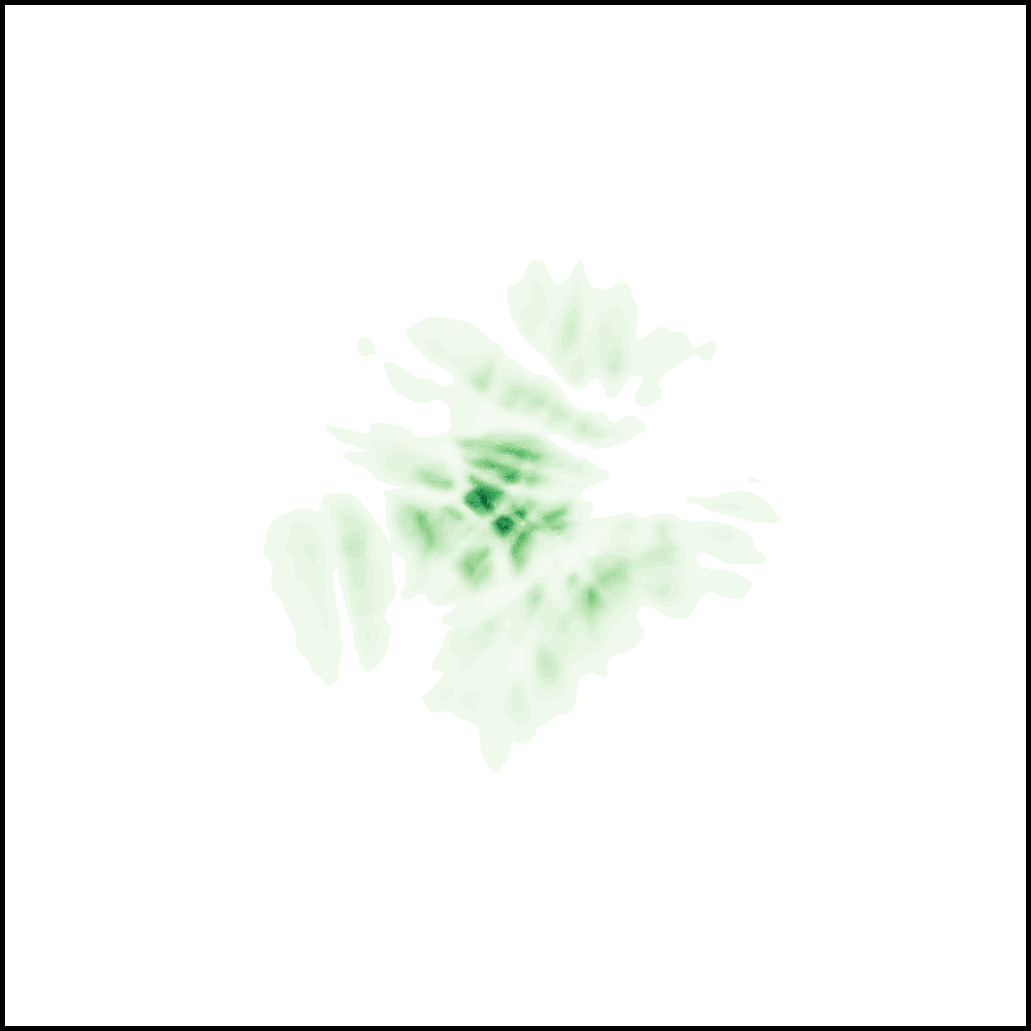}};
    \node[anchor=south] at (q.north) {aggregate posterior $q(\vz)$};
    \node[anchor=south] at (pi.north) {proposal $\pi(\vz)={\mathcal{N}(0,1)}$};
    \node[anchor=north] at (pi.south) {\small $\kld*{q(\vz)\!}{\!\pi(\vz)}\approx 0.4$};
    \node[anchor=south] (a_text) at (a.north) {$a(\vz)$ \MLP{2-100-100-1}};
    \node[anchor=north] at (a.south) {\small $Z\approx 0.06$};
    \node[anchor=south] at (p.north) {\Lars{} prior $p_{T=100}(\vz)$};
    \node[anchor=north] at (p.south) {\small $\kld*{q(\vz)\!}{\!p(\vz)}\approx 0.1$};
    \end{tikzpicture}
    \caption[]{Aggregate posterior and resampled prior that has been trained post-hoc on the pretrained regular VAE. Dynamic MNIST}
    \label{fig:app:2dlatent_finetune}
\end{figure}

\begin{figure}[htb]
    \centering
    \begin{tikzpicture}
    \node (classes_reject) {\includegraphics[height=5.5cm]{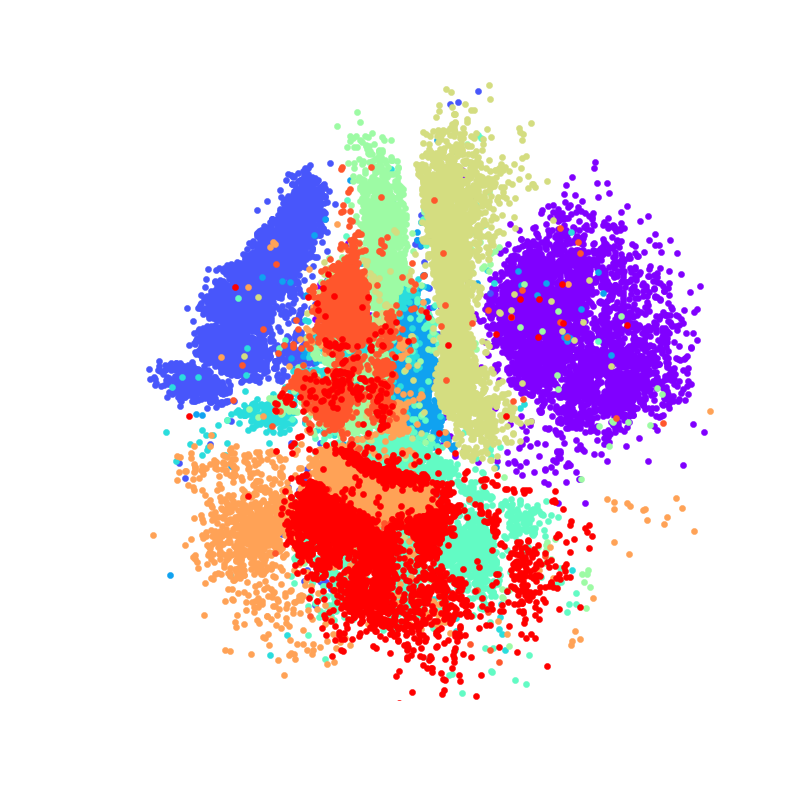}};
    \node[right=0.25cm of classes_reject] (classes_accept) {\includegraphics[height=5.5cm]{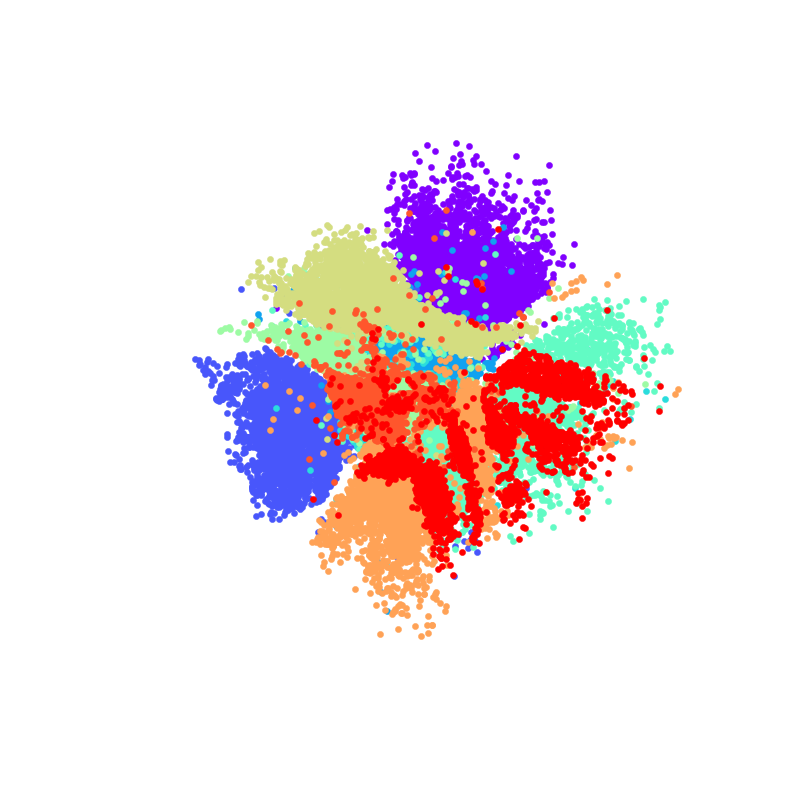}};
    \node[right=0.25cm of classes_accept] (classes_reject_small_a) {\includegraphics[height=5.5cm]{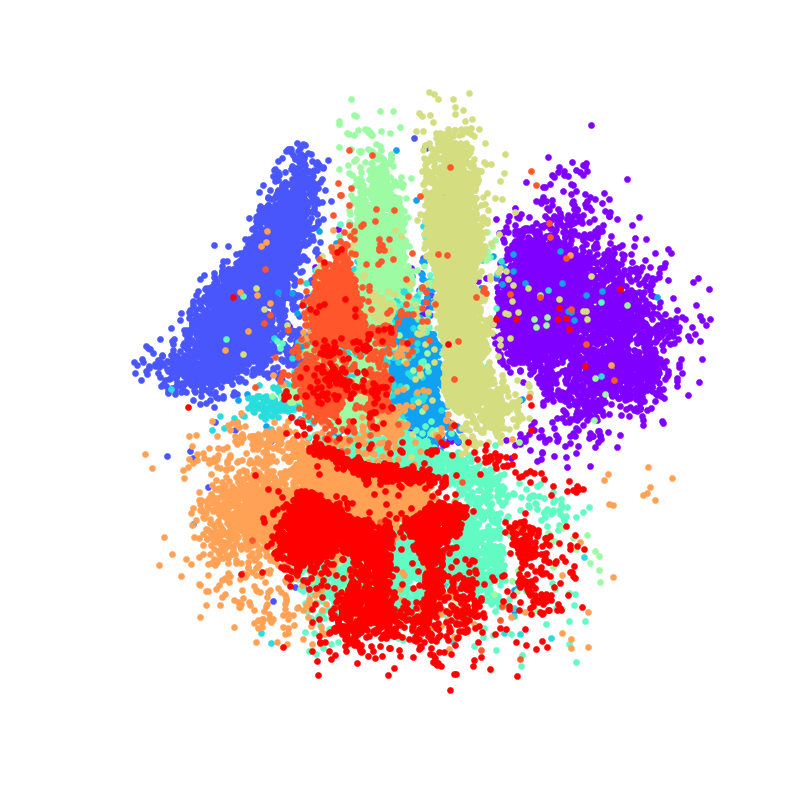}};
    \node[anchor=north] at (classes_reject.south) {\parbox{5cm}{\centering VAE with \Lars{} prior\\ $a=\MLP{2-100-100-1}$}};
    \node[anchor=north] at (classes_accept.south) {VAE with standard prior};
    \node[anchor=north] at (classes_reject_small_a.south) {\parbox{5cm}{\centering VAE with \Lars{} prior\\ $a=\MLP{2-20-20-1}$}};
    \end{tikzpicture}
    \caption[]{Embedding of the MNIST training data into the latent space of a VAE. Colours indicate the different classes and all plots have the same scale.}
    \label{fig:app:2dlatent_classes}
\end{figure}
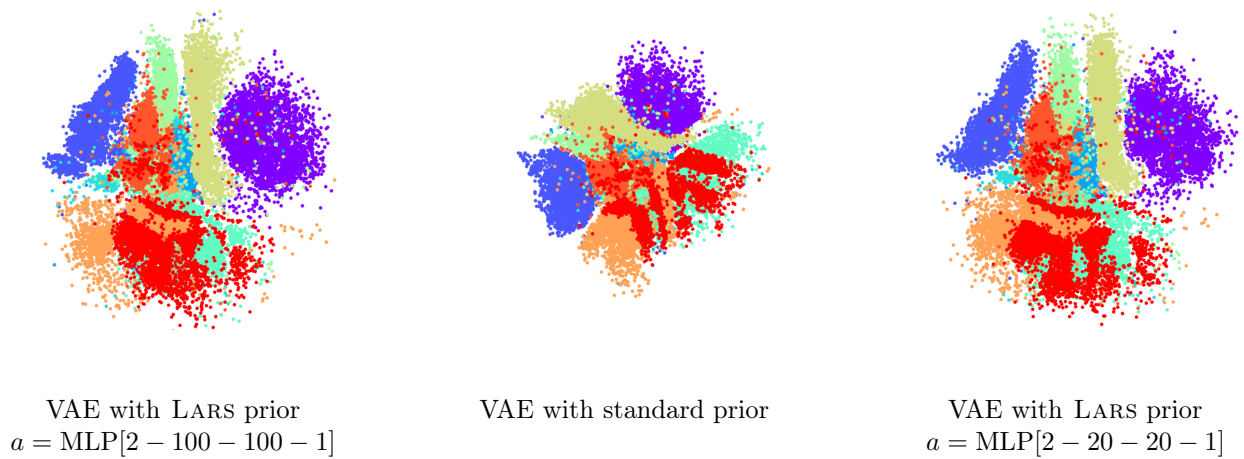

For visualization purposes, we also trained a VAE with a two-dimensional latent space, $d_\vz=2$. We trained both a regular VAE with standard Normal prior as well as VAEs with resampled priors. We considered three different cases that we detail below:
\begin{inlinelist}
    \item a VAE with a resampled prior with expressive/large network as  $a$ function;
    \item a VAE with a resampled prior with limited/small network as $a$ function; and
    \item a pretrained VAE to which we apply \Lars{} post-hoc, that is, we fixed encoder/decoder and only trained the accept function $a$.
\end{inlinelist}
We show the following:
\begin{itemize}
    \item \cref{fig:app:2dlatent_regular}: The aggregate posterior, that is, the mixture density of all encoded training data, $q(\vz) = \frac{1}{N}\sum_n^N q(\vz|\vx)$, as well as the standard Normal prior for a \textbf{regular VAE with standard Normal prior}. We find that the mismatch between standard Normal prior and aggregate posterior is $\kld*{q(\vz)\!}{\!p(\vz)}\approx 0.4$.
    \item \cref{fig:app:2dlatent_reject}: The aggregate posterior, proposal, accept function and resampled density for the\textbf{ VAE with jointly trained resampled prior} with high capacity/expressive accept function. We found that the \Lars{} prior matches the aggregate posterior very well and has $\kld*{q(\vz)\!}{\!p(\vz)}\approx 0.15$. We note that the aggregate posterior is spread out more compared to the regular VAE, see also \cref{fig:app:2dlatent_classes} and note that the KL between the proposal and the aggregate posterior is larger than for the regular VAE. The accept function slices out parts of the prior very effectively and redistributes the weight towards the sides. 
    \item \cref{fig:app:2dlatent_reject_small_a}: Same as \cref{fig:app:2dlatent_reject} but with a smaller network for the acceptance function. We used the same random seed for both experiments and notice that the structure of the resulting latent space is very similar but more coarsely partitioned compared to the previous case. The final KL between aggregate posterior and resampled prior is slightly larger, $\kld*{q(\vz)\!}{\!p(\vz)}\approx 0.20$, which we attribute to the lower flexibility of the $a$ function.
    \item \cref{fig:app:2dlatent_finetune}: We also consider the case of applying \Lars{} post hoc to the pretrained regular VAE, that is, we only learn $a$ on a fixed encoder/decoder. Thus, the aggregate posterior is identical to \cref{fig:app:2dlatent_regular}. However, we find that the acceptance function is able to modulate the standard Normal proposal to fit the aggregate posterior very well and even slightly better than in the jointly trained case. Note that the aggregate posterior is the same as in \cref{fig:app:2dlatent_regular} but that the \Lars{} prior matches it much better than the standard Normal proposal.
    \item \cref{fig:app:2dlatent_classes}: Scatter plots of the encoded means for all training data points for both the VAE with standard Normal prior as well as the \Lars{}/resampled prior with large and small network for the acceptance function. Colours indicate the different MNIST classes.
\end{itemize}

\FloatBarrier

\printbibliography
\end{refsection}
\end{document}

%% file: figures/2d_toy_figure.tex
\begin{tikzpicture}
\node (target) {\includegraphics[height=2cm]{figures/2d_toy_realnvp/target.pdf}};
\node[right=0.15cm of target] (a_high) {\includegraphics[height=2cm]{figures/2d_toy/a_high_v2.pdf}};
\node[right=0cm of a_high] (p_high)
{\includegraphics[height=2cm]{figures/2d_toy/p_high_v2.pdf}};

\node[right=0.15cm of p_high] (a_low) {\includegraphics[height=2cm]{figures/2d_toy/a_low_v2.pdf}};
\node[right=0.cm of a_low] (p_low) {\includegraphics[height=2cm]{figures/2d_toy/p_low_v2.pdf}};
\node[right=0.15cm of p_low] (pi_realnvp) {\includegraphics[height=2cm]{figures/2d_toy_realnvp/pi_nvp_v2.pdf}};
\node[right=0.cm of pi_realnvp] (a_realnvp) {\includegraphics[height=2cm]{figures/2d_toy_realnvp/a_v2.pdf}};
\node[right=0.cm of a_realnvp] (p_realnvp_lars) {\includegraphics[height=2cm]{figures/2d_toy_realnvp/p_v2}};
\draw[dashed] ($(target.south)!0.5!(a_high.south)- (0, 0.4)$) -- ($(target.north)!0.5!(a_high.north) + (0, 0.5)$);
\draw[dashed] ($(p_high.south)!0.5!(a_low.south)- (0, 0.4)$) -- ($(p_high.north)!0.5!(a_low.north) + (0, 0.5)$);
\draw[dashed] ($(p_low.south)!0.5!(pi_realnvp.south)- (0, 0.4)$) -- ($(p_low.north)!0.5!(pi_realnvp.north) + (0, 0.5)$);
\node[anchor=center] at (target.north) {target $q(\vz)$\strut};
\node[anchor=center] (p_high_text) at (p_high.north) {$p(\vz)$\strut};
\node[anchor=north] at (p_high.south) {\small $\kld*{q\!}{\!p}=0.06$};
\node[anchor=center] (a_high_text) at (a_high.north) {$a(\vz)$\strut};
\node[anchor=north] at (a_high.south) {\small $Z=0.09$};
\node[anchor=center] (p_low_text) at (p_low.north) {$p(\vz)$\strut};
\node[anchor=north] at (p_low.south) {\small $\kld*{q\!}{\!p}=0.66$};
\node[anchor=center] (a_low_text) at (a_low.north) {$a(\vz)$\strut};
\node[anchor=north] at (a_low.south) {\small $Z=0.17$};
\node[anchor=center] at (pi_realnvp.north) {$\pi_\mathrm{RealNVP}(\vz)$\strut};
\node[anchor=north] at (pi_realnvp.south) {\small $\kld*{q\!}{\!\pi}=1.21$};
\node[anchor=center] (a_realnvp_text) at (a_realnvp.north) {$a(\vz)$\strut};
\node[anchor=north] at (a_realnvp.south) {\small $Z=0.18$};
\node[anchor=center] at (p_realnvp_lars.north) {$p(\vz)$ \strut};
\node[anchor=north] at (p_realnvp_lars.south) {\small $\kld*{q\!}{\!p}=0.08$};
\node[anchor=south] at ($(a_realnvp_text.north)-(0,0.2)$) {\parbox{4cm}{\centering RealNVP + \Lars{}\\ {\small high capacity $a(\vz)$}}};
\node[anchor=south] at ($(a_high_text.north)!0.5!(p_high_text.north)-(0,0.2)$) {\parbox{4.5cm}{\centering$\mathcal{N}(0,1)$ + \Lars{}\\ {\small high capacity $a(\vz)$: MLP[10-10]}}}; %
\node[anchor=south] at ($(a_low_text.north)!0.5!(p_low_text.north)-(0,0.2)$) {\parbox{4cm}{\centering$\mathcal{N}(0,1)$ + \Lars{}\\ {\small low capacity $a(\vz)$: MLP[10]}}};
\end{tikzpicture}

%% file: figures/mnist_mlp_Z_trace.tex
\begin{tikzpicture}
\begin{semilogxaxis}[
	xlabel=Number $S$ of samples used to estimate $Z$,ylabel=estimate $Z_S$ for $Z$,
	 y label style={at={(-0.1,0.5)}},
	height=5cm,
	width=8cm,
	yticklabel style={
        /pgf/number format/fixed,
        /pgf/number format/precision=5
},scaled y ticks=false,
	xmin=1e5,
	xmax=2e10,]
\addplot[color=C1, line width=1.5pt] table {figures/1729353_wid5_Z_trace.txt};
\end{semilogxaxis}%
\end{tikzpicture}

%% file: figures/mnist_mlp_samples_ranked.tex
\begin{tikzpicture}
\begin{semilogyaxis}[
	xlabel=Index of sample ,ylabel=$a(\mathbf{z})$,
	height=5cm,
	width=8cm,
	ymin=1e-22,
	ymax=1,
	xmin=1e0,
	xmax=1e4,
	ytick={1e-22,1e-16, 1e-10,1e-4,1}]
\addplot[color=C1, line width=2pt] table {figures/samples/mnist_mlp_ranked/1729353_wid5.txt};
\end{semilogyaxis}%
\end{tikzpicture}